\documentclass[conference]{IEEEtran}

\usepackage{graphicx}
\usepackage{booktabs}

\usepackage[accsupp]{axessibility}  
\usepackage{caption}
\usepackage{bbm}
\usepackage{arydshln}

\usepackage{listings}
\usepackage{placeins}
\usepackage{nicefrac}
\usepackage{xurl}
\usepackage{enumitem}
\usepackage{tabularx}
\usepackage{nicefrac}
\usepackage{makecell}
\usepackage{xspace}
\usepackage{graphbox}
\usepackage{pifont}
\usepackage{enumitem}
\usepackage{colortbl}
\usepackage{multirow}
\usepackage{diagbox}
\usepackage{array}
\usepackage{wrapfig}
\usepackage{subcaption}
\usepackage{footmisc}
\usepackage{color}
\usepackage{stfloats}
\usepackage{slashbox}
\usepackage{diagbox}
\usepackage{float}
\usepackage{mathrsfs}
\usepackage{apptools}

\usepackage[pagebackref,breaklinks,colorlinks
]{hyperref}
\usepackage[capitalize,noabbrev]{cleveref}
\Crefname{equation}{Eq.}{Eqs.}

\usepackage[dvipsnames]{xcolor}
\usepackage{tcolorbox}

\newcolumntype{C}[1]{>{\centering\arraybackslash}p{#1}}
\newcolumntype{L}[1]{>{\raggedright\arraybackslash}p{#1}}
\newcolumntype{R}[1]{>{\raggedleft\arraybackslash}p{#1}}
\newlength\newl
\newlength\newlc

\newlength\colwidth
\newlength\figwidth

\newcommand{\vit}{ViT\xspace}

\newcommand{\clip}{CLIP\xspace}
\newcommand{\convnext}{ConvNeXt\xspace}

\newcommand{\apgdce}{APGD\textsubscript{CE}\xspace}

\newcommand{\imnet}{ImageNet\xspace}
\newcommand{\ours}{FARE\xspace}

\newcommand{\tecoa}{TeCoA\xspace}
\newcommand{\lipsim}{LipSim\xspace}

\newcommand{\lora}{LoRA\xspace}
\newcommand{\rclip}{R-CLIP\textsubscript{T}\xspace}
\newcommand{\rclipf}{R-CLIP\textsubscript{F}\xspace}
\newcommand{\rdinof}{R-DINO\textsubscript{F}\xspace}
\newcommand{\rdinoviif}{R-DINOv2\textsubscript{F}\xspace}
\newcommand{\oxford}{$\mathcal{R}$Oxford\xspace}
\newcommand{\paris}{$\mathcal{R}$Paris\xspace}

\def\R{\mathbb{R}}

\newcommand{\inner}[1]{\left\langle#1\right\rangle}

\def\R{\mathbb{R}}

\newcommand{\norm}[1]{\left\|#1\right\|}

\newcommand{\simfn}{\texttt{sim}\xspace}
\newcommand{\clf}{\texttt{clf}\xspace}

\definecolor{brightpurple}{RGB}{160,0,255}
\newcommand{\rev}[1]{
#1\xspace}

\newcommand\blfootnote[1]{%
  \begingroup
  \renewcommand\thefootnote{}\footnote{#1}%
  \addtocounter{footnote}{-1}%
  \endgroup
}

\newcommand{\eqcontr}{$^*$}

\definecolor{cvprblue}{rgb}{0.21,0.49,0.74}
\definecolor{lightgray}{rgb}{0.9,0.9,0.9}
\definecolor{BlueGray}{rgb}{1, 0.8, 0.8}
\definecolor{lightgreen}{rgb}{0.90, 0.99, 0.85}
\definecolor{darkgreen}{rgb}{.1, .85, .1}
\definecolor{newgray}{rgb}{0., 0., 0.} 
\definecolor{graygreen}{rgb}{.1, .75, .1} 
\definecolor{grayred}{rgb}{1., 0., 0.} 
\definecolor{lightorange}{rgb}{1., .9, 0.}
\definecolor{lighttred}{rgb}{1., .9, 0.8}
\definecolor{teaser1}{HTML}{FFCCBC}
\definecolor{teaser1}{HTML}{FFCCBC}

\usepackage{url}

\usepackage{amsmath,amsfonts,bm}

\def\eqref#1{equation~\ref{#1}}

\def\1{\bm{1}}

\def\vt{{\bm{t}}}

\def\vx{{\bm{x}}}
\def\vy{{\bm{y}}}
\def\vz{{\bm{z}}}

\DeclareMathAlphabet{\mathsfit}{\encodingdefault}{\sfdefault}{m}{sl}
\SetMathAlphabet{\mathsfit}{bold}{\encodingdefault}{\sfdefault}{bx}{n}

\def\vdelta{{\bm{\delta}}}

\DeclareMathOperator*{\argmax}{arg\,max}

\begin{document}

\title{Adversarially Robust CLIP Models Can Induce Better (Robust) Perceptual Metrics}

\author{
\IEEEauthorblockN{Francesco Croce\eqcontr\IEEEauthorrefmark{2} \hspace{3mm} Christian Schlarmann\eqcontr\IEEEauthorrefmark{3} \hspace{3mm} Naman Deep Singh\eqcontr\IEEEauthorrefmark{3} \hspace{3mm} Matthias Hein\IEEEauthorrefmark{3}}
\vspace{1.5mm}
\IEEEauthorblockN{\IEEEauthorrefmark{2}EPFL, Switzerland \hspace{3mm} \IEEEauthorrefmark{3}Tübingen AI Center, University of Tübingen, Germany}
}

\maketitle

\begin{abstract}
  Measuring perceptual similarity is a key tool in computer vision. In recent years perceptual metrics based on features extracted from neural networks with large and diverse training sets, e.g. CLIP, have become popular. At the same time, the metrics extracted from features of neural networks are not adversarially robust. In this paper we show that adversarially robust CLIP models, called \rclipf, obtained by unsupervised adversarial fine-tuning induce a \emph{better} and \emph{adversarially robust} perceptual metric that outperforms existing metrics in a zero-shot setting, and further matches the performance of state-of-the-art metrics while being robust after fine-tuning. Moreover, our perceptual metric achieves strong performance on related tasks such as robust image-to-image retrieval, which becomes especially relevant when applied to ``Not Safe for Work'' (NSFW) content detection and dataset filtering. While standard perceptual metrics can be easily attacked by a small perturbation completely degrading NSFW detection, our robust perceptual metric maintains high accuracy under an attack while having similar performance for unperturbed images. Finally, perceptual metrics induced by robust CLIP models have higher interpretability: feature inversion can show which images are considered similar, while text inversion can find what images are associated to a given prompt. This also allows us to visualize the very rich visual concepts learned by a CLIP model, including memorized persons, paintings and complex queries.
\end{abstract}


\section{Introduction}
\label{sec:intro}

\blfootnote{\eqcontr Equal contribution.}

A longstanding goal in computer vision is finding a \mbox{metric} which is able to accurately mimic the human perception of similarity of images.
This would benefit multiple tasks such as dataset filtering, image retrieval, copyright infringement discovery, and image quality assessment.
While the first approaches to perceptual metrics relied on analyzing statistical properties of the images \cite{wang2004image}, the development of deep learning brought metrics based on internal representations of trained models which are fine-tuned on human similarity judgments using Two Alternatives Forced Choice (2AFC) tasks.
The most prominent example is the LPIPS distance \cite{zhang2018unreasonable}.
More recently, the proximity in the embedding space of large 
foundation models 
such as CLIP \cite{radford2021clip}, DINO \cite{caron2021emerging}, and Masked Autoencoders (MAE) \cite{he2021masked} has been shown to effectively capture the semantic similarity of images \cite{fu2023learning}. Fine-tuning the representations of such models on novel 2AFC tasks lead to the recent DreamSim distance \cite{fu2023learning} providing significant improvements in perceptual similarity assessment.

\begin{figure*}[t]
\centering

\tcbset{colframe=white, left=0pt, right=0pt, top=2pt, bottom=2pt, boxrule=0pt, arc=5pt}

\begin{minipage}[c]{1.28\columnwidth}
\centering \small
\begin{tcolorbox}[colback=blue!10!white, width=\columnwidth]
\centering \small
\textbf{(Clean and robust) SOTA zero-shot 2AFC performance}\par\medskip

\begin{subfigure}{.96\columnwidth}
\centering \footnotesize
\figwidth=.9\columnwidth
\includegraphics[width=\figwidth, trim=1mm 1mm 0mm 1mm, clip]{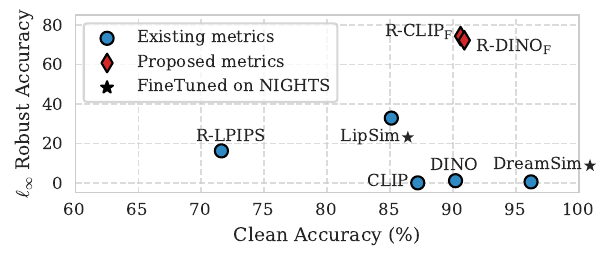}

    \caption{
    \rev{We report the clean and robust performance of several perceptual metrics on 
    the NIGHTS dataset \cite{fu2023learning}. The adversarially trained  \rclipf  (ConvNeXt-B) and \rdinof (ViT-B/16) models achieve higher both clean and robust accuracy than their non-robust counterparts (CLIP, DINO), and have SOTA zero-shot performance.}
     }
     \label{fig:teaser_2afc}
\end{subfigure}
\end{tcolorbox}



\begin{tcolorbox}[colback=green!10!white, width=\columnwidth]
\centering \small

\textbf{Robust detection of NSFW content}\par\medskip

\begin{subfigure}{.96\columnwidth}
\centering
\figwidth=.9\columnwidth
\includegraphics[width=\figwidth, trim=1mm 1mm 0mm 1mm, clip]{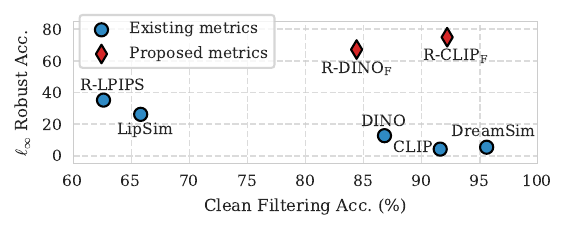}


    \caption{
    \rev{We test perceptual metrics to detect ``Not Safe for Work'' (NSFW) images via image-to-image retrieval. \rclipf achieves clean accuracy close to that of \clip and DreamSim ($>$90\%) while being robust against $\ell_\infty$-attacks ($\epsilon_\infty=8/255$) which aim to turn unsafe into safe images (75\% robust accuracy vs $<$40\% of competitors).}
    }
    \label{fig:teaser_nsfw}
\end{subfigure}
\end{tcolorbox}

\end{minipage}
\hfill
\begin{minipage}[c]{.73\columnwidth}
\centering \small

\begin{tcolorbox}[colback=orange!10!white, width=\columnwidth]
\centering \small
\textbf{Interpretability of perceptual metrics} 
\par\medskip
\begin{subfigure}{.96\columnwidth}
\figwidth=.48\columnwidth
\tabcolsep=2pt
\centering \footnotesize
\begin{tabular}{cc}
\multicolumn{2}{c}{Original}\\
\includegraphics[width=0.9\figwidth]{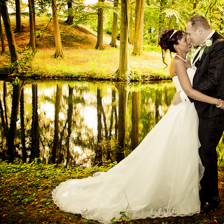} & \includegraphics[width=0.9\figwidth]{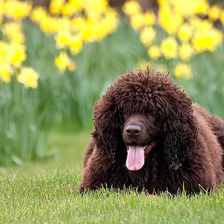} \\[1.2mm]
\multicolumn{2}{c}{Reconstructed with CLIP}\\
\includegraphics[width=0.9\figwidth]{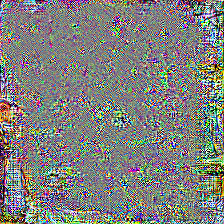} & \includegraphics[width=0.9\figwidth]{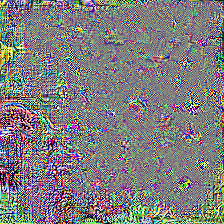} \\[1.2mm]
\multicolumn{2}{c}{Reconstructed with \rclipf (ours)}\\
\includegraphics[width=0.9\figwidth]{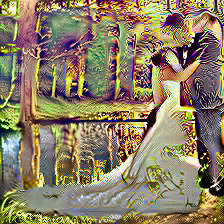} &
\includegraphics[width=0.9\figwidth]{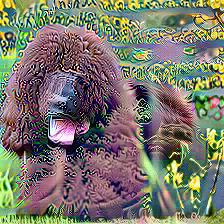}
\end{tabular}
\caption{\textbf{Feature inversion.} Starting from a grey image we maximize the cosine similarity to the embedding of an image  (original), once for CLIP and once for our \rclipf. With \mbox{\rclipf}, we get semantically correct reconstructions, whereas the clean CLIP model produces only adversarial noise.
}
\label{fig:teaser_feature_inversion}
\end{subfigure}
\end{tcolorbox}
\end{minipage}

\caption{Our perceptual metric \rclipf performs similar to DreamSim across tasks and is by far the most robust one.}
\label{fig:teaser}
\end{figure*}

A line of work has further focused on studying the adversarial robustness of perceptual similarity metrics, showing that, as extensively observed for image classifiers or segmentation models,
they are extremely brittle against even imperceptible perturbations \cite{kettunen2019elpips,sjogren2022identifying,ghildyal2022shifttolerant, ghildyal2023attacking}.
This might become especially problematic in tasks where an adversary has interest in bypassing automatic similarity checks, e.g. in image attribution, content filtering \cite{andriushchenko2022aria}, 
or ``Not Safe for Work'' (NSFW) detection in large scale datasets.
Recently, \rev{Ghazanfari et al.}~\cite{ghazanfari2023rlpips} proposed \rev{first} R-LPIPS, an empirically robust version of LPIPS against $\ell_p$-bounded adversarial perturbations, and \rev{later} introduced LipSim \cite{ghazanfari2023lipsim}, a first perceptual metric 
with certified robustness against $\ell_2$-bounded perturbations distilled from DreamSim.
However, 
their improvements in empirical and even more so provable robustness 
come at significant cost in clean performance, see \cref{fig:teaser}, which questions their practical utility as a perceptual metric as well as for applications like NSFW detection.

\rev{In this work, we first uncover how one can obtain perceptual metrics which simultaneously achieve SOTA clean and robust  zero-shot accuracy on 2AFC tasks.
In particular, we fine-tune the vision encoders of CLIP models with recent adversarial training methods \cite{Mao2022UnderstandingZAtecoa,schlarmann2024robust}.
The resulting models induce perceptual metrics that significantly outperform their non-robust counterparts (in a \emph{zero-shot setting}), and are even close to or better than specialized metrics fine-tuned on 2AFC datasets (\cref{fig:teaser_2afc}).
Moreover, these outperform previous approaches for robust perceptual metrics such as R-LPIPS and LipSim, which have even been trained on 2AFC datasets. 
Interestingly, these findings show that there exist tasks where the commonly observed trade-off between clean and robust performance \cite{Tsipras2019RobustnessOddsAcc, croce2020RobustBench} does not occur.
For example, the adversarial fine-tuning technique FARE \cite{schlarmann2024robust} yields robust \clip models, \rclipf, that achieve, after additional fine-tuning on 2AFC datasets, performance competitive with the recent state-of-the-art perceptual metric DreamSim but is additionally adversarially robust.
}
We also confirm this observation by extending FARE to the DINO architecture: the resulting \rdinof outperforms the original DINO as a perceptual metric.
Moreover, while previous work on perceptual metrics has solely focused on CLIP models with vision transformers (ViTs) as encoder, we show that the stronger inductive bias of convolutions in \convnext may be particularly helpful in this task.

\rev{Besides 2AFC datasets, an adversarially robust perceptual metric can be used for robust detection or content filtering via image-to-image retrieval, as commonly done in several tasks \cite{schuhmann2022laion5b}.
On an NSFW detection task, our \rclipf performs similarly to 
the original non-robust CLIP and DreamSim, while being significantly more robust (\cref{fig:teaser_nsfw}): in this case, adversarial robustness is particularly relevant as malicious actors have a clear incentive to target these detection models.}
We show similar findings for other tasks such as image retrieval and dataset filtering.

Finally, 
\rev{we illustrate, via feature inversion (inverting a given image embedding, \rev{see \cref{fig:teaser_feature_inversion}}) and text inversion (generating images from captions) experiments, an additional advantage of adversarially robust perceptual metrics, i.e. interpretability.
In fact, we can}
explore which images and concepts are considered similar by the perceptual metrics, as well as extract the ``visual concepts'' (e.g. famous paintings and people) memorized by CLIP during pre-training.
\\

\textbf{
Contributions.} In summary, our main findings include that

\begin{itemize}[itemsep=1.2mm, topsep=2mm, wide, parsep=1.2mm, 
left=0mm]
\item 
adversarially trained CLIP encoders achieve SOTA zero-shot (no fine-tuning on perceptual datasets) results on 2AFC tasks, significantly outperforming their clean equivalent, which is in contrast to most scenarios where adversarial training degrades clean performance.

\item 
robust CLIP models yield SOTA \textit{robust} perceptual metrics:
the CLIP vision encoders fine-tuned with $\ell_\infty$-adversarial training on ImageNet outperform existing robust perceptual metrics for 2AFC tasks. 

\item robust CLIP models enable robust image-to-image retrieval, which can be deployed for robust detection of NSFW content and filtering of harmful content from datasets.

\item 
\rev{when studying the effect of different types of vision encoders as perceptual metrics,} ConvNeXts, a convolutional architecture, 
often perform better
than ViTs, while previous work solely focused on transformer-based encoders. Moreover, we train the first adversarially robust DINO models and systematically compare it to \rclipf. 

\item the perceptual metrics induced by robust encoders have higher interpretability than with clean encoders, i.e. it is possible to visualize which type of similarity they encode with a straightforward optimization approach.\footnote{Our code and models are available at \url{https://github.com/fra31/perceptual-metrics}}

\end{itemize}

\section{Related Work}\label{sec:related}

\textbf{Perceptual metrics.}
Low-level pixel-based $\ell_p$-metrics and structural similarity SSIM \cite{wang2004image} do not capture well higher-order semantic similarity.
These are outperformed by metrics based on features extracted from neural networks trained on ImageNet, such as
LPIPS \cite{zhang2018unreasonable}, 
PIE-APP \cite{prashnani2018pieapp} and DISTS \cite{Ding_2020}. 
More recently, it has been shown that metrics induced by the features extracted by models trained on larger datasets and via self-supervised training, like CLIP \cite{radford2021clip}, DINO \cite{caron2021emerging} or MAE \cite{he2021masked}, are well aligned with human perception regarding semantic similarity \cite{fu2023learning}. 
DreamSim \cite{fu2023learning} is a recent fine-tuned ensemble of three of these models on the 2AFC dataset NIGHTS which shows the best alignment with human preferences on NIGHTS.
\rev{In our work we show how to improve the performance of CLIP- and DINO-based perceptual metrics.}
\\

\textbf{Adversarial robustness of perceptual metrics.}
Virtually all vision tasks tackled via neural networks are vulnerable to adversarial examples \cite{Szegedy2014AdvExamples}, 
and attacks in several threat models exist \cite{Carlini2017DetectionBypassing,Croce2020Autoattack,Laidlaw2019ThreatModelPerceptibility}.
The main empirical defense which works across different vision tasks is adversarial training \cite{Madry2018AT}. However, the price of having a robust model is typically a drop in performance.
Not surprisingly, perceptual metrics, including LPIPS and DreamSim, are also not robust to adversarial perturbations \cite{ghazanfari2023rlpips,ghazanfari2023lipsim,ghildyal2023attacking}.
In order to get a robust version R-LPIPS of the popular LPIPS metric, Ghazanfari et al. \cite{ghazanfari2023rlpips} perform adversarial training on the 2AFC 
fine-tuning task of the Berkeley-Adobe Perceptual Patch Similarity dataset (BAPPS) \cite{zhang2018unreasonable}.
LipSim \cite{ghazanfari2023lipsim} distills from DreamSim \cite{fu2023learning} a 1-Lipschitz network, and then fine-tunes it on the NIGHTS dataset \cite{fu2023learning}
to achieve certified adversarial robustness.
\rev{Conversely, we will leverage recent techniques for robust zero-shot classification with CLIP to obtain perceptual metrics with SOTA adversarial robustness.}
\\

\textbf{
\rev{Interpretability} of adversarially robust models.}
Feature inversion, i.e. finding an image which matches given features at the output layer, can be used to understand the inner workings of a network. However, it often yields highly distorted images without much semantic content \cite{mahendran2014understanding}.
Notably, adversarially robust models suffer significantly less from this problem,
and can be used to generate semantically meaningful images when maximizing the probability of a specific class \cite{santurkar2019imagesynthesis}.
This 
can be even exploited to generate visual counterfactuals (instance-specific explanations) for modern image classifiers~\cite{augustin2020ratio, boreiko2022sparsevisual}.
\rev{Similarly, robust classifiers are known to have more interpretable gradients than standard ones \cite{santurkar2019imagesynthesis}}.
\rev{Our work illustrates that such properties yield more interpretable perceptual metrics.}

\section{
Background} \label{sec:robust_clip}

\textbf{Adversarially Robust \clip Models.}
\clip models \cite{radford2021clip} consist of an image encoder $\phi: I \rightarrow \R^D$ and a text encoder $\psi: T \rightarrow \R^D$, which map different types of data into the same $D$-dimensional latent space.
The embedding of image-text pairs with corresponding semantic meaning are then aligned in the latent space via contrastive learning using large datasets of image-caption pairs.
These models attain good results in zero-shot classification performance: the $K$ class names are reformulated as text prompts, e.g. $\vt_k=$ ``\texttt{A photo of <class $k$>}'' for $k=1,\ldots,K$, and embedded via the text encoder as $\psi(\vt_k)$.
The predicted class for an image $\vx$ is the one whose text embedding has the highest cosine similarity to the image embedding. 
As for image classifiers obtained by supervised learning, the zero-shot CLIP classifiers are vulnerable to adversarial perturbations \cite{Fort2021CLIPadversarial, Mao2022UnderstandingZAtecoa}, in particular in the $\ell_p$-bounded threat models.

Recent works have extended adversarial training \cite{Madry2018AT} to \clip by fine-tuning the image encoder of an existing non-robust \clip model against $\ell_\infty$-bounded perturbations.
\tecoa \cite{Mao2022UnderstandingZAtecoa} 
performs supervised adversarial training with the loss
\begin{align} \label{eq:tecoa}
\mathcal{L}_\textrm{\tecoa}(f, \vx, y)= \max_{\norm{\vdelta}_\infty \leq \epsilon} -\log\left(\frac{e^{f_y(\vx + \vdelta)}}{\sum_{k=1}^K e^{f_k(\vx + \vdelta)}}\right),
\end{align}
where $(\vx, y)$ are image-label pairs from \imnet and $f_k(\vx)=\cos(\phi(\vx),\psi(\vt_k))$ 
(the resulting robust CLIP models are denoted as \rclip), 
while \ours \cite{schlarmann2024robust}
formulates the unsupervised learning problem
\begin{align} \label{eq:fare}
\mathcal{L}_\textrm{FARE}(\phi, \vx)= \max_{\norm{\vdelta}_\infty \leq \epsilon} \,\norm{\phi_\textrm{orig}(\vx)-\phi(\vx + \vdelta)}^2_2,
\end{align}
where one aims at low distortions in the embedding space compared to the original embedding $\phi_\textrm{orig}$ of clean images even after adversarial perturbations. \ours has the advantage that it preserves the original embedding and thus the resulting \rclipf models are compatible with the text embedding (\rev{beyond the ImageNet classes}) and can be used as a direct replacement of a \clip model for all downstream tasks, e.g. 
generative vision-language models \cite{schlarmann2024robust}.
Since \ours, in contrast to \tecoa, just requires the embedding from the vision encoder $\phi$ and no supervision via text, we use FARE to obtain adversarially robust DINO models which, unlike \clip, consist of a vision encoder only.
\rev{We highlight that neither Mao et al. \cite{Mao2022UnderstandingZAtecoa} nor Schlarmann et al. \cite{schlarmann2024robust} consider the effect of fine-tuning the CLIP vision encoder with adversarial training on the induced perceptual metric.}
\\

\textbf{\clip embedding induces a perceptual metric.}
To measure the similarity of two images $\vx_1, \vx_2 \in I$ it is common to use the cosine similarity of their embedding, i.e. \clip induces the similarity score
\begin{align}
\simfn(\vx_1, \vx_2) = 
\inner{\frac{\phi(\vx_1)}{\norm{\phi(\vx_1)}_2},\frac{\phi(\vx_2)}{\norm{\phi(\vx_2)}_2}}.
\label{eq:cossim}
\end{align}
This in turn induces the pseudometric\footnote{Note that without the square-root the triangle inequality is not fulfilled. 
}
\begin{align*}
d(\vx_1, \vx_2)=& \sqrt{1-\simfn(\vx_1, \vx_2)} \\
=& \frac{1}{\sqrt{2}}\norm{\frac{\phi(\vx_1)}{\norm{\phi(\vx_1)}_2}-\frac{\phi(\vx_2)}{\norm{\phi(\vx_2)}_2}}_2,
\end{align*}
which is used as a perceptual metric.
As shown in \cite{fu2023learning} this metric is well-aligned with human perception on the NIGHTS dataset in a 2AFC task even in a zero-shot setting, i.e. without fine-tuning on NIGHTS.
Moreover, similar scores can be obtained by using the the embedding of other types of encoders such as DINO and MAE.
\\

\textbf{2AFC datasets.}
In Two Alternative Forced Choice (2AFC) tasks, which have been long used in psychology to study human decision making \cite{Link2AFC_1975},
given a reference image $\vx_\textrm{ref}$ one has to decide which out of two images $\vx_1, \vx_2$ is most similar to the reference image (with ground truth label $y\in\{1, 2\}$).
Two popular 2AFC datasets for perceptual metrics are the BAPPS dataset \cite{zhang2018unreasonable}, used to tune the LPIPS distance based on features of AlexNet, and the NIGHTS dataset \cite{fu2023learning}, used to tune the DreamSim metric.
BAPPS contains low resolution ($64\times 64$) patches of real images perturbed with several types of corruptions, and compares their similarity to the original images. Conversely, NIGHTS includes high resolution synthetic images, and aims at capturing similarity in terms of attributes like pose, perspective, foreground color, number of items, and object shape. 
Given a perceptual metric or similarity score, 
one can formulate this problem as a binary classification task: with the \clip embedding we get the classifier
\begin{align}
\clf(\vx_1, \vx_2, \vx_\textrm{ref}) = [\simfn(\vx_\textrm{ref}, \vx_1), \simfn(\vx_\textrm{ref}, \vx_2)] 
\label{eq:classifier_2afc}
\end{align}
which predicts labels as $\argmax_{k=1, 2}\, \clf(\vx_1, \vx_2, \vx_\textrm{ref})$.
A classifier which performs well on such a 2AFC task is well-aligned with human perception.
In particular, CLIP-based perceptual metrics can be used zero-shot, or customized by fine-tuning on 2AFC training data.
\\

\textbf{Attacks on perceptual metrics.}
One can adversarially attack the classifier in~\cref{eq:classifier_2afc} in several ways, applying perturbations either on one of (or both) the test images $\vx_1, \vx_2$ or the reference image $\vx_\textrm{ref}$.
We consider the second option more intuitive as it may influence both similarity comparisons, or equivalently both logits of the classifier,
which mimics an attack scenario for image attribution or content filtering.
This is also in line with previous work in LipSim \cite{ghazanfari2023lipsim}.
The resulting optimization problem for the attack can be formulated as
\begin{align*}
\max_{\norm{\vdelta}_p \leq \epsilon_p}\mathcal{L}\big(\clf(\vx_1, \vx_2, \vx_\textrm{ref}+\vdelta), y\big) \quad \textrm{s.th.} \quad \vx_\textrm{ref} + \vdelta \in I,
\end{align*}
where the constraint $I$ ensures that the perturbed image $\vx_\textrm{ref}+\vdelta$ remains in the image domain.
Similar to image classification, this can be solved with several techniques, most commonly PGD-like attacks \cite{Madry2018AT,Croce2020Autoattack} on a classification loss $\mathcal{L}$, for example cross-entropy.

\section{Evaluation of Perceptual Metrics Induced by Robust Vision Encoders}\label{sec:exp}

In the following we study the effectiveness and robustness of the similarity metrics induced by standard and adversarially robust \clip and DINO models.
In particular, we analyze how the performance depends on their architecture, pre-training and adversarial fine-tuning scheme, both in the zero-shot setting and with fine-tuning on 2AFC datasets.
Finally, in \cref{sec:comparison} we compare \rev{the} robust \clip and DINO models to existing SOTA for standard and robust perceptual metrics. 
\\

\textbf{Setup.}
We consider three \clip models from the OpenCLIP library \cite{cherti2023openclip} with vision encoders using different backbones, i.e. \vit-B/32, \vit-B/16, \convnext-B, all pre-trained on LAION-2B \cite{schuhmann2022laion5b} and with similar number of parameters. 
To get adversarially robust versions of each model, we fine-tune them with \ours and \tecoa on \imnet in the $\ell_\infty$-threat model with radius $\epsilon_\infty=4/255$, following \cite{schlarmann2024robust}: we indicate them as R-CLIP\textsubscript{F} and \rclip respectively.
Moreover, DINO has been shown in \cite{fu2023learning} to induce a good perceptual metric in the zero-shot setting. 
Then, we fine-tune DINO \cite{caron2021emerging} with \vit-B/16 backbone and DINOv2 \cite{oquab2023dinov2} (\vit-B/14) with \ours, and denote the resulting models by \rdinof and \rdinoviif respectively.
Since DINO models do not include a text encoder, it is not possible to directly use \tecoa as it requires the textual encoding of the \imnet classes.
\rev{We note that, while these models are obtained applying existing fine-tuning techniques to additional architectures compared to the original works \cite{Mao2022UnderstandingZAtecoa, schlarmann2024robust}, we may refer to those as ours to differentiate them from the models directly taken from prior works.}
Finally, we report for completeness a larger CLIP model with ViT-L/14 as backbone \cite{radford2021clip}, whose robust versions are from \cite{schlarmann2024robust}.

We test adversarial robustness to $\ell_\infty$-bounded attacks of radius $\epsilon_\infty=4/255$ and $\ell_2$-bounded attacks of size $\epsilon_2=3$, as a proxy for unseen threat models (in fact, we would ideally want perceptual metrics which are robust across threat models, including those not seen at training time).
For computing the attacks we use APGD \cite{Croce2020Autoattack} on the cross-entropy loss for 100 iterations. 
For NIGHTS we report results on the entire test set. For BAPPS the clean accuracy is computed as average over the performance on the 6 dataset splits (entire test set), while for robust accuracy we use 1k examples for each split, and show the average results.
For both datasets the breakdown over splits can be found in Appendix~\ref{app:bapps},
and further details about the experimental setup in Appendix~\ref{app:experiments}.

\subsection{Fine-tuning for $\ell_\infty$-robustness makes \clip and DINO models more aligned with human perception} \label{sec:comparison_clip_models}

\begin{table}[t]
\centering
\small
\caption{\textbf{Comparison of \clip and DINO models with their robust versions on 2AFC tasks across architectures.}
We report clean and robust accuracy of both zero-shot and NIGHTS fine-tuned 
\clip and DINO models with different vision encoders (for models taken from prior work we provide the reference). \emph{All} robust zero-shot CLIP and DINO models (trained for $\ell_\infty$ with $\epsilon_\infty=4/255$) perform better than their clean counterparts in clean \textbf{and} robust accuracy (evalution with $\epsilon_\infty=4/255$ for $\ell_\infty$ and $\epsilon_2=3$ for $\ell_2$).}
\label{tab:comparison_clip_models}
\tabcolsep=1.9pt
\extrarowheight=2pt
\newl=8mm
\newlc=7mm
\begin{tabular}{L{14mm} C{19mm} *{2}{|
>{\columncolor{lightgreen}}C{\newlc} C{\newl} C{\newlc}}}

& & \multicolumn{3}{c}{NIGHTS} & \multicolumn{3}{c}{BAPPS} \\
Method & Encoder & {clean} & {$\ell_\infty$} &{$\ell_2$} & {clean} & $\ell_\infty$ &{$\ell_2$} \\

\toprule
\addlinespace[2mm]

\multicolumn{8}{l}{\textbf{Zero-shot models (DINO, CLIP)}} \\
 \toprule

  DINO & ViT-B/16 \cite{caron2021emerging} & 90.2 & 1.1 & 1.3 & 71.5 & 0.1 & 0.1 \\
  \rdinof &  ViT-B/16 & 90.9 & 72.2 & 73.2 & \textbf{74.4} & 23.7 & 18.2 \\

  \midrule

 \clip & ViT-L/14 \cite{radford2021clip} 
 & 81.7 & 0.0 & 0.0 & 65.6 & 0.1 & 0.1 \\
 \rclipf & ViT-L/14 \cite{schlarmann2024robust} 
 & 87.2 & 65.4 & 52.2  & 73.2 & 14.2 & 4.9 \\
\rclip  & ViT-L/14 \cite{schlarmann2024robust} 
& 89.1 & 74.9 & 72.1  & 74.0 & 21.5 & 12.1 \\
 \midrule
\clip 
& ViT-B/32 \cite{cherti2023openclip}& 
85.1 & 0.0 & 0.1 & 69.1 & 0.0 & 0.0 \\
 
 \rclipf & ViT-B/32 & 
 91.1 & 71.8 & 70.6 & 74.1 & 20.3 & 16.5 \\
 
 \rclip & ViT-B/32 & 
 91.0 & 79.1 & \textbf{79.7} & 74.1 & \textbf{29.2} & \textbf{27.0} \\
 \midrule
 \clip 
 & ViT-B/16 \cite{cherti2023openclip} 
 & 85.1 & 0.0 & 0.0 & 68.3 & 0.1 & 0.1 \\
 R-CLIP\textsubscript{F} & ViT-B/16 
 & 90.6 & 71.5 & 65.5 & 74.1 & 19.3 & 8.7 \\
 \rclip & ViT-B/16 & 
 91.9 & 79.4 & 77.1 & 74.0 & 27.8 & 19.6 \\
 \midrule
 \clip 
 & CnvNxt-B \cite{cherti2023openclip}
 & 87.2 & 0.0 & 0.0 & 68.2 & 0.0 & 0.0 \\
 R-CLIP\textsubscript{F} & CnvNxt-B & 
 90.6 & 74.3 & 66.1 & 74.0 & 19.0 & 6.0\\
 \rclip & CnvNxt-B & 
 \textbf{92.3} & \textbf{81.9} & 78.5 & 74.1 & 26.8 & 15.8 \\
 \midrule
 \addlinespace[2mm]

\multicolumn{8}{l}{\textbf{DINO and \clip fine-tuned with LoRA on NIGHTS}} \\
 \toprule

 DINO & ViT-B/16 \cite{fu2023learning} & 94.5 & 3.6 & 6.2 & 72.8 & 0.0 & 0.1 \\
 \rdinof & ViT-B/16 & 95.0 & 78.2 & 79.2 & 74.8 & 25.8 & \textbf{25.5} \\
 \midrule

 \clip & ViT-B/32 
 & 94.8 & 0.5 & 0.9 & 72.2 & 0.0 & 0.0  \\
 \rclipf & ViT-B/32  
 & 95.3 & 80.8 & 81.1 & \textbf{75.1} & 22.2 & 13.9 \\
 \rclip& ViT-B/32  
 & 94.3 & 80.8 & 82.2 & 74.2 & 21.6 & 23.1 \\
 \midrule
 \clip & ViT-B/16  
 & 94.5 & 0.0 & 0.0 & 71.3 & 0.0  & 0.0  \\
 \rclipf & ViT-B/16  
 & \textbf{95.7} & 80.9 & 78.6 & 74.5 & 25.2 & 16.7 \\
 \rclip & ViT-B/16  
 & 94.6 & 81.5 & 81.2 & 74.4 & 28.4 & 24.1 \\
 
 \midrule
 \clip & CnvNxt-B
 & 95.4 & 0.0 & 0.0 & 71.2 & 0.0 & 0.0 \\
 R-CLIP\textsubscript{F} & CnvNxt-B& 
 95.3 & 85.6 & 81.6 & 74.9 & \textbf{30.2} & 20.1 \\
  \rclip & CnvNxt-B& 
  95.0 & \textbf{87.2} & \textbf{84.5} & 74.7 & 29.2 & 20.0 \\

 \bottomrule
\end{tabular}
\end{table}

\begin{table*}[t]
\centering
\small
\caption{\textbf{Comparison to SOTA (robust) perceptual metrics on 2AFC tasks.}
We evaluate clean and robust performance on both NIGHTS and BAPPS datasets:
our zero-shot \rclipf and \rclip either outperform (zero-shot) 
or are close to (\lora fine-tuning) the baselines in clean accuracy, while achieving higher robust accuracy.
For NIGHTS all results are computed on the entire test set.
For BAPPS, clean performance is over the full test set, while robust accuracy is computed on 1k images for each split.
The adversarial perturbations are optimized with APGD at radii $\epsilon_\infty=4/255$ and $\epsilon_2=3$.
\textsuperscript{\dag} LipSim-Pretrained is distilled from DreamSim which in turn is fine-tuned on NIGHTS.
$^*$ indicates models not available and robustness could not be evaluated but expected to be similar to DreamSim (Ensemble + LoRA).
} 
\label{tab:comparison_baselines_2afc}
\tabcolsep=2pt
\extrarowheight=1.25pt
\newl=11mm
\newlc=11mm
\begin{tabular}{L{30mm} L{26mm} C{22mm} C{18mm} *{2}{|
>{\columncolor{lightgreen}}C{\newlc} C{\newl} C{\newlc}}}

& & & \multirow{2}{*}{\makecell{Fine-Tuning\\ Dataset}} & \multicolumn{3}{c}{NIGHTS} & \multicolumn{3}{c}{BAPPS} \\
Perceptual Model & Variant & Encoder &  & {clean} & {$\ell_\infty$} &{$\ell_2$} & {clean} & $\ell_\infty$ &{$\ell_2$}
\\

\toprule
\addlinespace[1.5mm]

\multirow{2}{*}{
Clean models
}
& DINO \cite{caron2021emerging} & ViT-B/16 & None & 90.2 & 1.1 & 1.3 & 71.5 & 0.1 & 0.1 \\
& CLIP \cite{cherti2023openclip} & ConvNeXt-B & None &87.2 & 0.0 & 0.0 & 68.2 & 0.0 & 0.0 \\
\midrule
\multirow{5}{*}{DreamSim~\cite{fu2023learning} }
& 
\clip \cite{cherti2023openclip} + \lora & ViT-B/32 & NIGHTS & 95.4 & 1.8 & 3.3 & 73.1 & 0.0 & 0.0 \\
& CLIP \cite{radford2021clip} + LoRA & ViT-B/32 & NIGHTS & 93.9 & 0.1 & 0.3 & 69.9 & 0.0 & 0.0\\
& DINO \cite{caron2021emerging} + LoRA & ViT-B/16 & NIGHTS & 94.5 & 3.6 & 6.2 & 72.8 & 0.0 & 0.1 \\
\cmidrule{2-10}
& Ensemble$^*$ & ViT-B/16 ($\times$3) & None & 90.8  & - & - & -& - & -\\
& Ensemble + LoRA & ViT-B/16 ($\times$3) & NIGHTS & \textbf{96.2} & 0.5 & 0.9 & 73.1 & 0.0 & 0.0\\
 \midrule
 \addlinespace[1.5mm]

\multirow{1}{*}{Robust LPIPS~\cite{ghazanfari2023rlpips}} & 
 R-LPIPS & AlexNet & BAPPS & 71.6 & 16.2 & 26.9& 72.8 & 7.0 & 12.3 \\
 \midrule
 \addlinespace[1.5mm]

\multirow{3}{*}{\lipsim~\cite{ghazanfari2023lipsim}} &
Pretrained & SLL & NIGHTS\textsuperscript{\dag}
 & 86.6 & 8.6 & 26.5 &74.2 & 1.1 & 7.4 \\
& $\textrm{Margin}_{0.2}$ & SLL & NIGHTS & 88.5 & 23.1 & 46.6 & 74.0 & 5.8 & 15.1 \\
& $\textrm{Margin}_{0.5}$ & SLL & NIGHTS & 85.1 & 32.8 & 53.1 & 73.1 & 7.0 & 12.3\\
\midrule
\addlinespace[1.5mm]

\multirow{6}{*}{
\makecell[l]{Robust \clip\\ and DINO (ours)}
} 
& \rdinof &  ViT-B/16 & None & 90.9 & 72.2 & 73.2 & 74.4 & 23.7 & 18.2 \\
& \rdinof + LoRA & ViT-B/16 & NIGHTS & 95.0 & 78.2 & 79.2 & 74.8 & 25.8 & \textbf{25.5} \\
\cmidrule{2-10}
&
\rclipf & ConvNeXt-B & None & 90.6 & 74.3 & 66.1 & 74.0 & 19.0 & 6.0 \\
& \rclipf + \lora & ConvNeXt-B & NIGHTS & 95.3 & 85.6 & 81.6 & \textbf{74.9} & \textbf{30.2} & 20.1 \\
\cmidrule{2-10}
& \rclip & ConvNeXt-B & None & 92.3 & 81.9 & 78.5 & 74.1 & 26.8 & 15.8 \\
& \rclip + \lora & ConvNeXt-B &  NIGHTS  & 95.0 & \textbf{87.2} & \textbf{84.5} & 74.7 & 29.2 & 20.0 \\

\bottomrule
 
\end{tabular}
\end{table*}

\textbf{Zero-shot perceptual metrics.}
The top part of Table~\ref{tab:comparison_clip_models}
reports the clean and robust zero-shot accuracy of the various perceptual metrics on the test set of NIGHTS and BAPPS. 
The robust \clip models achieve significantly higher clean accuracy than their original clean \clip counterparts, from which they have been fine-tuned.
The improvements are consistent across all encoder architectures, adversarial fine-tuning schemes (\ours, \tecoa), as well as datasets, in the range of 3-7\% of clean performance.
The same is true for \rev{the} robust versions of DINO and DINOv2 models, i.e. \rdinof and \rdinoviif (the results for DINOv2 models are reported in \cref{tab:additional_models_nights} in Appendix).
This is remarkable as adversarial robustness is typically associated with a loss in performance \cite{Tsipras2019RobustnessOddsAcc, croce2020RobustBench},
\rev{as also observed for \tecoa and \ours on non-perceptual similarity tasks in \cite{Mao2022UnderstandingZAtecoa, schlarmann2024robust}.}
We hypothesize that the robustness to imperceptible $\ell_\infty$-perturbation leads to an emphasis of robust features, which are likely more correlated with higher order semantic concepts.
Non-robust features, which are expected not to encode semantic information, are instead suppressed by adversarial training.
\rev{
Evidence that robust features are more semantically meaningful is given by the fact that robust models, and in particular classifiers, have more interpretable gradients and generative properties than standard ones \cite{santurkar2019imagesynthesis, augustin2020ratio, boreiko2022sparsevisual}. It has also been shown that robust models’ decisions are biased more by shape than by texture of an image \cite{zhang2019interpreting}, while the opposite is true for non-robust models \cite{geirhos2018imagenet}. Notably, also humans have a shape bias \cite{geirhos2018imagenet}.
}

As expected, the similarity metrics induced by clean models are not adversarially robust (their robust accuracy is consistently near zero).
Conversely, using the robust embedding of \rdinof, \rclipf and \rclip yields robust perceptual metrics in both $\ell_\infty$- and $\ell_2$-threat models.
We observe that the supervised adversarial fine-tuning of \tecoa gives higher robustness in all cases, and typically better clean accuracy, than \ours on 2AFC-tasks, but FARE fine-tuned models generalize better to other tasks, see Section \ref{sec:image_retrieval}.
Overall, these experiments show that in this case the clean vs robust accuracy trade-off which has been observed in several tasks, see e.g. \cite{Tsipras2019RobustnessOddsAcc}, is even reversed, and adversarial training is beneficial for both clean \emph{and} robust performance.
\\

\textbf{Comparison of backbones for (robust) zero-shot perceptual metrics.}
While \rev{Fu et al.}~\cite{fu2023learning} have analyzed 
\rev{different pre-trained models}
(\clip, DINO and MAE), they all share ViTs as backbone for the vision encoders.
However, \convnext, an architecture built on convolution, has been shown to perform on par with vision transformers (and sometimes better) on several vision tasks \cite{liu2022convnet, Woo2023ConvNeXtV2, singh2023revisiting}. 
Interestingly, \cref{tab:comparison_clip_models} illustrates that in our setup the \rclip ConvNeXt-B model achieves higher zero-shot clean and robust accuracy in the $\ell_\infty$-threat model than the two vision transformers of similar size (ViT-B/16, ViT-B/32) on NIGHTS, and same clean performance but slightly worse robust accuracy on BAPPS.
Also, the \rclipf with ConvNeXt-B model is the most robust one among the FARE models on NIGHTS, and comparable in clean accuracy on NIGHTS and BAPPS to the ViTs but slightly less robust on BAPPS. 
In line with \cite{fu2023learning}, we observe that the larger robust ViT-L/14 models of \cite{schlarmann2024robust} perform worse than our smaller ViT-B networks. 
Finally, \rdinof generally achieves comparable or better results than  \rclipf with both the same and other encoder types.

Despite the zero-shot FARE models perform a bit worse than \tecoa, we will see in \cref{sec:image_retrieval} that the robust FARE models (especially CLIP) generalize much better to other tasks, which is plausible given that FARE preserves the original embedding while being robust.
\\

\textbf{Fine-tuning on the NIGHTS dataset.}
Next, 
we fine-tune the encoders 
following the setup of \cite{fu2023learning}, i.e. with clean training, 
using LoRA \cite{hu2022lora} to update the entire network (see Appendix~\ref{app:experiments} for details and results with MLP probing on top of the frozen encoders). 

When fine-tuning with LoRA on NIGHTS, the robust models still perform better than standard ones on BAPPS, while clean performance on NIGHTS is quite similar with a small but consistent improvement for the FARE-models.
Regarding robustness, the \rclip models are slightly better but the gap to \rclipf is much smaller than in the zero-shot setup.
The fine-tuned robust DINO model \rdinof attains, on average, slightly worse robust accuracy than \rclipf and \rclip.
Notably, \lora-fine-tuning the adversarially trained backbones allows the similarity metric to retain, and even improve, robustness in the $\ell_p$-threat models.
While this might be unexpected, we speculate that the fine-tuning 
allows the models to rely on a subset of a small number of task-specific features for classification, since the NIGHTS benchmark is of limited difficulty.
This means that 
thanks to the robust pre-training the relevant features are highly robust, and the further fine-tuning down-weights the importance of the non-robust features, thus leading to the improvement in robustness.

Finally, fine-tuning on NIGHTS even brings slight benefits to the performance on BAPPS, as noticed by \cite{fu2023learning}.
In total we see that the \convnext-B architecture performs best regarding robustness and thus we fix this as our architecture for the clean and robust CLIP models for the remainder of the paper, but show results for all robust CLIP models in the Appendix.

\subsection{Comparison to SOTA (robust) perceptual metrics} \label{sec:comparison}

Next, we compare \rclipf,  \rclip (with \convnext-B as backbone since it gives mostly better results than the ViT-Bs, see discussion above) and \rdinof to SOTA methods for clean and robust perceptual metrics for 2AFC tasks, and summarize the results in \cref{tab:comparison_baselines_2afc}.
Additional results (including a breakdown over dataset splits) can be found in Appendix~\ref{app:additional_experiments}.

The DreamSim-Ensemble \cite{fu2023learning} concatenates the features of three ViTs (the original \clip \cite{radford2021clip}, DINO \cite{caron2021emerging}, and a CLIP model from OpenCLIP \cite{cherti2023openclip}) to obtain the features for computing perceptual similarity:
this achieved SOTA results on NIGHTS both in the zero-shot setting and with LoRA-fine-tuning, although at increased inference cost (since three encoders are needed).
Additionally, as a lightweight alternative, \rev{Fu et al.}~\cite{fu2023learning} provide single-branch models (OpenCLIP, CLIP, DINO) fine-tuned with \lora on NIGHTS.
\rclip, with a single encoder, outperforms the DreamSim-Ensemble in both the zero-shot setup and when fine-tuning a task-specific MLP head (see Appendix~\ref{app:bapps}), while being worse only for LoRA fine-tuning.
Both \rclipf and \rclip are comparable or better than the single-encoder DreamSim models (all use \lora fine-tuning) on NIGHTS,
and perform similar or slightly better than the LipSim models \cite{ghazanfari2023lipsim} on BAPPS.

In the context of robust perceptual metrics, \lipsim~\cite{ghazanfari2023lipsim} 
Pretrained model attains certified $\ell_2$-robustness by distilling DreamSim on \imnet, while the $\textrm{Margin}_{0.2}$ and $\textrm{Margin}_{0.5}$ models are further fine-tuned on NIGHTS.
The main goal of \lipsim is certified $\ell_2$-robustness, but \rev{Ghazanfari et al.}~\cite{ghazanfari2023lipsim} also report good performance in empirical robustness.
Moreover, \rev{Ghazanfari et al.}~\cite{ghazanfari2023rlpips} propose a robust version of LPIPS trained on the BAPPS dataset.
All robust zero-shot \clip and DINO models outperform LipSim and R-LPIPS both in clean and robust accuracy by large margins: for example, on NIGHTS, 
\rclip achieves 49.1\% and 25.4\% higher robust accuracy in $\ell_\infty$ and $\ell_2$ respectively than the most robust baseline \lipsim-$\textrm{Margin}_{0.5}$, while having even 7.2\% better clean performance.
Similarly, on BAPPS 
\rev{the robust CLIP and DINO encoders} achieve higher robust accuracy than the baselines.
Finally, DreamSim as well as the clean CLIP and DINO do not have any non-trivial robustness.

\subsection{\rev{Additional robustness evaluations}}
\label{sec:other_evalutions}

\rev{In Appendix~\ref{app:bapps} we perform sanity checks of the robustness evaluation for our metrics.
First, our attack yields zero robustness for sufficiently large perturbations to rule out potential problems in the optimization (\cref{fig:curves}).
Second, increasing the attack iterations does not significantly reduce the robustness (\cref{fig:attack-iters}).
Then, to exclude gradient-masking issues, we verify that the black-box Square Attack \cite{AndriushchenkoC20Square} does not improve the evaluation of the white-box attacks.
Finally, we report in \cref{tab:zero-shot-imnet} the clean and robust accuracy of the CLIP models on both ImageNet and 13 zero-shot classification datasets (we follow the evaluation protocol of \cite{schlarmann2024robust}, see Appendix~\ref{app:other_tasks}): as expected, the \rclipf and \rclip models demonstrate robustness but at the expense of clean accuracy.}

\definecolor{newgray}{rgb}{.6, .6, .6}

\begin{table*}[t]
\centering\small
\caption{\textbf{Robust NSFW detection.} We consider two scenarios: \textit{(i)} the query images are safe (class $\mathcal{S}$) and the attacker target is the unsafe class $\mathcal{U}$, and \textit{(ii)} the opposite, i.e. when the query images are from $\mathcal{U}$ and the target is the safe class $\mathcal{S}$ (in practice, the most relevant attack direction).
We report 
the fraction of points classified into each of the three classes, including the buffer class $\mathcal{B}$ (results for incorrect classes in grey) with and without 
adversarial attack.
\rclipf achieves clean accuracy similar to \clip and the DreamSim-Ensemble, with a significantly higher robust accuracy against $\ell_\infty$-bounded attacks.
}
\tabcolsep=1.1pt
\extrarowheight=2pt
\newl=8mm
\newlc=8mm
\begin{tabular}{L{25mm} L{25mm} C{20mm} |
>{\columncolor{lightgreen}}C{\newlc}
>{\columncolor{lightgreen}}C{\newlc}
>{\columncolor{lightgreen}}C{\newlc}
C{\newlc}C{\newlc}C{\newlc}|
>{\columncolor{lightgreen}}C{\newlc}
>{\columncolor{lightgreen}}C{\newlc}
>{\columncolor{lightgreen}}C{\newlc}
C{\newlc}C{\newlc}C{\newlc}}
& & & \multicolumn{6}{c}{Query: safe ($\mathcal{S})$ $\rightarrow$ Target: unsafe ($\mathcal{U}$)} & \multicolumn{6}{|c}{Query: unsafe ($\mathcal{U})$ $\rightarrow$ Target: safe ($\mathcal{S}$)}\\
 \cline{4-9} \cline{10-15}
\addlinespace[0.5mm]
\multirow{2}{*}{Perceptual Model} &\multirow{2}{*}{Variant} & \multirow{2}{*}{Encoder} & \multicolumn{3}{c}{clean} & \multicolumn{3}{c|}{$\ell_\infty (\nicefrac{8}{255})$}  & \multicolumn{3}{c}{clean} & \multicolumn{3}{c}{$\ell_\infty (\nicefrac{8}{255})$} \\
& &  & $\mathcal{S}\uparrow$& \textcolor{newgray}{$\mathcal{B}\downarrow$}& \textcolor{newgray}{$\mathcal{U}\downarrow$}& $\mathcal{S}\uparrow$ & \textcolor{newgray}{$\mathcal{B}\downarrow$} &  \textcolor{newgray}{$\mathcal{U}\downarrow$} & \textcolor{newgray}{$\mathcal{S}\downarrow$} &  \textcolor{newgray}{$\mathcal{B}\downarrow$} & $\mathcal{U}\uparrow$ &  \textcolor{newgray}{$\mathcal{S}\downarrow$} &  \textcolor{newgray}{$\mathcal{B}\downarrow$} & $\mathcal{U}\uparrow$\\
\toprule
\addlinespace[1.5mm]

\multirow{2}{*}{\makecell[l]{Clean models}}
 & DINO \cite{caron2021emerging} & \vit-B/16  & 79.0 &\textcolor{newgray}{ 10.8 }&\textcolor{newgray}{ 10.2 }& 4.8 &\textcolor{newgray}{ 12.7 }&\textcolor{newgray}{ 82.5 }&\textcolor{newgray}{ 2.8 }&\textcolor{newgray}{ 10.4 }& 86.8 &\textcolor{newgray}{ 77.1 }&\textcolor{newgray}{ 10.2 }& 12.7 \\
 & \clip \cite{cherti2023openclip} & \convnext-B & \textbf{89.4} &\textcolor{newgray}{ 6.6 }&\textcolor{newgray}{ 4.0 }& 1.0 &\textcolor{newgray}{ 9.8 }&\textcolor{newgray}{ 89.2 }&\textcolor{newgray}{  0.2 }&\textcolor{newgray}{ 8.2 }&91.6 &\textcolor{newgray}{ 89.0 }&\textcolor{newgray}{ 6.8 }& 4.2 \\
 \midrule
\addlinespace[1.5mm]

\multirow{2}{*}{DreamSim \cite{fu2023learning}}
 & DINO \cite{caron2021emerging} + \lora &  \vit-B/16  & 72.2 &\textcolor{newgray}{ 14.0 }&\textcolor{newgray}{13.8 }& 5.0 &\textcolor{newgray}{ 11.4 }&\textcolor{newgray}{ 83.6 }&\textcolor{newgray}{ 0.6 }&\textcolor{newgray}{6.6 }& 92.8 &\textcolor{newgray}{ 63.8 }&\textcolor{newgray}{ 16.6 }& 19.6 \\
 & Ensemble + \lora &  ViT-B/16 ($\times$3) & 88.6 &\textcolor{newgray}{ 7.8 }&\textcolor{newgray}{3.6 }&0.8 &\textcolor{newgray}{ 3.2 }&\textcolor{newgray}{ 96.0 }&\textcolor{newgray}{0.2 }&\textcolor{newgray}{4.2 }& 	\textbf{95.6} &\textcolor{newgray}{ 84.0 }&\textcolor{newgray}{ 10.6 }& 5.4 \\
\midrule
\addlinespace[1.5mm]

\multirow{1}{*}{Robust LPIPS \cite{ghazanfari2023rlpips}}
 & R-LPIPS & AlexNet & 40.2 &\textcolor{newgray}{ 25.6 }&\textcolor{newgray}{ 34.2 }& 7.0 &\textcolor{newgray}{ 38.0 }&\textcolor{newgray}{ 55.0 }&\textcolor{newgray}{ 10.2 }&\textcolor{newgray}{ 27.2 }& 62.6 &\textcolor{newgray}{ 41.2 }&\textcolor{newgray}{ 23.6 }& 35.2 \\
\midrule
\addlinespace[1.5mm]

\multirow{2}{*}{\lipsim \cite{ghazanfari2023lipsim}}
 & Pretrained & SLL & 66.4 &\textcolor{newgray}{ 20.8 }&\textcolor{newgray}{ 12.8 }& 9.6 &\textcolor{newgray}{ 32.0}&\textcolor{newgray}{ 58.4 }&\textcolor{newgray}{5.6 }&\textcolor{newgray}{ 32.8 }& 61.6 &\textcolor{newgray}{ 61.6 }&\textcolor{newgray}{ 12.6 }& 25.8 \\
 & Margin$_{0.5}$ & SLL & 69.2 &\textcolor{newgray}{ 16.0 }&\textcolor{newgray}{14.8 }& 35.0 &\textcolor{newgray}{ 15.2}&\textcolor{newgray}{ 49.8 }&\textcolor{newgray}{ 21.6 }&\textcolor{newgray}{ 12.6}& 65.8&\textcolor{newgray}{ 50.2 }&\textcolor{newgray}{ 23.6 }& 26.2 \\
 \midrule
 \addlinespace[1.5mm]

\multirow{6}{*}{\makecell[l]{Robust CLIP and\\ DINO (ours)}}
 & \rdinof &  \vit-B/16  & 80.2 &\textcolor{newgray}{ 8.8 }&\textcolor{newgray}{ 11.0 }& 54.6 &\textcolor{newgray}{ 8.7 }&\textcolor{newgray}{ 36.7 }&\textcolor{newgray}{ 12.0 }&\textcolor{newgray}{ 3.6 }& 84.4 &\textcolor{newgray}{ 26.9 }&\textcolor{newgray}{ 6.0 }& 67.1 \\
 & \rdinof + \lora &  \vit-B/16  & 69.0 & \textcolor{newgray}{21.4} &\textcolor{newgray}{9.6} & \textbf{58.6} &\textcolor{newgray}{20.4} & \textcolor{newgray}{21.0} & \textcolor{newgray}{6.0} & \textcolor{newgray}{13.6} & 80.4 & \textcolor{newgray}{7.0} & \textcolor{newgray}{20.6} & 72.4 \\
 \cmidrule{2-15}
 & \rclipf & \convnext-B & 88.6 &\textcolor{newgray}{4.2 }&\textcolor{newgray}{ 7.2 }& 50.6 &\textcolor{newgray}{ 15.2 }&\textcolor{newgray}{ 34.2 }&\textcolor{newgray}{  1.2 }&\textcolor{newgray}{6.6 }& 92.2 &\textcolor{newgray}{ 18.6 }&\textcolor{newgray}{ 6.4 }& 75.0 \\
 & \rclipf + \lora & \convnext-B & 78.4 & \textcolor{newgray}{ 10.8 } & \textcolor{newgray}{ 10.8 } & 57.1 & \textcolor{newgray}{ 21.7 } & \textcolor{newgray}{ 21.2 } &  \textcolor{newgray}{ 1.4 } & \textcolor{newgray}{ 13.6 } & 85.0 & \textcolor{newgray}{ 9.0 } & \textcolor{newgray}{ 12.2 } & \textbf{78.8}\\
 \cmidrule{2-15}
 & \rclip & \convnext-B & 74.2 &\textcolor{newgray}{ 9.0 }&\textcolor{newgray}{ 16.8 }& 46.6 &\textcolor{newgray}{ 21.0 }&\textcolor{newgray}{ 32.4 }&\textcolor{newgray}{  9.8 }&\textcolor{newgray}{ 11.2 }& 79.0 &\textcolor{newgray}{ 8.2 }&\textcolor{newgray}{ 21.4 }& 70.4 \\
 & \rclip + \lora & \convnext-B & 80.2 & \textcolor{newgray}{ 11.0 } & \textcolor{newgray}{ 8.8 } & 53.1 & \textcolor{newgray}{ 25.0 } & \textcolor{newgray}{ 21.9 }&  \textcolor{newgray}{ 2.0 }& \textcolor{newgray}{ 18.0 }& 80.0 & \textcolor{newgray}{ 4.8 }& \textcolor{newgray}{ 29.6 } & 65.6\\

\bottomrule
\label{tab:robust-retrieval-short}
\end{tabular}
\end{table*}

\section{Robust Image-to-Image Retrieval}
\label{sec:image_retrieval}

Besides 2AFC tasks which quantify alignment with human perception, probably one of the most popular application scenarios of perceptual metrics is image-to-image retrieval. In the following we illustrate that the strong performance (both clean and robust) of \rdinof, \rclipf and \rclip (without and with fine-tuning on NIGHTS) generalizes to this task in different test-cases, when compared to the SOTA baselines introduced in the previous section. 
In particular, we illustrate in \cref{sec:nsfw_detection} and Appendix~\ref{sec:dataset_filtering} (in synthetic scenarios, as a proof of concept) that robust perceptual metrics might be especially useful in sensitive tasks like NSFW detection, where malicious actors have a clear incentive to circumvent the detector:
as we show in the following, these could introduce adversarial perturbations in their images which are visually not noticeable but bypass the detection mechanisms.

\subsection{Robust NSFW detection} \label{sec:nsfw_detection}

An approach to detection leverages image-to-image retrieval: given a set of labeled images representative of different classes, one can use a (perceptual) metric to find the nearest neighbor of a query image $\vx$ in the retrieval pool, i.e. that with highest perceptual similarity.
Then, the class of the nearest neighbor is assigned to the query image, and potentially detected. 
\\

\textbf{Setup.} To test the different perceptual metrics on this task, we sample 500 images each from a public dataset\footnote{\url{https://huggingface.co/datasets/deepghs/nsfw_detect}} 
for the classes `neutral' 
(which constitutes the safe class denoted $\mathcal{S}$)
`p*rn' 
(the unsafe class $\mathcal{U}$) and `s*xy' 
(as a buffer class $\mathcal{B}$ which can be used to indicate images on which the model is uncertain)
as retrieval pools.
Then, we select test sets of 1000 images for the $\mathcal{S}$ and $\mathcal{U}$ classes, disjoint of the retrieval sets.
As classification rule, we compute the cosine similarity for each query image to all 1500 retrieval images (3 classes), and select the class of the image with maximal similarity.
For computing the adversarial attacks, we sample a set $Y$ of 16 images belonging to the target class (but not included in the retrieval set) and minimize the average distance of their normalized embeddings to that of query image $\vx$.
This yields the optimization problem
\begin{align} 
\min_{\norm{\vdelta}_p \leq \epsilon_p}\, \sum_{\vy \in Y} \norm{\frac{\phi(\vx + \vdelta)}{\norm{\phi(\vx + \vdelta)}_2} - \frac{\phi(\vy)}{\norm{\phi(\vy)}_2}}_2^2 \;\;
\textrm{s. th.} \;\; \vx + \vdelta \in I, 
\label{eq:image-retrieval-nsfw}
\end{align}
which is optimized with 200 iterations of APGD at $\ell_\infty$-radius of $\epsilon = 8/255$.
We notice that \cref{eq:image-retrieval-nsfw} corresponds to maximizing the sum of the cosine similarities $\simfn(\vx + \vdelta, \vy)$ (see~\cref{eq:cossim}) to the images $\vy$ from the target class, and thus is  a targeted attack.
While we use the gradient from the perceptual model, we do not assume access to the retrieval set, which can be interpreted as a grey-box scenario of an attacker with only partial knowledge of the target system.
We consider this as a realistic scenario, as 
publicly available foundation models like CLIP or DINO are nowadays frequently used for image retrieval but the retrieval set would be unknown to the attacker.
\\

\textbf{Results.}
\cref{tab:robust-retrieval-short} shows the clean and robust performance of the perceptual metrics on images of both the safe ($\mathcal{S}$) and unsafe ($\mathcal{U}$) classes (for each type of query, we report the breakdown of the classification into each class to account for the effect of the buffer class $\mathcal{B}$).
\rclipf performs on par with clean \clip (as well \rdinof with DINO) and only marginally below the DreamSim-Ensemble.
We observe that in this task \rclipf outperforms \rclip, likely because the unsupervised fine-tuning of \ours better preserves the original features on a larger set of the input image space compared to \tecoa.
Moreover, fine-tuning
\rdinof and \rclipf with \lora on NIGHTS notably degrades the performance, showing that these perceptual metrics, unlike the zero-shot ones, may overfit to the 2AFC tasks.
Finally, R-LPIPS and LipSim attain significantly lower clean accuracy than the other methods, possibly due to their simpler architecture.

Then, we compute robust accuracy for both safe and unsafe queries: on images of $\mathcal{U}$ perturbed via attacks with target $\mathcal{S}$, \rclipf retains 75.0\% accuracy, significantly higher than both the clean baselines, which are very vulnerable to these attacks (e.g. \clip and the DreamSim ensemble have around 5\% accuracy), and the robust R-LPIPS (35.2\%) and LipSim (26.2\%) metrics.
This case, equivalent to making unsafe images not being detected, is the most concerning goal of a malicious actor, and \rev{the results of} \rclipf and \rdinof suggest that it is viable to achieve robustness while preserving clean detection performance.
Additional results can be found in \cref{tab:robust-retrieval} in Appendix~\ref{sec:image-retrieval-nsfw}.

\textbf{The LAION NSFW detector.} 
LAION-5B is one of the largest and most popular image-text pairs datasets \cite{schuhmann2022laion5b}.
After links to child sex abuse material (CSAM) were found in LAION-5B, it went through a revision 
to filter out links to CSAM data based on hashes of known CSAM. 
Additionally, they used the LAION NSFW detector \cite{laionNSFW2022}: this consists of a shallow head on top of the embeddings of a CLIP ViT-L/14 model, and is trained as a binary NSFW classifier (safe vs unsafe) on a training set of 240k samples. With this, they could filter out 176M images (about 3\% of LAION-5B) and release 
\textit{re-LAION-5B-research-safe}.

\begin{table}[t]
\centering\small
\caption{\textbf{Attack on the  LAION NSFW detector (safe ($\mathcal{S}$) vs unsafe ($\mathcal{U}$))}. We compute the perturbations against the LAION detector with the grey-box attack from \cref{eq:image-retrieval-nsfw} without knowledge about its additional classifier (white-box robustness is zero).
For the unsafe class the LAION NSFW detector works very well but is not robust, whereas \rclipf reaches similar clean accuracy but is significantly robust. Combining \rclipf and the LAION detector yields high clean and robust accuracy for the unsafe class. 
}
\tabcolsep=2pt
\extrarowheight=2pt
\newl=9mm
\newlc=9mm
\begin{tabular}{L{18mm} C{23mm} *{2}{|
C{\newlc}>{\columncolor{lightgreen}} C{\newl}}}
  & & \multicolumn{2}{c}{Clean} & \multicolumn{2}{c}{$\ell_\infty(\nicefrac{8}{255})$}\\
 \cline{3-6}
\addlinespace[0.5mm]
Method & Encoder &$\mathcal{S}$ & $\mathcal{U}$ & $\mathcal{S}$ & $\mathcal{U}$ \\
\toprule
\addlinespace[0.5mm]
LAION-det & \vit-L/14 & 98.2 & 99.6 & 13.1 & 28.2 \\
\midrule
{\clip}
& \convnext-B & 95.4 & 99.4 & 0.0 & 4.2 \\
{\rclipf}
& \convnext-B &92.2 & 98.4 & 56.4 & 82.8\\
 \midrule

\makecell[l]{LAION-det +\\ \rclipf} & \makecell{ViT-L/14 +\\ \convnext-B} & 89.6 & 99.8  &9.3 & 91.0\\
\bottomrule
\label{tab:laion-nsfw}
\end{tabular}
\end{table}

\begin{figure*}[t] \centering 
 \includegraphics[align=c, width=1.8\columnwidth]{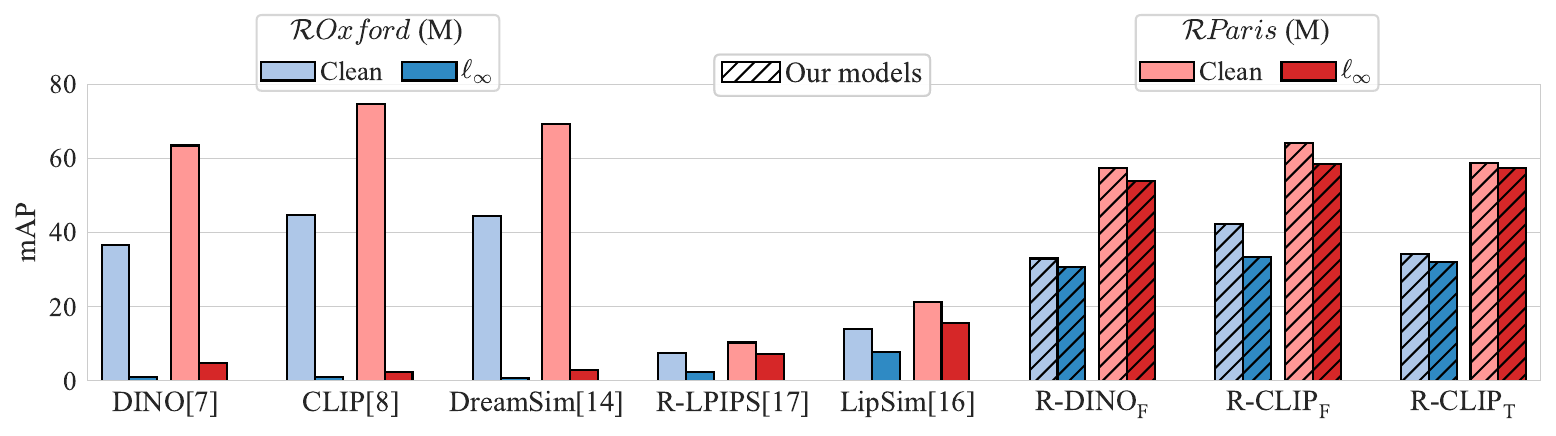}
 \\
 \caption{\textbf{Nearest neighbors retrieval on \oxford and \paris.}
 We report clean and robust mAP of different methods on the Medium (M) sets of \oxford (blue hue) and \paris (red hue): our \rclipf (zero-shot with \convnext backbone) achieves clean performance not far from the clean models, while having significantly higher robustness ($\epsilon_\infty=4/255$).}
 \label{fig:nn-retrieval}
\end{figure*}

An adversarial attack on the LAION NSFW detector could allow malicious actors to manipulate NSFW images to be considered safe and included in future versions of the dataset.
We attack the LAION NSFW detector with our grey-box attack from \cref{eq:image-retrieval-nsfw}, so that results are comparable to those 
of the detection performed with the perceptual metrics via retrieval.
This means the attacker does not have knowledge of the classifier on top of the CLIP embedding, but only access to the CLIP model
(a full white-box attack on the LAION NSFW detector yields 0\% accuracy for the unsafe class, see Appendix~\ref{sec:image-retrieval-nsfw}).
In \cref{tab:laion-nsfw}, we compare the LAION detector to our simple nearest neighbor based detector with the original CLIP 
and \rclipf, which are based on a much smaller retrieval set (less than 1\%) than the training data of the LAION detector.
As the LAION NSFW detector only distinguishes between safe and unsafe, we recompute the performance of CLIP and \rclipf on this restricted set, 
i.e. without buffer class.
Notably, both \clip and \rclipf, while being worse on the safe data, perform very well on the unsafe part, but only \rclipf shows strong robustness. 
As the main safety concern is to filter out the unsafe data, these results are quite strong given the small retrieval pool, which illustrate 
the potential usefulness of our \rclipf model for NSFW detection. 

Finally, as a simple fix to the missing robustness of the LAION NSFW detector, we combine their detector with our \rclipf detector by rejecting an input as unsafe if one of them classifies it as unsafe.
In \cref{tab:laion-nsfw}, we report robustness of the combined detector by considering the attacks on both models \emph{independently} and then taking the `or'-operation which yields a lower bound on the robustness. 
Combining the two detectors retains or improves clean and robust accuracy for the unsafe class at the price of more errors for the safe class. As the main goal is to detect unsafe images, the loss in more safe images is likely acceptable given the potential harm if unsafe images remain in such an important dataset for the community. While this experiment is more of a proof of concept, we think that robust NSFW detection deserves more attention.

\begin{figure}[t]
\centering
\tabcolsep=3pt

\includegraphics[align=c, width=1.\columnwidth]{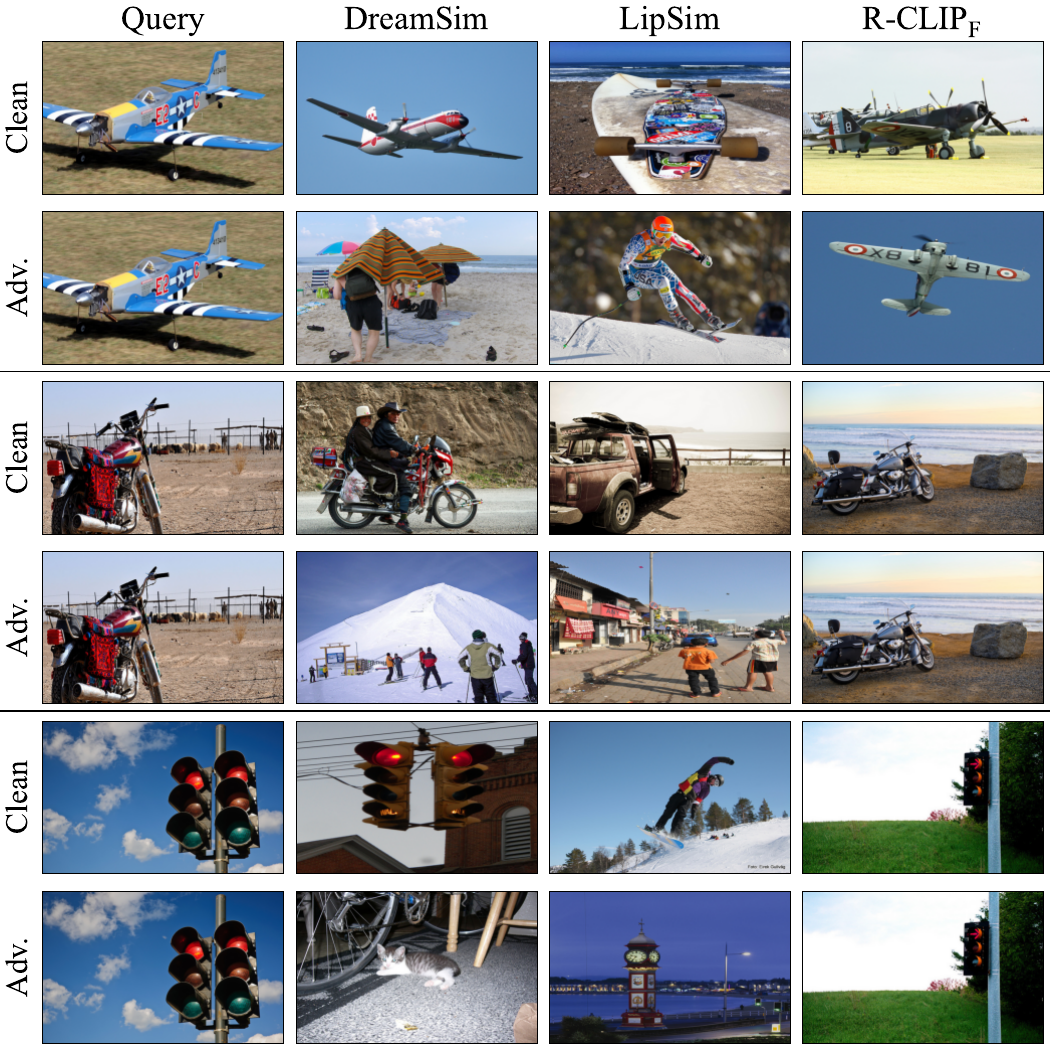}

\caption{\textbf{Qualitative analysis on MS-COCO.} We show the nearest neighbors retrieved for random query images from MS-COCO (first column) by the DreamSim-Ensemble, LipSim and \rclipf (\convnext), before (``Clean'' row) and after (``Adv.'' row) adding the adversarial perturbation to the query image ($\epsilon_\infty=4/255$). For clean images \rclipf and DreamSim have both semantically correct nearest neighbors whereas LipSim is off in some cases. Only \rclipf maintains semantically correct nearest neighbors under adversarial perturbations. 
}
\label{fig:coco-short}
\end{figure}

\subsection{Nearest neighbors retrieval} \label{sec:nn_retrieval}
Another instance of image-to-image retrieval is finding in the retrieval pool images which are similar to or share the same subject with the query image.
For example, in the revisited Oxford 
and Paris 
datasets~\cite{radenovic2018revisiting} the task is to retrieve different images of a given landmark.
In this case an attacker might want to modify the query image so that similar images are not found, for example to bypass image attribution methods \cite{andriushchenko2022aria}. 
This amounts to, in contrast to the one in the previous section, an untargeted attack with the goal of distorting the embedding of the perturbed image as much as possible compared to the original image.
Thus, to test the adversarial robustness of a metric in this task, we optimize a perturbation $\vdelta$ to maximize the squared $\ell_2$ distance between the (normalized) embedding of $\vx$ and $\vx + \vdelta$, i.e. 
\begin{align}
\max_{\norm{\vdelta}_p \leq \epsilon_p}\, \norm{\frac{\phi(\vx + \vdelta)}{\norm{\phi(\vx + \vdelta)}_2} - \frac{\phi(\vx)}{\norm{\phi(\vx)}_2}}_2^2 \quad
\textrm{s. th.} \quad \vx + \vdelta \in I, 
\label{eq:image-retrieval}
\end{align}
which is equivalent to minimizing the cosine similarity $\simfn(\vx + \vdelta, \vx)$, see~\cref{eq:cossim} (we use $\ell_\infty$-bounded attacks with $\epsilon_\infty=4/255$, and 100 iterations of APGD).
We note that 
even in this case the attacker does not require access to the retrieval set. 
Further details are provided in Appendix~\ref{sec:details_image_retrieval}.

{\figwidth=1.8\columnwidth
\begin{figure*}[t]
\centering
\footnotesize
    \begin{minipage}[t]{.025\textwidth}
         \vspace{11mm}
         \rotatebox{90}{\textbf{Original}}
    \end{minipage}%
    \begin{minipage}[t]{.98\textwidth}
        \vspace{0pt}
            \begin{subfigure}[t]{0.16\textwidth}
                    \includegraphics[width=\textwidth]{ 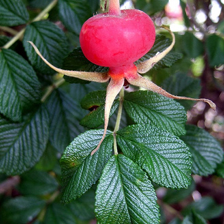 }
            \end{subfigure}%
                \hspace{0.5mm}%
            \begin{subfigure}[t]{0.16\textwidth}
                    \includegraphics[width=\textwidth]{ 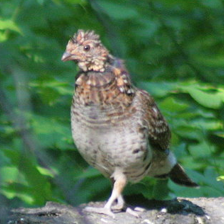 }
            \end{subfigure}%
                \hspace{0.5mm}%
            \begin{subfigure}[t]{0.16\textwidth}
                    \includegraphics[width=\textwidth]{ 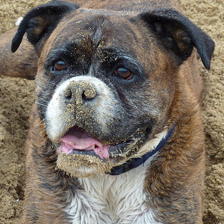 }
            \end{subfigure}%
                \hspace{0.5mm}%
            \begin{subfigure}[t]{0.16\textwidth}
                    \includegraphics[width=\textwidth]{ 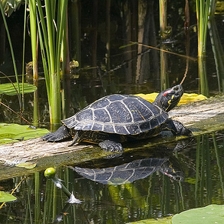 }
            \end{subfigure}%
                \hspace{0.5mm}%
            \begin{subfigure}[t]{0.16\textwidth}
                    \includegraphics[width=\textwidth]{ 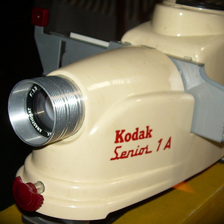 }
            \end{subfigure}%
                \hspace{0.5mm}%
            \begin{subfigure}[t]{0.16\textwidth}
                    \includegraphics[width=\textwidth]{ 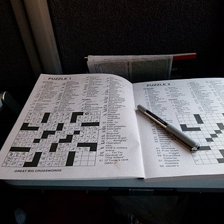 }
            \end{subfigure}%
    \end{minipage}

    \vspace{1mm}
    \begin{minipage}[t]{.025\textwidth}
         \vspace{11mm}
         \rotatebox{90}{\textbf{CLIP}}
    \end{minipage}%
    \begin{minipage}[t]{.98\textwidth}
        \vspace{0pt}
            \begin{subfigure}[t]{0.16\textwidth}
                    \includegraphics[width=\textwidth]{ 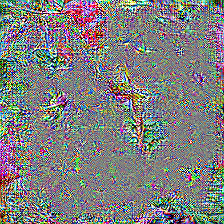 }
            \end{subfigure}%
                \hspace{0.5mm}%
            \begin{subfigure}[t]{0.16\textwidth}
                    \includegraphics[width=\textwidth]{ 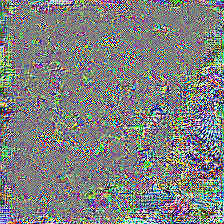 }
            \end{subfigure}%
                \hspace{0.5mm}%
            \begin{subfigure}[t]{0.16\textwidth}
                    \includegraphics[width=\textwidth]{ 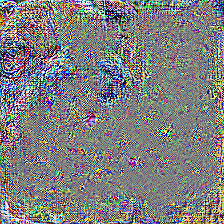 }
            \end{subfigure}%
                \hspace{0.5mm}%
            \begin{subfigure}[t]{0.16\textwidth}
                    \includegraphics[width=\textwidth]{ 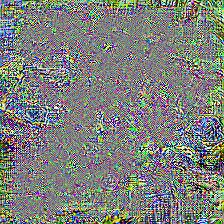 }
            \end{subfigure}%
                \hspace{0.5mm}%
            \begin{subfigure}[t]{0.16\textwidth}
                    \includegraphics[width=\textwidth]{ 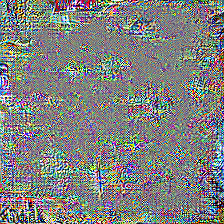 }
            \end{subfigure}%
                \hspace{0.5mm}%
            \begin{subfigure}[t]{0.16\textwidth}
                    \includegraphics[width=\textwidth]{ 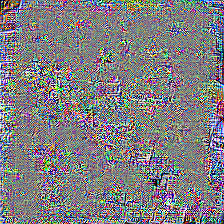 }
            \end{subfigure}%
    \end{minipage}

    \vspace{1mm}
    \begin{minipage}[t]{.025\textwidth}
         \vspace{9mm}
         \rotatebox{90}{\textbf{\rdinof}}
    \end{minipage}%
    \begin{minipage}[t]{.98\textwidth}
        \vspace{0pt}
            \begin{subfigure}[t]{0.16\textwidth}
                    \includegraphics[width=\textwidth]{ 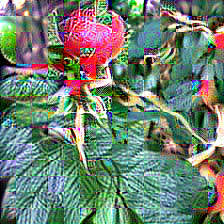 }
            \end{subfigure}%
                \hspace{0.5mm}%
            \begin{subfigure}[t]{0.16\textwidth}
                    \includegraphics[width=\textwidth]{ 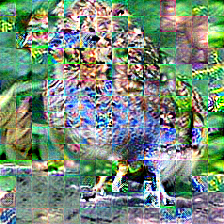 }
            \end{subfigure}%
                \hspace{0.5mm}%
            \begin{subfigure}[t]{0.16\textwidth}
                    \includegraphics[width=\textwidth]{ 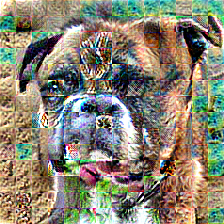 }
            \end{subfigure}%
                \hspace{0.5mm}%
            \begin{subfigure}[t]{0.16\textwidth}
                    \includegraphics[width=\textwidth]{ 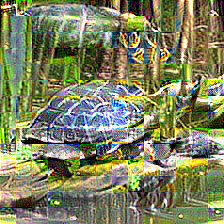 }
            \end{subfigure}%
                \hspace{0.5mm}%
            \begin{subfigure}[t]{0.16\textwidth}
                    \includegraphics[width=\textwidth]{ 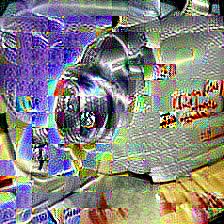 }
            \end{subfigure}%
                \hspace{0.5mm}%
            \begin{subfigure}[t]{0.16\textwidth}
                    \includegraphics[width=\textwidth]{ 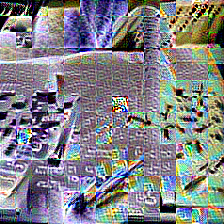 }
            \end{subfigure}%
    \end{minipage}

    \vspace{1mm}
    \begin{minipage}[t]{.025\textwidth}
         \vspace{9mm}
         \rotatebox{90}{\textbf{\rclipf}}
    \end{minipage}%
    \begin{minipage}[t]{.98\textwidth}
        \vspace{0pt}
            \begin{subfigure}[t]{0.16\textwidth}
                    \includegraphics[width=\textwidth]{ 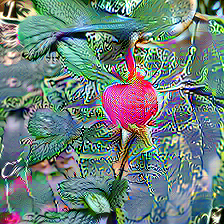 }
            \end{subfigure}%
                \hspace{0.5mm}%
            \begin{subfigure}[t]{0.16\textwidth}
                    \includegraphics[width=\textwidth]{ 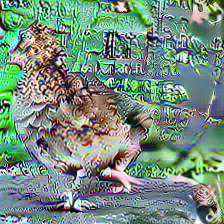 }
            \end{subfigure}%
                \hspace{0.5mm}%
            \begin{subfigure}[t]{0.16\textwidth}
                    \includegraphics[width=\textwidth]{ 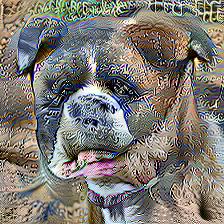 }
            \end{subfigure}%
                \hspace{0.5mm}%
            \begin{subfigure}[t]{0.16\textwidth}
                    \includegraphics[width=\textwidth]{ 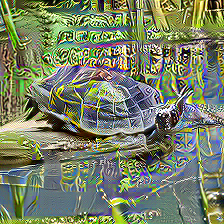 }
            \end{subfigure}%
                \hspace{0.5mm}%
            \begin{subfigure}[t]{0.16\textwidth}
                    \includegraphics[width=\textwidth]{ 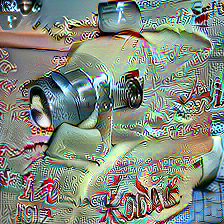 }
            \end{subfigure}%
                \hspace{0.5mm}%
            \begin{subfigure}[t]{0.16\textwidth}
                    \includegraphics[width=\textwidth]{ 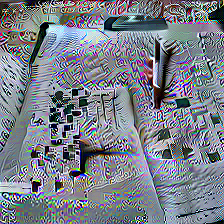 }
            \end{subfigure}%
    \end{minipage}

    \vspace{1mm}
    \begin{minipage}[t]{.025\textwidth}
         \vspace{9mm}
         \rotatebox{90}{\textbf{\rclip}}
    \end{minipage}%
    \begin{minipage}[t]{.98\textwidth}
        \vspace{0pt}
            \begin{subfigure}[t]{0.16\textwidth}
                    \includegraphics[width=\textwidth]{ 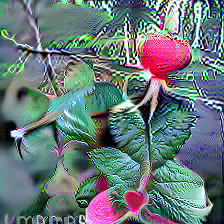 }
            \end{subfigure}%
                \hspace{0.5mm}%
            \begin{subfigure}[t]{0.16\textwidth}
                    \includegraphics[width=\textwidth]{ 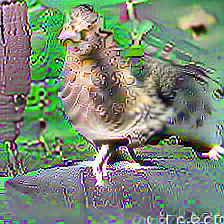 }
            \end{subfigure}%
                \hspace{0.5mm}%
            \begin{subfigure}[t]{0.16\textwidth}
                    \includegraphics[width=\textwidth]{ 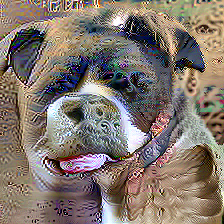 }
            \end{subfigure}%
                \hspace{0.5mm}%
            \begin{subfigure}[t]{0.16\textwidth}
                    \includegraphics[width=\textwidth]{ 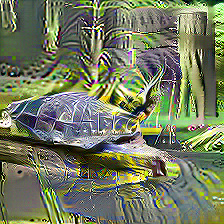 }
            \end{subfigure}%
                \hspace{0.5mm}%
            \begin{subfigure}[t]{0.16\textwidth}
                    \includegraphics[width=\textwidth]{ 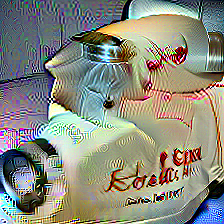 }
            \end{subfigure}%
                \hspace{0.5mm}%
            \begin{subfigure}[t]{0.16\textwidth}
                    \includegraphics[width=\textwidth]{ 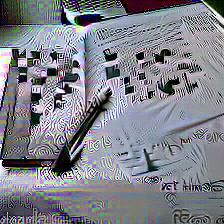 }
            \end{subfigure}%
    \end{minipage}

    \vspace{1ex}
\caption{\textbf{Feature inversion.} We reconstruct images from the embedding of respective models by optimizing a randomly initialized image to maximize similarity in the embedding space. Distinct features of the original images are reconstructed.}
\label{fig:features-inv}
\end{figure*}
}

In \cref{fig:nn-retrieval}  we show the results for the medium split of \oxford and \paris, while the complete results including the hard split can be found in \cref{tab:retrieval-roxford}. We observe that \rev{the} robust perceptual metrics (with zero-shot evaluation) achieve significantly higher robust mean Average Precision (mAP) than the baseline (close to zero for both \clip and DreamSim), at the cost of a small degradation in clean performance, with \rclipf performing best.
At the same time, they have higher mAP for both the clean and robust case than the LipSim and R-LPIPS models.
Similarly to NSFW detection, \rclipf achieves better retrieval results than \rclip and \rdinof.
Also, \cref{tab:retrieval-roxford} (in Appendix) shows the perceptual metrics fine-tuned with \lora on NIGHTS often attain lower performance than the zero-shot models, in particular for \clip and \rclipf.
Since the fine-tuned models are typically not competitive with the zero-shot ones in image-to-image retrieval, we omit them in the remaining evaluations. 

Finally, we qualitatively test the perceptual metrics for image retrieval on the MS-COCO dataset~\cite{lin2014coco-dataset}, which includes images with a broader set of subjects.
In \cref{fig:coco-short}, for each query image we show its nearest neighbor, among a random subset of 15k images from the training data, as identified by the similarity score induced by different models.
In the second row of each block, we show the same for the adversarially perturbed reference image ($\epsilon_\infty=4/255$).
\rclipf, which achieved the best retrieval results among the robust models in the previous tasks, finds, for clean queries, neighbors of similar quality compared to DreamSim, while preserving the semantic content of the retrieved image for the perturbed input in contrast to DreamSim and LipSim.
More examples are included in \cref{fig:retrieval1} and \cref{fig:retrieval2} in Appendix.

{
\figwidth=1.8\columnwidth
\begin{figure*}[t]
\centering
\footnotesize
\begin{tabular}{c@{\hspace{1mm}}|@{\hspace{1mm}}c}
\begin{subfigure}[b]{0.16\textwidth}
    \caption*{\footnotesize Original}
    \includegraphics[width=\textwidth]{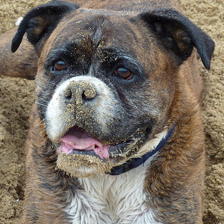}
\end{subfigure}%
\hspace{0.5mm}%
\begin{subfigure}[b]{0.16\textwidth}
    \caption*{\footnotesize Reconstructed \#1}
    \includegraphics[width=\textwidth]{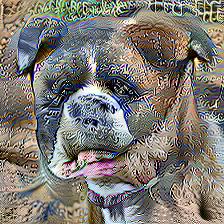}
\end{subfigure}%
\hspace{0.5mm}%
\begin{subfigure}[b]{0.16\textwidth}
    \caption*{\footnotesize Reconstructed \#2}
    \includegraphics[width=\textwidth]{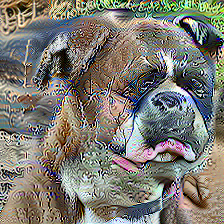}
\end{subfigure}%
&
\begin{subfigure}[b]{0.16\textwidth}
    \caption*{\footnotesize Original}
    \includegraphics[width=\textwidth]{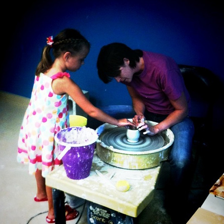}
\end{subfigure}%
\hspace{0.5mm}%
\begin{subfigure}[b]{0.16\textwidth}
    \caption*{\footnotesize Reconstructed \#1}
    \includegraphics[width=\textwidth]{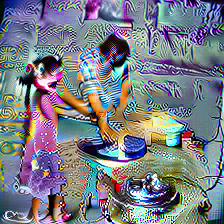}
\end{subfigure}%
\hspace{0.5mm}%
\begin{subfigure}[b]{0.16\textwidth}
    \caption*{\footnotesize Reconstructed \#2}
    \includegraphics[width=\textwidth]{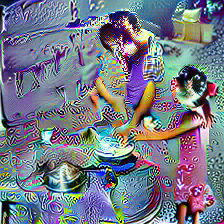}
\end{subfigure}%
\\
\begin{subfigure}[b]{0.16\textwidth}
    \includegraphics[width=\textwidth]{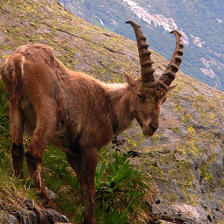}
\end{subfigure}%
\hspace{0.5mm}%
\begin{subfigure}[b]{0.16\textwidth}
    \includegraphics[width=\textwidth]{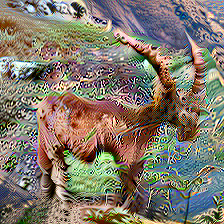}
\end{subfigure}%
\hspace{0.5mm}%
\begin{subfigure}[b]{0.16\textwidth}
    \includegraphics[width=\textwidth]{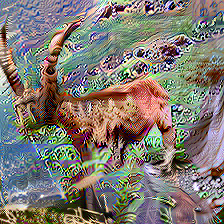}
\end{subfigure}%
&
\begin{subfigure}[b]{0.16\textwidth}
    \includegraphics[width=\textwidth]{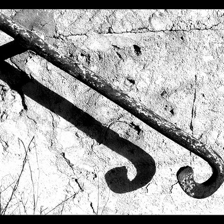}
\end{subfigure}%
\hspace{0.5mm}%
\begin{subfigure}[b]{0.16\textwidth}
    \includegraphics[width=\textwidth]{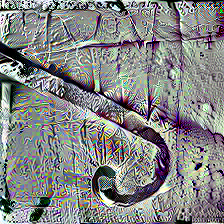}
\end{subfigure}%
\hspace{0.5mm}%
\begin{subfigure}[b]{0.16\textwidth}
    \includegraphics[width=\textwidth]{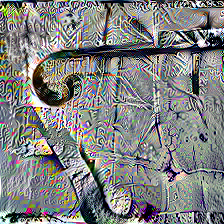}
\end{subfigure}%
\end{tabular}
\caption{\textbf{Feature inversion variants}. Varying the random seeds for the initialization, when using \rclipf, recovers multiple images for the same target feature. These are sometimes horizontally flipped 
but preserve the original semantic content.} \label{fig:image-variants}
\end{figure*}
}

\section{Visual Concepts of Robust \clip Models}

We have shown in \cref{sec:exp} and \cref{sec:image_retrieval} that robust \clip vision encoders can yield effective and robust perceptual metrics.
In the following, we explore the interpretability of such metrics, i.e. 
whether it is possible to analyze the types of similarity and concepts encoded in the underlying model.
While the better generative properties of adversarially trained image classifiers compared to clean ones are known \cite{santurkar2019imagesynthesis, augustin2020ratio, boreiko2022sparsevisual}, we show that they extend to \rev{the} robust CLIP models. Thereby, we can use this property towards explainability of the induced perceptual similarity metric (\textit{feature inversion}). Moreover, it allows us to generate visual explanations with respect to arbitrary text (\textit{text inversion}).

\subsection{Feature inversion} \label{sec:image_inversion}

To study which images are considered similar (or identical) by our perceptual metric, we aim at finding images mapped to the same embedding vector. 
We can formulate such task as finding an image $\hat{\vx}$ which maximizes the similarity to the embedding $\phi(\vx)$ of a given reference image $\vx$, i.e.,
\begin{align}
\argmax_{\hat \vx \in I}\; \simfn(\hat \vx, \vx) = \argmax_{\hat \vx \in I}\; \cos(\phi(\hat \vx), \phi(\vx)).
\label{eq:image_inversion}
\end{align}
While the problem is trivially solved by $\hat{\vx} = \vx$, the solution is not necessarily unique, 
and since we assume access to the encoded image $\phi(\vx)$ only (rather than $\vx$), it can only be approximated.
The search space $I$ is the space of all images and thus very large. 
We initialize a noisy grey image and optimize it with APGD, 
see Appendix~\ref{app:inversions} for details.
Since the formulation is analogous to that of adversarial attacks \rev{(see \cref{eq:image-inv-det})}, if the encoder $\phi$ is not robust it will not be possible to find meaningful solutions 
without imposing additional image priors \cite{mahendran2014understanding}.
This is illustrated in \cref{fig:features-inv,fig:features-inv-2} (in Appendix), where optimizing the input to match a target embedding results in non-interpretable images for clean \clip.
Conversely, using a robust encoder allows us to recover quite accurate versions of the original images, where subjects, colors and structure are well approximated, including small details like text.
This is remarkable as the embedding space of \mbox{\convnext-B} is only 640-dimensional, 
and we use simple constrained optimization.
The images obtained with \rclip are slightly sharper that with \rclipf, while 
those given by \rdinof 
show the features of the original image but often exhibit a square patches structure, likely due to the input image being divided into large patches by the \vit (while \rclipf and \rclip have \convnext as backbone). 
\rev{A quantitative evaluation of the higher quality of image reconstructed by robust metrics is presented in \cref{tab:feat-inv-quant}.}

In \cref{fig:image-variants}, we show the effect of varying initialization on the optimization results: interestingly, the reconstructed images differ mainly in non-semantic aspects, such as small translations or horizontal flip.
These examples show that the perceptual metric given by the robust \clip seems to capture the semantic content of the images well and ignores other aspects, as humans do.

{\figwidth=1.8\columnwidth
\begin{figure*}[t]
\centering
\footnotesize
    \begin{minipage}[t]{.025\textwidth}
         \vspace{0pt}
    \end{minipage}%
    \hspace{1ex}%
    \begin{minipage}[t]{.98\textwidth}
        \vspace{0pt}
            \begin{minipage}[t]{0.166\textwidth}
                \centering
                The Last Supper by Leonardo da Vinci
            \end{minipage}%
            \begin{minipage}[t]{0.166\textwidth}
                \centering
                The Scream by Edvard Munch
            \end{minipage}%
            \begin{minipage}[t]{0.166\textwidth}
                \centering
                Yoshua Bengio
            \end{minipage}%
            \begin{minipage}[t]{0.166\textwidth}
                \centering
                An astronaut riding a horse
            \end{minipage}%
            \begin{minipage}[t]{0.166\textwidth}
                \centering
                An octopus\\playing chess
            \end{minipage}%
            \begin{minipage}[t]{0.166\textwidth}
                \centering
                A dragon playing\\the piano
            \end{minipage}%
    \end{minipage}

    \vspace{1mm}
    \begin{minipage}[t]{.025\textwidth}
         \vspace{11mm}
         \rotatebox{90}{\textbf{CLIP}}
    \end{minipage}%
    \begin{minipage}[t]{.98\textwidth}
        \vspace{0pt}
            \begin{subfigure}[t]{0.16\textwidth}
                    \includegraphics[width=\textwidth]{ 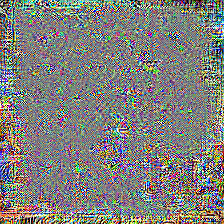 }
            \end{subfigure}%
                \hspace{0.5mm}%
            \begin{subfigure}[t]{0.16\textwidth}
                    \includegraphics[width=\textwidth]{ 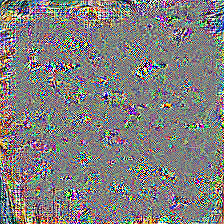 }
            \end{subfigure}%
                \hspace{0.5mm}%
            \begin{subfigure}[t]{0.16\textwidth}
                    \includegraphics[width=\textwidth]{ 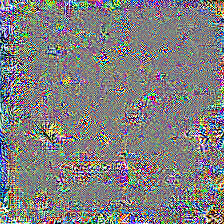 }
            \end{subfigure}%
                \hspace{0.5mm}%
            \begin{subfigure}[t]{0.16\textwidth}
                    \includegraphics[width=\textwidth]{ 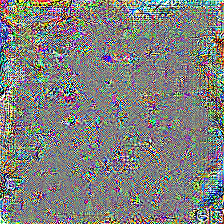 }
            \end{subfigure}%
                \hspace{0.5mm}%
            \begin{subfigure}[t]{0.16\textwidth}
                    \includegraphics[width=\textwidth]{ 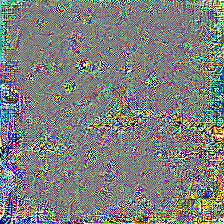 }
            \end{subfigure}%
                \hspace{0.5mm}%
            \begin{subfigure}[t]{0.16\textwidth}
                    \includegraphics[width=\textwidth]{ 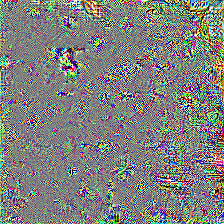 }
            \end{subfigure}%
    \end{minipage}

    \vspace{1mm}
    \begin{minipage}[t]{.025\textwidth}
         \vspace{9mm}
         \rotatebox{90}{\textbf{\rclipf}}
    \end{minipage}%
    \begin{minipage}[t]{.98\textwidth}
        \vspace{0pt}
            \begin{subfigure}[t]{0.16\textwidth}
                    \includegraphics[width=\textwidth]{ 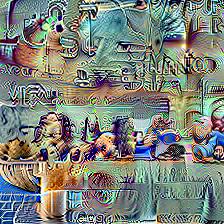 }
            \end{subfigure}%
                \hspace{0.5mm}%
            \begin{subfigure}[t]{0.16\textwidth}
                    \includegraphics[width=\textwidth]{ 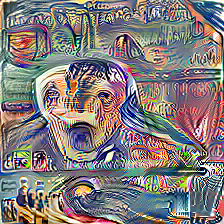 }
            \end{subfigure}%
                \hspace{0.5mm}%
            \begin{subfigure}[t]{0.16\textwidth}
                    \includegraphics[width=\textwidth]{ 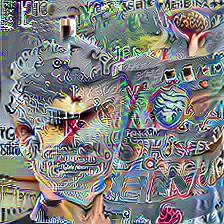 }
            \end{subfigure}%
                \hspace{0.5mm}%
            \begin{subfigure}[t]{0.16\textwidth}
                    \includegraphics[width=\textwidth]{ 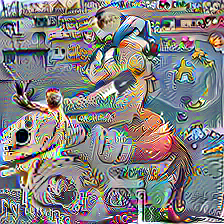 }
            \end{subfigure}%
                \hspace{0.5mm}%
            \begin{subfigure}[t]{0.16\textwidth}
                    \includegraphics[width=\textwidth]{ 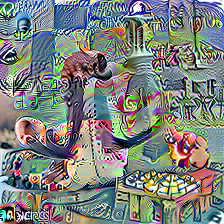 }
            \end{subfigure}%
                \hspace{0.5mm}%
            \begin{subfigure}[t]{0.16\textwidth}
                    \includegraphics[width=\textwidth]{ 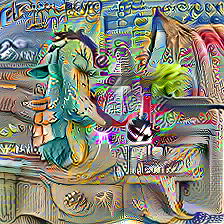 }
            \end{subfigure}%
    \end{minipage}

    \vspace{1mm}
    \begin{minipage}[t]{.025\textwidth}
         \vspace{9mm}
         \rotatebox{90}{\textbf{\rclip}}
    \end{minipage}%
    \begin{minipage}[t]{.98\textwidth}
        \vspace{0pt}
            \begin{subfigure}[t]{0.16\textwidth}
                    \includegraphics[width=\textwidth]{ 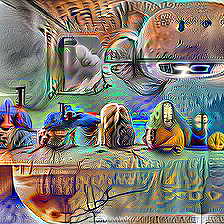 }
            \end{subfigure}%
                \hspace{0.5mm}%
            \begin{subfigure}[t]{0.16\textwidth}
                    \includegraphics[width=\textwidth]{ 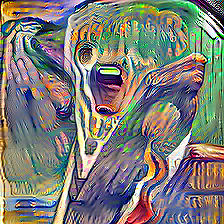 }
            \end{subfigure}%
                \hspace{0.5mm}%
            \begin{subfigure}[t]{0.16\textwidth}
                    \includegraphics[width=\textwidth]{ 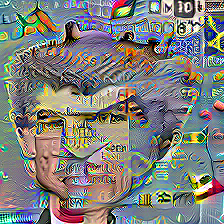 }
            \end{subfigure}%
                \hspace{0.5mm}%
            \begin{subfigure}[t]{0.16\textwidth}
                    \includegraphics[width=\textwidth]{ 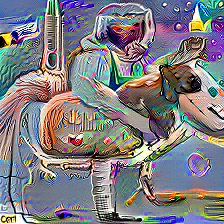 }
            \end{subfigure}%
                \hspace{0.5mm}%
            \begin{subfigure}[t]{0.16\textwidth}
                    \includegraphics[width=\textwidth]{ 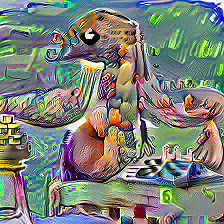 }
            \end{subfigure}%
                \hspace{0.5mm}%
            \begin{subfigure}[t]{0.16\textwidth}
                    \includegraphics[width=\textwidth]{ 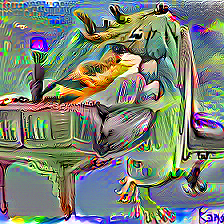 }
            \end{subfigure}%
    \end{minipage}

    \vspace{1ex}
\caption{\textbf{Text inversion.} We show visual concepts encoded in (robust) \clip models by optimizing randomly initialized images to match the given text prompts in the embedding space (see \cref{sec:text-inv}). We are able to extract rich and meaningful visual concepts from robust \clip models, while clean \clip models yield adversarial noise.}
\label{fig:text-inv}
\end{figure*}
}

\subsection{Text inversion}\label{sec:text-inv}

Since \clip includes a text encoder $\psi$ which maps to the same embedding space as the image encoder $\phi$,
we can also explore which images are associated to a given text prompt by our robust perceptual metric.
This enables visualizing which concepts can be represented, and then captured, by the metric.
In practice, given a target text $\vt$, we can solve
\begin{align}
    \argmax_{\vx \in I}\; \simfn(\vx, \vt) = \argmax_{\vx \in I}\; \cos(\phi(\vx), \psi(\vt))
    \label{eq:text_retrieval}
\end{align}
to find an image $\vx$ which has high similarity to $\vt$.
Similarly to the features inversion experiments, we optimize $\vx$ in \cref{eq:text_retrieval} with APGD starting from a noisy grey image.

While for non-robust models this mainly produces noise, with our robust \clip models clearly recognizable features of the target text are visible, see \cref{fig:text-inv}.
This shows how \clip has memorized popular subjects during training, including paintings and public figures: adversarial fine-tuning emphasizes the reliance of robust features, and allows us to extract such memorized information via optimizing the similarity score.
As feature inversion produces more realistic images than text inversion, we hypothesize that text inversion is more difficult particularly due to the modality gap~\cite{liang2022mindthegap}.

A recent work \cite{ganz2024clipag} has shown that jointly optimizing multiple augmentations of an image to align with the given text query can provide smoother images (when combined on a robust \clip encoder).
We test such approach in \cref{fig:text-inv-aug} in Appendix: even in this case, while the clean model can yield meaningful images, the results of using \rclipf and \rclip are clearly of higher quality.
Finally, we note that the approach of \cite{ganz2024clipag}, unlike our formulation in \cref{eq:text_retrieval}, does not correspond to the embedding of the vision encoder, since the features derived from the various augmented images are averaged.
Then, our robust encoders allow us to explore the visual concepts of the perceptual metric used in practice, with a straightforward optimization algorithm.

\section{Conclusion}

\textbf{Discussion.}
We have shown that fine-tuning \clip and DINO models with adversarial training provides perceptual metrics which significantly better align with human judgment than using the original clean models.
\rev{The resulting} metrics achieve SOTA performance for single encoders on 2AFC tasks, both in the zero-shot setup and after fine-tuning on perceptual data.
At the same time, such metrics inherit the adversarial robustness of the vision encoders, outperforming existing methods for robust perceptual metrics.
These findings uncover that there exist tasks where adversarial robustness is beneficial for clean performance.

Moreover, \rev{we show that 
applying the unsupervised adversarial fine-tuning \rev{of \ours \cite{schlarmann2024robust}} to a CLIP model with \convnext as backbone yields}
a perceptual metric which performs close to clean models on several image-to-image retrieval (safety-critical) tasks with significantly higher robustness.
This suggests the potential of applying adversarially robust encoders in sensitive tasks such as robust NSFW content detection and dataset filtering, especially since we show that existing classifiers like the LAION NSFW detector are vulnerable to adversarial perturbations.
We also note that the models specialized on a 2AFC task (e.g. DreamSim
) typically achieve worse performance than the zero-shot models on the retrieval tasks, suggesting some form of overfitting.

Finally, the robust encoders allow us to explore which visual concepts are taken into account for computing similarity, and what information is encoded in \clip,
by a direct and simple optimization approach.
\\

\textbf{Limitations.}
While we have tested various architectures, we did not consider different pre-training (and potentially fine-tuning) datasets and techniques, which is an important factor for their performance and could influence the perceptual metrics.
Moreover, for our  
NSFW detection we use only a small pool of retrieval images as downloading larger number of NSFW images on our servers is problematic and thus our results can likely be improved. 
\\

\textbf{Future work.}
In future work, it would be interesting to explore if robust models are also useful in other tasks such as image-text alignment for guiding generative models or image quality assessment.
Moreover, it would be important to better understand what makes robust encoders particularly effective on the 2AFC tasks we have tested, and which types of perceptual similarity they capture.

\section*{Acknowledgements}
We thank the International Max Planck Research School for Intelligent Systems (IMPRS-IS) for supporting CS and NDS.
We acknowledge support from the Deutsche Forschungsgemeinschaft (DFG, German Research Foundation) under Germany’s Excellence Strategy (EXC number 2064/1, project number 390727645), as well as in the priority program SPP 2298, project number 464101476.
We are also thankful for the support of Open Philanthropy and the Center for AI Safety Compute Cluster. Any opinions, findings, and conclusions or recommendations expressed in this material are those of the authors and do not necessarily reflect the views of the sponsors.

\bibliographystyle{abbrv}

\clearpage

\begin{appendices}

\section{Experimental Details}\label{app:experiments}

Here we provide details about the models and setup used in the experiments.

\subsection{Models and evaluation} \label{app:model-details}

\textbf{\clip models.}
We use the vision encoders from the OpenCLIP library, and in particular those of \clip models pre-trained on LAION-2B. The specific model identifiers are listed in \cref{tab:identifiers}.
We fine-tune with \ours and \tecoa for 2 epochs for the $\ell_\infty$-threat model with radius $\epsilon_\infty=4/255$, following the scheme in \cite{schlarmann2024robust}.
For fine-tuning on NIGHTS we follow the scheme of \cite{fu2023learning} (for the \convnext-B encoder we apply \lora on the fully connected layers of the MLPs).
\\

\begin{table}[h]
    \centering
    \tabcolsep=1.5pt
    \small
    \caption{\textbf{Model keys from OpenCLIP of different pre-trained encoders.}}
    \begin{tabular}{C{18mm}|C{65mm}}
         Encoder&  Identifier key\\
         \toprule
         \vit-B/32 & \texttt{CLIP-ViT-B-32-laion2B-s34B-b79K} \\
         \vit-B/16 & \texttt{CLIP-ViT-B-16-laion2B-s34B-b88K} \\
         \convnext-B & \makecell{\texttt{CLIP-convnext\_base\_w-laion2B-} \\ \texttt{s13B-b82K-augreg}} \\
         \bottomrule
    \end{tabular}
    
    \label{tab:identifiers}
\end{table}

\textbf{Baselines.}
We use the original DreamSim models, including three single encoders (OpenCLIP, \clip, DINO) and the corresponding ensemble, all fine-tuned on NIGHTS, as publicly available.\footnote{\url{https://github.com/ssundaram21/dreamsim}}
The \lipsim
metric uses a Semi-Definite program based Lipschitz Layers (SLL) convolutional network from~\cite{araujo2023unified} as the backbone. 
In the evaluation we use the original \lipsim models.\footnote{\url{https://github.com/SaraGhazanfari/lipsim}}
Finally, for R-LPIPS we use the model\footnote{\url{https://github.com/SaraGhazanfari/R-LPIPS}} trained for $\ell_\infty$-robustness on the reference image (on the BAPPS dataset).
\\

\textbf{Evaluation.}
For all models we use image resolution of 224x224, including the clean \clip with \convnext-B encoder which was pre-trained at 256x256.
While the NIGHTS dataset already contains high resolution images, which are then resized and cropped (the exact pre-processing depends on the model), the BAPPS dataset is typically used at 64x64 resolution, then we upsample the images to 224x224.
The adversarial perturbations are similarly applied on the 224x224 images.
Similar to~\cite{fu2023learning}, we use the final class token embedding as the output feature for all transformer-based models. For \convnext-based models we select the final representation.

\subsection{Visual Concepts}\label{app:inversions}

The images to be optimized are always initialized grey (all pixels $0.5$) with small additive uniform noise in $[-\nicefrac{8}{255}, \nicefrac{8}{255}]$. Then we run APGD \cite{Croce2020Autoattack} for 500 iterations with initial step-size 200 and set the $\ell_2$ radius to 100, which acts as a regularizer.

\rev{In particular, for feature inversion we solve the optimization problem 
\begin{align} 
\min_{\norm{\vdelta}_2 \leq \epsilon}\, \norm{\frac{\phi(\vz + \vdelta)}{\norm{\phi(\vz + \vdelta)}_2} - \frac{\phi(\vx)}{\norm{\phi(\vx)}_2}}_2^2 \;\;
\textrm{s. th.} \;\; \vz + \vdelta \in I, 
\label{eq:image-inv-det}
\end{align}
where $\vz$ is the initialization described above and $\vx$ the original image to reconstruct.
Also, we can rephrase text inversion as
\begin{align} 
\min_{\norm{\vdelta}_2 \leq \epsilon}\, \norm{\frac{\phi(\vz + \vdelta)}{\norm{\phi(\vz + \vdelta)}_2} - \frac{\psi(\vt)}{\norm{\psi(\vt)}_2}}_2^2 \;\;
\textrm{s. th.} \;\; \vz + \vdelta \in I, 
\label{eq:text-inv-det}
\end{align}
where $\vt$ is the target text prompt. 
We see that these formulations are similar to that of $\ell_2$-bounded adversarial attacks, and thus for non-robust image encoders the optimization finds non-interpretable images.
}

\section{Additional Experiments} \label{app:additional_experiments} 

In the following we provide additional evaluations of our robust \clip and DINO models and the induced perceptual metrics.

\subsection{Detailed Results on 2AFC Datasets} \label{app:bapps}

\textbf{Zero-shot performance on the NIGHTS dataset.}
Table~\ref{tab:additional_models_nights}
reports the clean and robust accuracy of clean and robust \clip and DINO models across different architectures on the test set of the NIGHTS dataset on both its \imnet and non-\imnet splits,\footnote{The \imnet split contains images generated from classes included in \imnet, see \cite{fu2023learning} for details} and the average over the entire set.
The advantage of the robust encoders can be observed on both test splits.
\\

\textbf{Fine-tuning different encoders on NIGHTS.}
To complement the results of \cref{tab:comparison_clip_models}, in \cref{tab:additional_models_nights} we show the performance of the perceptual metrics obtained by fine-tuning the \clip models with different backbones (ViT-B/32, ViT-B/16 and \convnext-B) and pre-training (clean, \ours, \tecoa) on the NIGHTS dataset as well as DINO (ViT-B/16) and DINOv2 (ViT-B/14) and their robust versions obtained with \ours.
When fine-tuning an MLP on top of the frozen encoder, the robust backbones preserve, across architectures, the advantage in both clean and robust accuracy they show in the zero-shot setup compared to the clean \clip.
With \lora, all backbones achieve similar clean performance, but the metrics based on robust \clip and DINO encoders are the only ones with non-trivial robustness.
\\

\textbf{Comparison to single DreamSim models.}
Additionally, we report in \cref{tab:additional_models_nights} the results of the variants of DreamSim which use a single ViT as encoder \cite{fu2023learning} and are fine-tuned on NIGHTS with \lora.
We observe that our \rclip with \convnext-B backbone plus MLP matches or improves the performance of 2 out of 3 DreamSim models, although it keeps the encoder unchanged (zero-shot setting).
Moreover, several of ours model fine-tuned with \lora perform on par with the best DreamSim model (that is OpenCLIP ViT-B/32 pre-trained on LAION-400M, while our \clip models have been pre-trained on LAION-2B).
Finally, the single DreamSim models come with robust accuracy close to zero in both threat models, unlike our metrics.
\\

\begin{figure}[t] \centering 
\tabcolsep=1.1pt
\begin{tabular}{*{2}{C{.5\columnwidth}}} 
$\ell_\infty$-threat model & $\ell_2$-threat model \\
\multicolumn{2}{c}{
\includegraphics[width=\columnwidth]{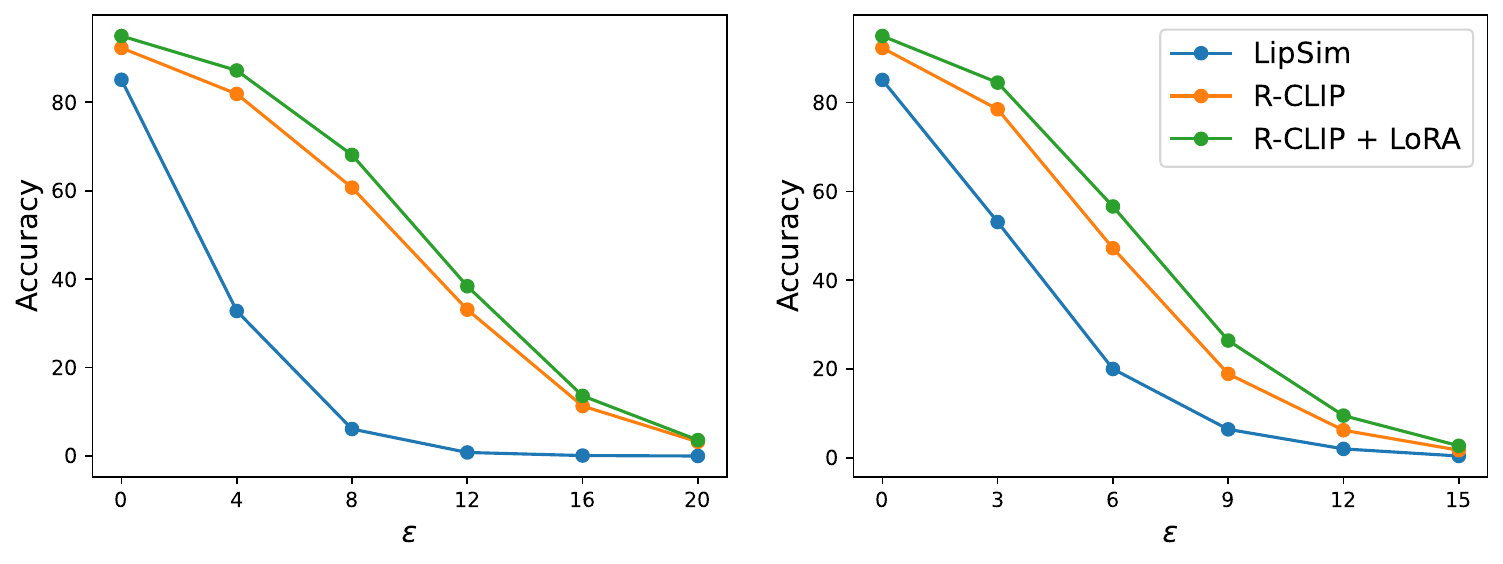}
}
\end{tabular}
\vspace{-2mm}
\caption{\textbf{Robustness at different perturbation radii for NIGHTS.} We show the robust accuracy for our \rclip (\convnext-B backbone, zero-shot and fine-tuned with \lora) and \lipsim when varying the perturbation radii, for both  $\ell_\infty$ (left, the $\epsilon$ values are shown scaled to the [0, 255] range) and $\ell_2$ (right) bounded attacks. \rclip models outperform \lipsim across perturbation sizes.
} \label{fig:curves}
\end{figure}

\textbf{Varying perturbation radius.}
We test our \rclip (zero-shot and with \lora fine-tuning, \convnext-B backbone) and the most robust \lipsim model when varying the perturbation radius for both $\ell_\infty$- and $\ell_2$-threat models.
\cref{fig:curves} shows the clean and robust accuracy of each model on the NIGHTS test set. 
We observe that our models attain higher robust accuracy than \lipsim across radii, while reaching zero at sufficiently large values.\\

\begin{figure}[t] \centering 
\includegraphics[width=\columnwidth]{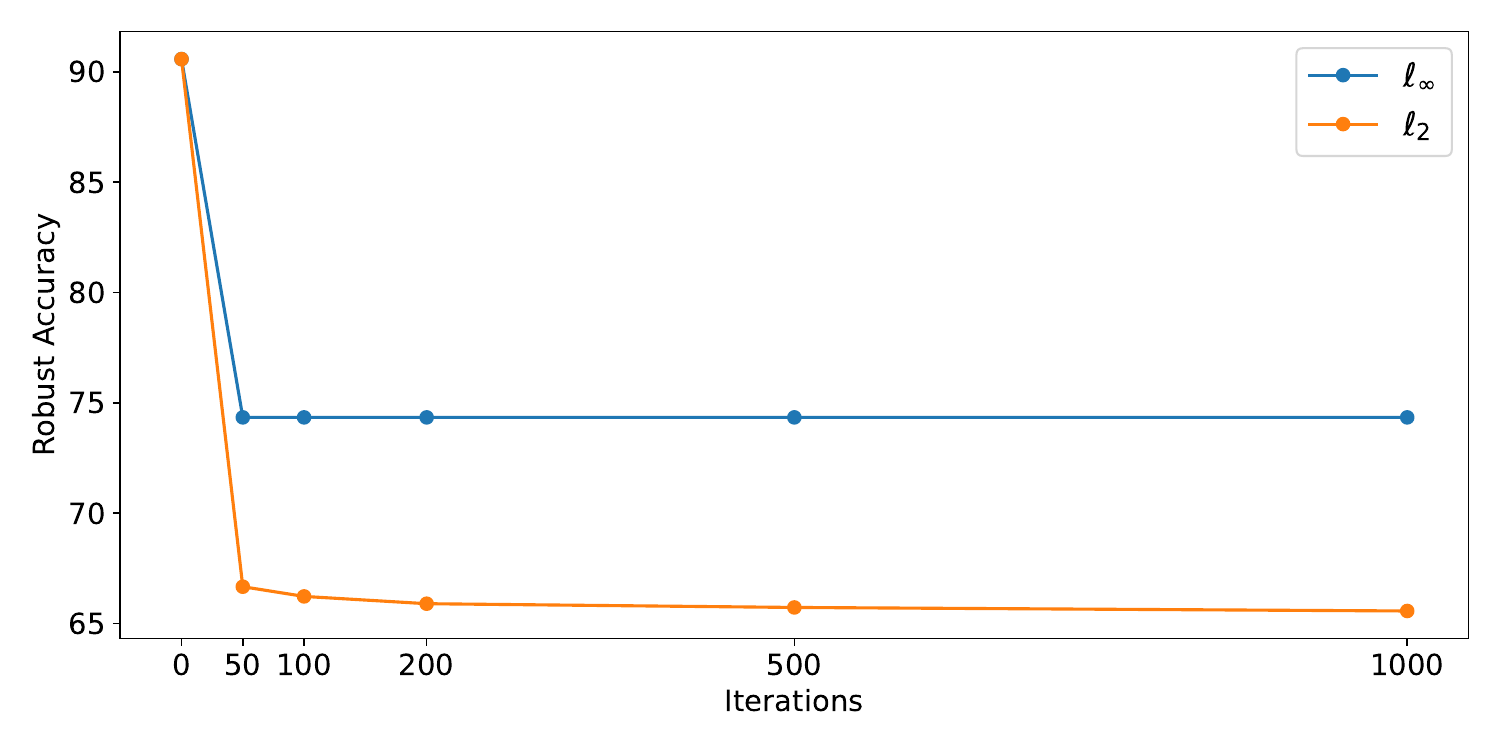}
\vspace{-6mm}
\caption{\rev{
\textbf{Attack iterations ablation.} We run the attack against \rclipf (\convnext-B backbone, zero-shot) with increasing number of iterations. For the $\ell_\infty$ threat model ($\epsilon = 4/255$) the robust accuracy remains constant. For the $\ell_2$ threat model ($\epsilon = 3$) we observe a slight decrease that levels off.
}
} \label{fig:attack-iters}
\end{figure}

\begin{table}[h]
\centering\small
\caption{\rev{
\textbf{Attack iterations ablation.} We run the attack against \rclipf (\convnext-B backbone, zero-shot) with increasing number of iterations. Here we show the exact values corresponding to \cref{fig:attack-iters}.
}
}
\begin{tabular}{c|cccccc}
    \toprule
    Iterations & 0 & 50 & 100 & 200 & 500 & 1000 \\
    \midrule
    $\ell_\infty$ & 90.6 &  74.3 &  74.3 &  74.3 &  74.3 &  74.3\\
    $\ell_2$ & 90.6 &  66.7 &  66.1 &  65.9 &  65.7 &  65.6 \\
    \bottomrule
\end{tabular}
\label{tab:attack-iters}
\end{table}

\rev{
\textbf{Varying number of iterations.} We further test the effect of varying the number of attack iterations in APGD.
As shown in \cref{fig:attack-iters} (details in \cref{tab:attack-iters}) the robustness in the $\ell_\infty$-threat model of \rclipf (\convnext-B backbone, zero-shot) remains stable when increasing the iterations up to 1000. For $\ell_2$ we observe only a small decrease (0.5\%) when going from 100 to 1000 iterations, and \cref{fig:attack-iters} shows that the robust accuracy has stabilized.
}
\\

\rev{
\textbf{Square Attack.} To further test the robustness evaluation we employ Square Attack, which is a black-box attack that does not rely on gradients. We run Square Attack against \rclipf (\convnext backbone, zero-shot) with 5000 queries on top of the APGD attack used in the main paper. The attack does not yield any additional adversarial examples over APGD neither for $\ell_\infty$ ($\epsilon=\nicefrac{4}{255})$ nor for $\ell_2$ ($\epsilon=3$) threat models, thus validating our robustness evaluation.
}
\\

\textbf{Detailed comparison on BAPPS.}
\cref{tab:details_bapps} shows the breakdown of the clean performance of the various perceptual metrics over the 6 splits of BAPPS (the entire validation set is used for this).
Consistently with NIGHTS, the adversarially trained \clip and DINO encoders provide a significant improvement compared to their clean counterparts.
Also, our models used in the zero-shot setup outperform the DreamSim ones, and are on par with the \lipsim metrics (both fine-tuned on NIGHTS).
Fine-tuning \rclipf and \rclip on NIGHTS yields some small but consistent increase in clean accuracy.
Finally, \cref{tab:details_robustness_bapps} reports the robust accuracy for all metrics in both $\ell_p$-threat models: similar to NIGHTS, \rclip and \rdinof outperforms the existing methods.
Interestingly, in this case the \rclip with \vit-B/32 backbone shows better results than the other architectures.

{\figwidth=1.8\columnwidth
\begin{figure*}[t]
\centering
\footnotesize
    \begin{minipage}[t]{.025\textwidth}
         \vspace{11mm}
         \rotatebox{90}{\textbf{Original}}
    \end{minipage}%
    \begin{minipage}[t]{.98\textwidth}
        \vspace{0pt}
            \begin{subfigure}[t]{0.16\textwidth}
                    \includegraphics[width=\textwidth]{ 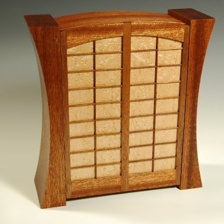 }
            \end{subfigure}%
                \hspace{0.5mm}%
            \begin{subfigure}[t]{0.16\textwidth}
                    \includegraphics[width=\textwidth]{ 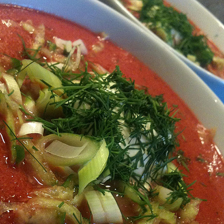 }
            \end{subfigure}%
                \hspace{0.5mm}%
            \begin{subfigure}[t]{0.16\textwidth}
                    \includegraphics[width=\textwidth]{ 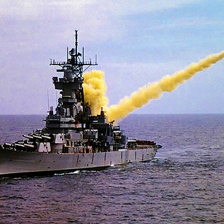 }
            \end{subfigure}%
                \hspace{0.5mm}%
            \begin{subfigure}[t]{0.16\textwidth}
                    \includegraphics[width=\textwidth]{ 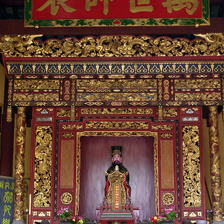 }
            \end{subfigure}%
                \hspace{0.5mm}%
            \begin{subfigure}[t]{0.16\textwidth}
                    \includegraphics[width=\textwidth]{ 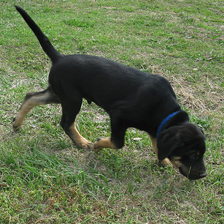 }
            \end{subfigure}%
                \hspace{0.5mm}%
            \begin{subfigure}[t]{0.16\textwidth}
                    \includegraphics[width=\textwidth]{ 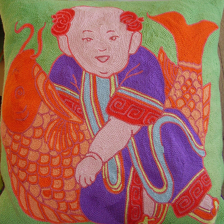 }
            \end{subfigure}%
    \end{minipage}

    \vspace{1mm}
    \begin{minipage}[t]{.025\textwidth}
         \vspace{11mm}
         \rotatebox{90}{\textbf{CLIP}}
    \end{minipage}%
    \begin{minipage}[t]{.98\textwidth}
        \vspace{0pt}
            \begin{subfigure}[t]{0.16\textwidth}
                    \includegraphics[width=\textwidth]{ 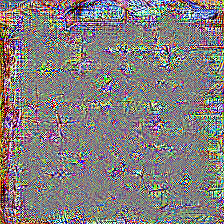 }
            \end{subfigure}%
                \hspace{0.5mm}%
            \begin{subfigure}[t]{0.16\textwidth}
                    \includegraphics[width=\textwidth]{ 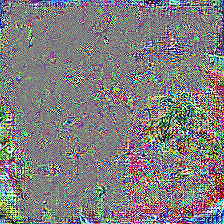 }
            \end{subfigure}%
                \hspace{0.5mm}%
            \begin{subfigure}[t]{0.16\textwidth}
                    \includegraphics[width=\textwidth]{ 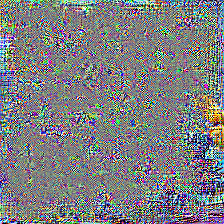 }
            \end{subfigure}%
                \hspace{0.5mm}%
            \begin{subfigure}[t]{0.16\textwidth}
                    \includegraphics[width=\textwidth]{ 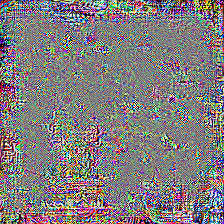 }
            \end{subfigure}%
                \hspace{0.5mm}%
            \begin{subfigure}[t]{0.16\textwidth}
                    \includegraphics[width=\textwidth]{ 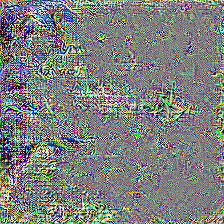 }
            \end{subfigure}%
                \hspace{0.5mm}%
            \begin{subfigure}[t]{0.16\textwidth}
                    \includegraphics[width=\textwidth]{ 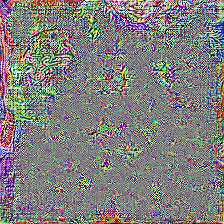 }
            \end{subfigure}%
    \end{minipage}

    \vspace{1mm}
    \begin{minipage}[t]{.025\textwidth}
         \vspace{9mm}
         \rotatebox{90}{\textbf{\rdinof}}
    \end{minipage}%
    \begin{minipage}[t]{.98\textwidth}
        \vspace{0pt}
            \begin{subfigure}[t]{0.16\textwidth}
                    \includegraphics[width=\textwidth]{ 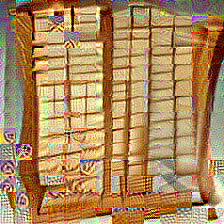 }
            \end{subfigure}%
                \hspace{0.5mm}%
            \begin{subfigure}[t]{0.16\textwidth}
                    \includegraphics[width=\textwidth]{ 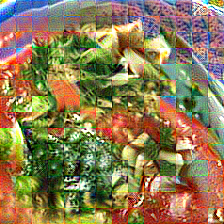 }
            \end{subfigure}%
                \hspace{0.5mm}%
            \begin{subfigure}[t]{0.16\textwidth}
                    \includegraphics[width=\textwidth]{ 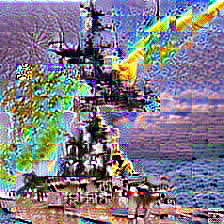 }
            \end{subfigure}%
                \hspace{0.5mm}%
            \begin{subfigure}[t]{0.16\textwidth}
                    \includegraphics[width=\textwidth]{ 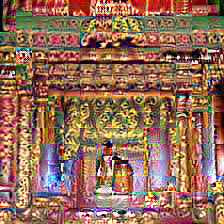 }
            \end{subfigure}%
                \hspace{0.5mm}%
            \begin{subfigure}[t]{0.16\textwidth}
                    \includegraphics[width=\textwidth]{ 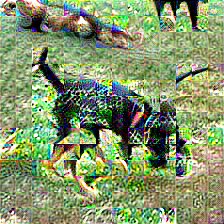 }
            \end{subfigure}%
                \hspace{0.5mm}%
            \begin{subfigure}[t]{0.16\textwidth}
                    \includegraphics[width=\textwidth]{ 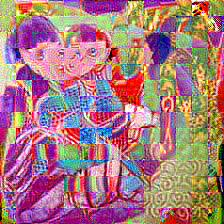 }
            \end{subfigure}%
    \end{minipage}

    \vspace{1mm}
    \begin{minipage}[t]{.025\textwidth}
         \vspace{9mm}
         \rotatebox{90}{\textbf{\rclipf}}
    \end{minipage}%
    \begin{minipage}[t]{.98\textwidth}
        \vspace{0pt}
            \begin{subfigure}[t]{0.16\textwidth}
                    \includegraphics[width=\textwidth]{ 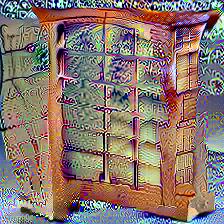 }
            \end{subfigure}%
                \hspace{0.5mm}%
            \begin{subfigure}[t]{0.16\textwidth}
                    \includegraphics[width=\textwidth]{ 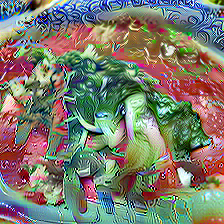 }
            \end{subfigure}%
                \hspace{0.5mm}%
            \begin{subfigure}[t]{0.16\textwidth}
                    \includegraphics[width=\textwidth]{ 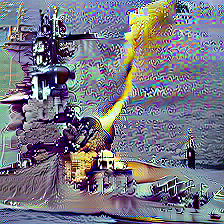 }
            \end{subfigure}%
                \hspace{0.5mm}%
            \begin{subfigure}[t]{0.16\textwidth}
                    \includegraphics[width=\textwidth]{ 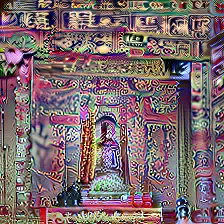 }
            \end{subfigure}%
                \hspace{0.5mm}%
            \begin{subfigure}[t]{0.16\textwidth}
                    \includegraphics[width=\textwidth]{ 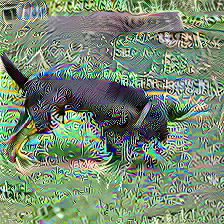 }
            \end{subfigure}%
                \hspace{0.5mm}%
            \begin{subfigure}[t]{0.16\textwidth}
                    \includegraphics[width=\textwidth]{ 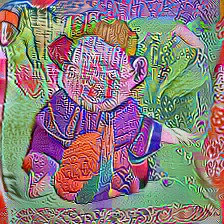 }
            \end{subfigure}%
    \end{minipage}

    \vspace{1mm}
    \begin{minipage}[t]{.025\textwidth}
         \vspace{9mm}
         \rotatebox{90}{\textbf{\rclip}}
    \end{minipage}%
    \begin{minipage}[t]{.98\textwidth}
        \vspace{0pt}
            \begin{subfigure}[t]{0.16\textwidth}
                    \includegraphics[width=\textwidth]{ 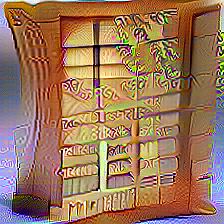 }
            \end{subfigure}%
                \hspace{0.5mm}%
            \begin{subfigure}[t]{0.16\textwidth}
                    \includegraphics[width=\textwidth]{ 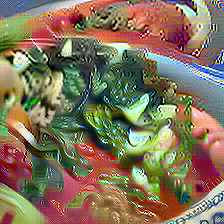 }
            \end{subfigure}%
                \hspace{0.5mm}%
            \begin{subfigure}[t]{0.16\textwidth}
                    \includegraphics[width=\textwidth]{ 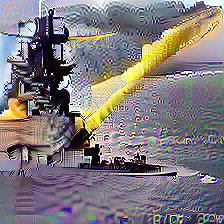 }
            \end{subfigure}%
                \hspace{0.5mm}%
            \begin{subfigure}[t]{0.16\textwidth}
                    \includegraphics[width=\textwidth]{ 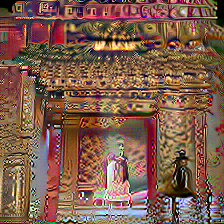 }
            \end{subfigure}%
                \hspace{0.5mm}%
            \begin{subfigure}[t]{0.16\textwidth}
                    \includegraphics[width=\textwidth]{ 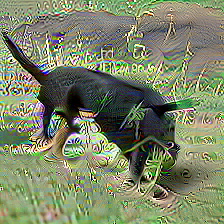 }
            \end{subfigure}%
                \hspace{0.5mm}%
            \begin{subfigure}[t]{0.16\textwidth}
                    \includegraphics[width=\textwidth]{ 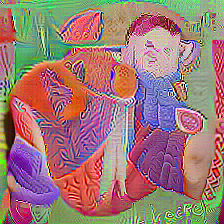 }
            \end{subfigure}%
    \end{minipage}

    \vspace{1ex}
\caption{\textbf{Feature inversion.} We show additional feature inversion images (see \cref{sec:image_inversion}).}
\label{fig:features-inv-2}
\end{figure*}
\begin{figure*}[t]
\centering
\footnotesize
    \begin{minipage}[t]{.025\textwidth}
         \vspace{0pt}
    \end{minipage}%
    \hspace{1ex}%
    \begin{minipage}[t]{.98\textwidth}
        \vspace{0pt}
            \begin{minipage}[t]{0.166\textwidth}
                \centering
                The Last Supper by Leonardo da Vinci
            \end{minipage}%
            \begin{minipage}[t]{0.166\textwidth}
                \centering
                The Scream by Edvard Munch
            \end{minipage}%
            \begin{minipage}[t]{0.166\textwidth}
                \centering
                Yoshua Bengio
            \end{minipage}%
            \begin{minipage}[t]{0.166\textwidth}
                \centering
                An astronaut riding a horse
            \end{minipage}%
            \begin{minipage}[t]{0.166\textwidth}
                \centering
                An octopus\\playing chess
            \end{minipage}%
            \begin{minipage}[t]{0.166\textwidth}
                \centering
                A dragon playing\\the piano
            \end{minipage}%
    \end{minipage}

    \vspace{1mm}
    \begin{minipage}[t]{.025\textwidth}
         \vspace{11mm}
         \rotatebox{90}{\textbf{CLIP}}
    \end{minipage}%
    \begin{minipage}[t]{.98\textwidth}
        \vspace{0pt}
            \begin{subfigure}[t]{0.16\textwidth}
                    \includegraphics[width=\textwidth]{ 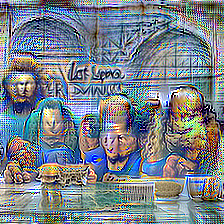 }
            \end{subfigure}%
                \hspace{0.5mm}%
            \begin{subfigure}[t]{0.16\textwidth}
                    \includegraphics[width=\textwidth]{ 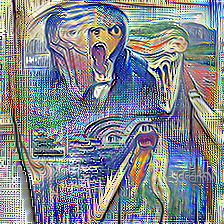 }
            \end{subfigure}%
                \hspace{0.5mm}%
            \begin{subfigure}[t]{0.16\textwidth}
                    \includegraphics[width=\textwidth]{ 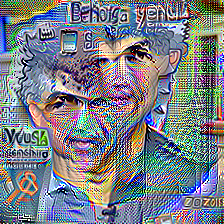 }
            \end{subfigure}%
                \hspace{0.5mm}%
            \begin{subfigure}[t]{0.16\textwidth}
                    \includegraphics[width=\textwidth]{ 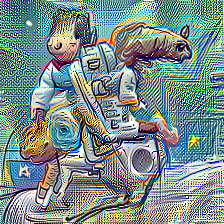 }
            \end{subfigure}%
                \hspace{0.5mm}%
            \begin{subfigure}[t]{0.16\textwidth}
                    \includegraphics[width=\textwidth]{ 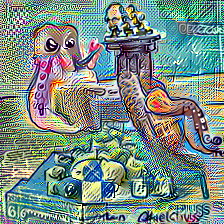 }
            \end{subfigure}%
                \hspace{0.5mm}%
            \begin{subfigure}[t]{0.16\textwidth}
                    \includegraphics[width=\textwidth]{ 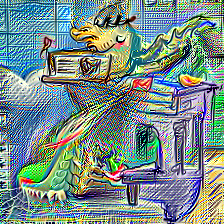 }
            \end{subfigure}%
    \end{minipage}

    \vspace{1mm}
    \begin{minipage}[t]{.025\textwidth}
         \vspace{9mm}
         \rotatebox{90}{\textbf{\rclipf}}
    \end{minipage}%
    \begin{minipage}[t]{.98\textwidth}
        \vspace{0pt}
            \begin{subfigure}[t]{0.16\textwidth}
                    \includegraphics[width=\textwidth]{ 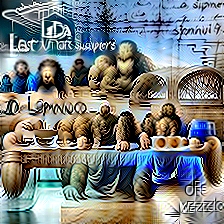 }
            \end{subfigure}%
                \hspace{0.5mm}%
            \begin{subfigure}[t]{0.16\textwidth}
                    \includegraphics[width=\textwidth]{ 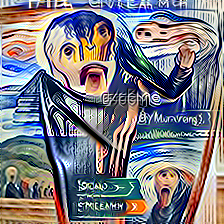 }
            \end{subfigure}%
                \hspace{0.5mm}%
            \begin{subfigure}[t]{0.16\textwidth}
                    \includegraphics[width=\textwidth]{ 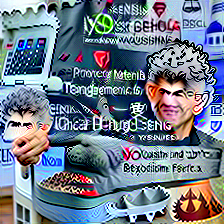 }
            \end{subfigure}%
                \hspace{0.5mm}%
            \begin{subfigure}[t]{0.16\textwidth}
                    \includegraphics[width=\textwidth]{ 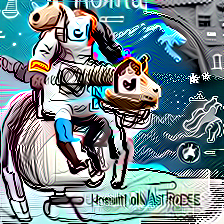 }
            \end{subfigure}%
                \hspace{0.5mm}%
            \begin{subfigure}[t]{0.16\textwidth}
                    \includegraphics[width=\textwidth]{ 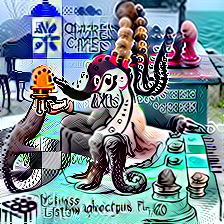 }
            \end{subfigure}%
                \hspace{0.5mm}%
            \begin{subfigure}[t]{0.16\textwidth}
                    \includegraphics[width=\textwidth]{ 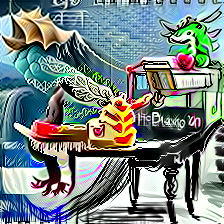 }
            \end{subfigure}%
    \end{minipage}

    \vspace{1mm}
    \begin{minipage}[t]{.025\textwidth}
         \vspace{9mm}
         \rotatebox{90}{\textbf{\rclip}}
    \end{minipage}%
    \begin{minipage}[t]{.98\textwidth}
        \vspace{0pt}
            \begin{subfigure}[t]{0.16\textwidth}
                    \includegraphics[width=\textwidth]{ 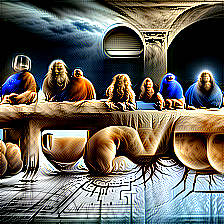 }
            \end{subfigure}%
                \hspace{0.5mm}%
            \begin{subfigure}[t]{0.16\textwidth}
                    \includegraphics[width=\textwidth]{ 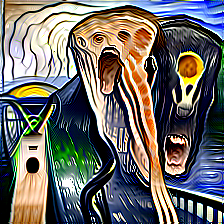 }
            \end{subfigure}%
                \hspace{0.5mm}%
            \begin{subfigure}[t]{0.16\textwidth}
                    \includegraphics[width=\textwidth]{ 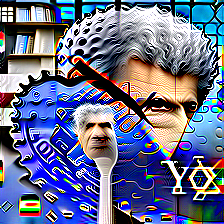 }
            \end{subfigure}%
                \hspace{0.5mm}%
            \begin{subfigure}[t]{0.16\textwidth}
                    \includegraphics[width=\textwidth]{ 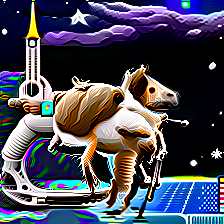 }
            \end{subfigure}%
                \hspace{0.5mm}%
            \begin{subfigure}[t]{0.16\textwidth}
                    \includegraphics[width=\textwidth]{ 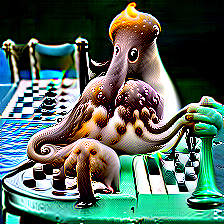 }
            \end{subfigure}%
                \hspace{0.5mm}%
            \begin{subfigure}[t]{0.16\textwidth}
                    \includegraphics[width=\textwidth]{ 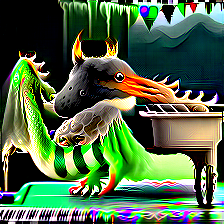 }
            \end{subfigure}%
    \end{minipage}

    \vspace{1ex}
\caption{\textbf{Text inversion with multi-view augmentations.} Using multi-view augmentations in the text-inversion process improves image quality across models, robust models still yield best images.
Note that perceptual metrics do not use multi-view augmentations at inference.}
\label{fig:text-inv-aug}
\end{figure*}
}

\subsection{Feature inversion}

\textbf{Additional images.} 
We show in \cref{fig:features-inv-2} additional examples of feature inversion as described in \cref{sec:image_inversion}.
\\

\rev{
\textbf{Quantitative evaluation.} 
We report the cosine similarity between original images and images that are reconstructed via feature inversions in Table~\ref{tab:feat-inv-quant}. We evaluate clean and robust \convnext-based \clip models and \vit-B/16-based DINO models. 100 sample images from COCO are reconstructed with each perceptual metric, and subsequently their similarity to the original images is computed with all metrics.
Thereby, we can measure the agreement of the various perceptual metrics on the image quality.
We see in \cref{tab:feat-inv-quant} that, generally, models assign highest cosine similarity to images that where reconstructed with the same model.
However, both clean DINO and \rdinof give higher similarity to the reconstructions obtained from \rclipf and \rclip compared to those from clean CLIP. Moreover, the same observations hold true when using CLIP models as a judge and DINO models for generation.
This aligns with human perception, as the images reconstructed via clean models display only adversarial noise (see \cref{fig:features-inv,fig:features-inv-2}). Notably, this observation holds across architectures (\convnext vs \vit-B/16) and pre-training methods (\clip vs DINO).
}

\begin{table}[t]
\centering\small
\caption{\rev{\textbf{Quantitative evaluation of feature inversion.} We measure the average cosine similarities between original and reconstructed images as generated and measured by \convnext-based \clip models and \vit-B/16-based DINO models. The robust perceptual metrics consistently assign a low score to feature inversion generated by clean models, while preferring those generated by robust models, thus aligning with human perception (see Figure~\ref{fig:features-inv}).
}
}
\tabcolsep=2pt
\begin{tabular}{L{4mm}L{15mm}C{10mm}C{12mm}C{12mm}C{12mm}C{13mm}}
    
     & & \multicolumn{5}{c}{\textbf{Evaluate w/}} \\
     \addlinespace[2pt]
     & & \clip & \rclip & \rclipf & DINO & \rdinof \\
     \toprule
     \multirow{3}{*}{\rotatebox{90}{\textbf{Generate w/}}}
     & \multicolumn{1}{l|}{\clip} & 0.99 &   0.64 &   0.34 &   0.07 &   0.09 \\
     & \multicolumn{1}{l|}{\rclip} &    0.64 &   1.00 &   0.94 &   0.49 &   0.65 \\
     & \multicolumn{1}{l|}{\rclipf} & 0.70 &   0.96 &   1.00 &   0.43 &   0.53 \\
     & \multicolumn{1}{l|}{DINO} & 0.26 &   0.66 &   0.30 &   1.00 &   0.31 \\
     & \multicolumn{1}{l|}{\rdinof} & 0.52 &   0.82 &   0.66 &   0.82 &   1.00 \\
     \bottomrule
\end{tabular}
\label{tab:feat-inv-quant}
\end{table}

\subsection{Text inversion
}
\label{app:text-inv-alternative}

\textbf{Optimization with multiple augmentations.}
It has been observed that integrating augmentations of multiple views of the input image into the forward pass improves the quality of images generated with gradient-based optimization \cite{ganz2024clipag,kazemi2024inverting,fort2024ensemble}. We test whether this also helps in our text-inversion setting.
To this end, we run the optimization following the setup of \cite{ganz2024clipag}, i.e. using the Adam optimizer \cite{kingma2015adam} for 1000 iterations with initial step-size 0.1 and independently augmenting 32 views of the image via diffaugment \cite{zhao2020diffaugment} with color, translation and cutout augmentations. The results are shown in \cref{fig:text-inv-aug}. We observe that this procedure yields interpretable images even for the clean \clip model, but robust clip models are superior in quality, in particular they have significantly less high-frequency noise.
However, we are more interested in the interpretability of the perceptual metric than in text-to-image generation. Evaluating the model simultaneously on multiple views of the input image conceptually deviates from the perceptual metric used at inference. Thereby, we use direct optimization of the perceptual metrics in the main paper.

\subsection{Evaluation of models on different tasks} \label{app:other_tasks}

\textbf{Robustness on \imnet and zero-shot classification.}
It is interesting to see if the performance on the perceptual metric task is correlated with other properties of \clip models like zero-shot classification. 
Therefore, we test the clean and robust models on \imnet (note that the adversarial fine-tuning is done on this dataset, in a supervised setting for \tecoa and unsupervised for \ours) and on 13 zero-shot image classification datasets, similar to~\cite{schlarmann2024robust}. In \cref{tab:zero-shot-imnet}, we report the clean and robust accuracy for the $\ell_\infty$-threat model at perturbation strengths of $2/255$ and~$4/255$. Robustness is computed with the first two attacks of AutoAttack \cite{Croce2020Autoattack}, i.e. APGD on the cross-entropy and 
targeted DLR loss. As expected, the two epoch adversarial fine-tuning results in a decay in clean performance with a significant robustness gain across architectures. Thus, while we see for zero-shot classification the usual robustness-accuracy trade-off, this does not happen for the induced perceptual metric on the 2AFC tasks. Exploring this difference is an interesting future research direction.
\\

\textbf{Performance on the THINGS dataset.}
In~\cref{tab:things}, we show how different perceptual models perform on the THINGS dataset~\cite{hebartthings}, which contains image triplets with categorical variations and classifies the odd-one-out. Our \rclipf performs the best followed by the clean \clip, whereas all fine-tuned models are notably worse.
This finding is in line with~\cite{fu2023learning}, who drew the similar conclusion that fine-tuning on NIGHTS degrades the performance on THINGS.

\subsection{Robust NSFW classification}
\label{sec:image-retrieval-nsfw}

\begin{table*}[t]
\centering\small
\caption{\textbf{Robust NSFW detection.} We consider two scenarios: \textit{(i)} query images are from $\mathcal{S}$ and target is $\mathcal{U}$ and, \textit{(ii)} query images are from $\mathcal{U}$ and target is from $\mathcal{S}$. We report for both cases the fraction of points allocated to each of the 3 classes with and without (clean) adversarial attacks. 
}
\tabcolsep=2pt
\extrarowheight=1.5pt
\newl=8mm
\newlc=8mm
\begin{tabular}{L{16mm} C{22mm} |
>{\columncolor{lightgreen}}C{\newlc}
>{\columncolor{lightgreen}}C{\newlc}
>{\columncolor{lightgreen}}C{\newlc}
C{\newlc}C{\newlc}C{\newlc}|
>{\columncolor{lightgreen}}C{\newlc}
>{\columncolor{lightgreen}}C{\newlc}
>{\columncolor{lightgreen}}C{\newlc}
C{\newlc}C{\newlc}C{\newlc}}
 & & \multicolumn{6}{c}{Query: $\mathcal{S}\quad $ Target: $\mathcal{U}$} & \multicolumn{6}{c}{Query: $\mathcal{U}\quad $Target: $\mathcal{S}$}\\
 \cline{3-8} \cline{9-14}
\addlinespace[0.5mm]
 &  & \multicolumn{3}{c}{clean} & \multicolumn{3}{c|}{$\ell_\infty (\nicefrac{8}{255})$}  & \multicolumn{3}{c}{clean} & \multicolumn{3}{c}{$\ell_\infty (\nicefrac{8}{255})$} \\
Method & Encoder & $\mathcal{S}$ & $\mathcal{B}$ &  $\mathcal{U}$ & $\mathcal{S}$ &  $\mathcal{B}$ &  $\mathcal{U}$ & $\mathcal{S}$ &  $\mathcal{B}$ & $\mathcal{U}$ &  $\mathcal{S}$ &  $\mathcal{B}$ & $\mathcal{U}$\\
\toprule
\addlinespace[1.5mm]
\multicolumn{10}{l}{Perceptual model: \textbf{\lipsim}}\\
\toprule
Pretrained & SLL &66.4 & 20.8 & 12.8& 9.6 & 32.0& 58.4 &5.6 & 32.8 & 61.6 & 61.6 & 12.6 &25.8 \\
Margin$_{0.5}$ & SLL & 69.2 & 16.0 &14.8 & 35.0 & 15.2& 49.8 & 21.6 & 12.6& 65.8& 50.2 & 23.6 & 26.2 \\
\midrule
\addlinespace[1.5mm]
\multicolumn{10}{l}{Perceptual model: \textbf{Robust LPIPS}}\\
\toprule
 R-LPIPS & AlexNet & 40.2 & 25.6 & 34.2 & 7.0 & 38.0 & 55.0 & 10.2 & 27.2 & 62.6 & 41.2 & 23.6 & 35.2 \\
\midrule
\addlinespace[1.5mm]
\multicolumn{10}{l}{Perceptual model: \textbf{DreamSim}}\\
\toprule
DINO &  \vit-B/16  & 72.2 & 14.0 &13.8 & 5.0 & 11.4 & 83.6 & 0.6 &6.6 & 92.8 & 63.8 & 16.6 & 19.6\\
Ensemble &  ViT-B/16 ($\times$3) & 88.6 & 7.8 &3.6 &0.8 & 3.2 & 96.0 &0.2 &4.2 & \textbf{95.6} & 84.0 & 10.6 & 5.4\\
\midrule
\addlinespace[1.5mm]
\multicolumn{10}{l}{Perceptual model: \textbf{DINO}}\\
\toprule
DINO & \vit-B/16 & 79.0 & 10.8 & 10.2 & 4.8 & 12.7 & 82.5 & 2.8 & 10.4 & 86.8 & 77.1 & 10.2 & 12.7 \\
\rdinof &  \vit-B/16  & 80.2 & 8.8 & 11.0 & 54.6 & 8.7 & 36.7 & 12.0 & 3.6 & 84.4 & 26.9 & 6.0 & 67.1 \\
\midrule
\addlinespace[1.5mm]
\multicolumn{10}{l}{Perceptual model: \textbf{\clip}}\\
\toprule
  \clip & ViT-B/16 & \textbf{89.6} & 4.8 & 5.6 & 0.0 & 6.5 & 93.5 & 0.2 & 7.2 & 92.6 & 88.4 & 7.6 & 4.0\\
  \rclipf & ViT-B/16 &  85.2 & 7.4 & 7.4 & 54.4 & 15.2 & 30.4 & 2.4 & 2.8 & 94.8 & 18.4 & 7.2 & 74.4\\
 \rclip & ViT-B/16 & 64.2 & 26.2 & 9.6 & 44.8 & 11.0 & 44.2 & 44.8 & 23.0 & 32.2 &  46.8 & 16.4 & 36.8\\
 \midrule
  \clip & \convnext-B & 89.4 & 6.6 & 4.0 & 1.0 & 9.8 & 89.2 &  0.2 & 8.2 &91.6 & 89.0 & 6.8 & 4.2\\
  \rclipf & \convnext-B & 88.6 &4.2 & 7.2 & 50.6 & 15.2 & 34.2 &  1.2 &6.6 & 92.2 & 18.6 & 6.4 & 75.0 \\
 \rclip & \convnext-B & 74.2 &9.0 & 16.8 & 46.6 & 21.0 & 32.4 &  9.8 & 11.2 & 79.0 & 8.2 &21.4 &70.4\\
 \midrule
 \addlinespace[1.5mm]

\multicolumn{10}{l}{Perceptual model: \textbf{\clip + \lora}}\\
\toprule
  \clip & \convnext-B & 86.2 & 7.2 & 6.6 & 0.6 & 24.8 & 74.6 & 0.4 & 23.4 & 76.2 & 86.7 & 6.9 & 6.5 \\
  \rclipf & \convnext-B & 78.4 & 10.8 & 10.8 & \textbf{57.1} & 21.7 & 21.2 &  1.4 & 13.6 & 85.0 & 9.0 & 12.2 & \textbf{78.8} \\
 \rclip & \convnext-B & 80.2 & 11.0 & 8.8 & 53.1 & 25.0 & 21.9 &  2.0 & 18.0 & 80.0 & 4.8 & 29.6 & 65.6 \\

  \bottomrule

 \label{tab:robust-retrieval}
\end{tabular}
\end{table*}

Image-to-image retrieval can be used for content filtering. For example, detecting ``Not Safe for Work'' (NSFW) images is a pressing problem as modern training datasets are often scrapped from the web~\cite{radford2021clip, schuhmann2022laion5b}, and unsafe content needs to be discarded from such datasets.
Naturally, safe-guarding filtering models against malicious users becomes an important concern.
In this section, we expand on the experimental setup for the robust NSFW detection task presented in \cref{sec:nsfw_detection}.

In~\cref{tab:robust-retrieval} we provide the detailed results of detection accuracy of different perceptual models which 
complement \cref{tab:robust-retrieval-short}.
The original \clip model performs best in clean performance on safe images (class $\mathcal{S}$), whereas the DreamSim ensemble on unsafe queries (class $\mathcal{U}$).
LipSim and R-LPIPS models have instead relatively low clean accuracy.
On this task, \rclip gets significantly worse accuracy than \clip, possibly due to the supervised fine-tuning which degrades the performance on image distributions far from that \imnet.
\rclipf, which relies on unsupervised fine-tuning, performs in fact on par with \clip, while  achieving robust detection rate of 50.6\% and 75.0\% on queries from class $\mathcal{S}$ and $\mathcal{U}$ respectively.
Conversely, the performance of both \clip and the DreamSim ensemble on adversarially perturbed images degrades below 6\%, and only the DreamSim DINO model has non-trivial robustness.
\lora fine-tuning on NIGHTS substantially degrades the clean performance of \clip and \rclipf, although preserves or slightly improves robustness.
Finally, \rdinof achieves similar clean performance to DINO, with significantly higher robust accuracy.
\\

\textbf{White-box evaluation of the LAION NSFW detector.} 
In~\cref{tab:laion-nsfw}, we only attacked the vision encoder (\vit-L/14) of the LAION NSFW detector model, assuming that the attacker does not have access to the small classifier (an MLP with 4 layers) on top of that. If the attacker has access to the full model (vision encoder and safety classifier), they can attack it in an end-to-end white-box fashion. To emulate this, we attack the full detector using APGD as optimizer for 200 iterations at an $\ell_\infty$ radius of $\epsilon=8/255$. As one would expect, this model is completely non-robust, with robust accuracy of 0\% for both cases, i.e. whether the query is from class $\mathcal{S}$ or $\mathcal{U}$. Similar to~\cref{tab:laion-nsfw} in the main part, we formulate this task as a 2-class problem and omit the buffer ($\mathcal{B}$) class.

\subsection{Nearest neighbors retrieval} \label{sec:details_image_retrieval}

\textbf{Quantitative results for image retrieval.}
We here expand the results reported in \cref{sec:nn_retrieval}.
We consider the Medium (M) and Hard (H) splits of the revisited Oxford (\oxford) and Paris (\paris) image retrieval datasets~\cite{oxfordparis, radenovic2018revisiting}, whose specific task is to find the images portraying the same landmark as the query image, 
and report the mean Average Precision (mAP). 
For both \oxford and \paris the number of query images is 70 whereas the data pool for retrieval contains 5k and 6.3k images respectively, and we always evaluate with image size $224\times 224$ with single-scale, unlike~\cite{caron2021emerging} who evaluate $\mathcal{R}$Paris at a higher resolution with multi-scale view.
For adversarial evaluation, we use an $\ell_\infty$ radius of $\epsilon=4/255$, 
and APGD (100 iterations) as optimizer. 
\cref{tab:retrieval-roxford} shows the detailed results summarized in \cref{fig:nn-retrieval}: the clean \clip models (ViT-B/16 and \convnext backbones) yield the best clean performance, 
with the DreamSim-Ensemble slightly worse,
but are completely non-robust under attack. 
Conversely, the LipSim and R-LPIPS models are marginally robust, but have low clean performance. 
Our \rclipf models attain clean mAP very close to the DreamSim models, but, unlike the baselines, do not suffer significant degradation 
against adversarial attacks.
\rclipf (unsupervised) models again attain much better clean performance, in some case close to that of the original \clip, than the \rclip (supervised) models.
Finally, fine-tuning with \lora on NIGHTS mostly degrades the clean performance of \clip and \rclipf, while slightly improving that of \rclip. This suggests that fine-tuning on a specific perceptual tasks (2AFC in this case) might not translate into improvements on other tasks.
\\

\begin{table*}[t]
\centering\small
\caption{\textbf{Quantitative robust image-to-image retrieval.} We show the clean and robust mAP (mean Average Precision) on both Medium (M) and Hard (H) splits of datasets proposed by~\cite{oxfordparis} for the image retrieval task as formulated  in~\cite{radenovic2018revisiting}. The best performing model in each column is highlighted.}
\tabcolsep=2pt
\extrarowheight=1.5pt
\newl=8mm
\newlc=8mm
\begin{tabular}{L{16mm} C{22mm} *{2}{|
>{\columncolor{lightgreen}}C{\newlc}>{\columncolor{lightgreen}}C{\newlc}  C{\newl} C{\newl}}}
 &  & \multicolumn{4}{c}{\oxford} & \multicolumn{4}{|c}{\paris}\\
 \cline{3-6} \cline{7-10}
\addlinespace[0.5mm]
 &  & \multicolumn{2}{c}{clean} & \multicolumn{2}{c}{$\ell_\infty (\nicefrac{4}{255})$}  & \multicolumn{2}{c}{clean} & \multicolumn{2}{c}{$\ell_\infty (\nicefrac{4}{255})$} \\
Method & Encoder & M & H&  M & H &  M & H &  M & H\\
\toprule
\addlinespace[1.5mm]
\multicolumn{10}{l}{Perceptual model: \textbf{\lipsim}}\\
\toprule
Pretrained & SLL &14.7 & 2.1&  6.4&1.6&30.6 &9.0 &19.2 & 6.2\\
Margin$_{0.5}$ & SLL & 13.9 & 2.1 & 7.7& 1.6 & 21.2 & 7.3 & 15.6 & 5.3   \\
\midrule
\addlinespace[1.5mm]
\multicolumn{10}{l}{Perceptual model: \textbf{DreamSim}}\\
\toprule
OpenClip &  \vit-B/32  & 39.5 & 12.2 & 0.9 & 0.4 & 64.2 & 37.6 & 3.0 & 1.6 \\
DINO &  \vit-B/16  & 31.1 & 8.0 & 1.2 & 0.6 & 59.0 & 30.2 & 5.4 & 2.2 \\
Ensemble &  ViT-B/16 ($\times$3)  & 44.5 & 15.1 & 0.8 & 0.4 & 69.4 & 43.3 & 3.2 & 1.4 \\
\midrule
\addlinespace[1.5mm]

\multicolumn{10}{l}{Perceptual model: \textbf{Robust LPIPS}}\\
\toprule
 R-LPIPS & AlexNet & 7.5 & 1.7 & 2.4 & 0.8 & 10.3 & 3.0 & 7.2 & 1.8\\
\midrule
\addlinespace[1.5mm]
\multicolumn{10}{l}{Perceptual model: \textbf{DINO}}\\
\toprule
DINO & \vit-B/16 & 36.6 & 9.6 & 1.0 & 0.5 & 63.5 & 36.4 & 4.8& 2.0\\
\rdinof & \vit-B/16 & 33.0 & 7.8 & 30.6 & 7.2 & 57.5 & 29.2 & 54.0 & 26.0 \\
\midrule
\addlinespace[1.5mm]
\multicolumn{10}{l}{Perceptual model: \textbf{\clip}}\\
\toprule
\clip & \vit-B/16 & \textbf{47.2} & 16.0 & 1.0 & 0.5 & 74.3 & 51.8 & 2.8 & 1.9 \\
\rclipf & \vit-B/16 & 37.0&	10.0 &	29.2 & 9.2 & 59.7 & 33.2 & 55.9 & 28.2\\
\rclip & \vit-B/16 & 31.6&	8.0 &	 27.9 & 7.7 & 53.1 & 26.1 & 49.8 & 23.4\\
  
  \midrule
  \clip & \convnext-B & 44.7 & 14.4 & 0.9 & 0.5 & \textbf{74.6} & \textbf{52.0} & 2.4 & 1.7\\
  \rclipf & \convnext-B & 42.2 & 13.0	& \textbf{33.4} & 10.1 & 64.1 & 37.2 & \textbf{58.6} & \textbf{32.2}\\
 \rclip & \convnext-B & 34.1 & 10.3	& 32.1 & 9.3 & 58.8 & 32.2 & 57.4 & 30.2\\
   \midrule
   \addlinespace[1.5mm]
\multicolumn{10}{l}{Perceptual model: \textbf{\clip + \lora}}\\
\toprule
  \clip & \convnext-B & 45.3 & \textbf{16.2} & 0.8 & 0.4 & 70.5 & 46.3 & 2.1 & 1.3 \\
  \rclipf & \convnext-B & 34.6 & 10.5 & 33.4 & \textbf{10.4} & 57.8 & 28.6 & 55.0 & 26.0\\
 \rclip & \convnext-B & 36.1 & 10.5 & 32.9 & 9.4 & 59.0 & 32.6 & 58.0 & 31.1 \\
 
  \bottomrule

 \label{tab:retrieval-roxford}
\end{tabular}
\end{table*}

\begin{figure*}[p]
\centering
\includegraphics[width=1.7\columnwidth]{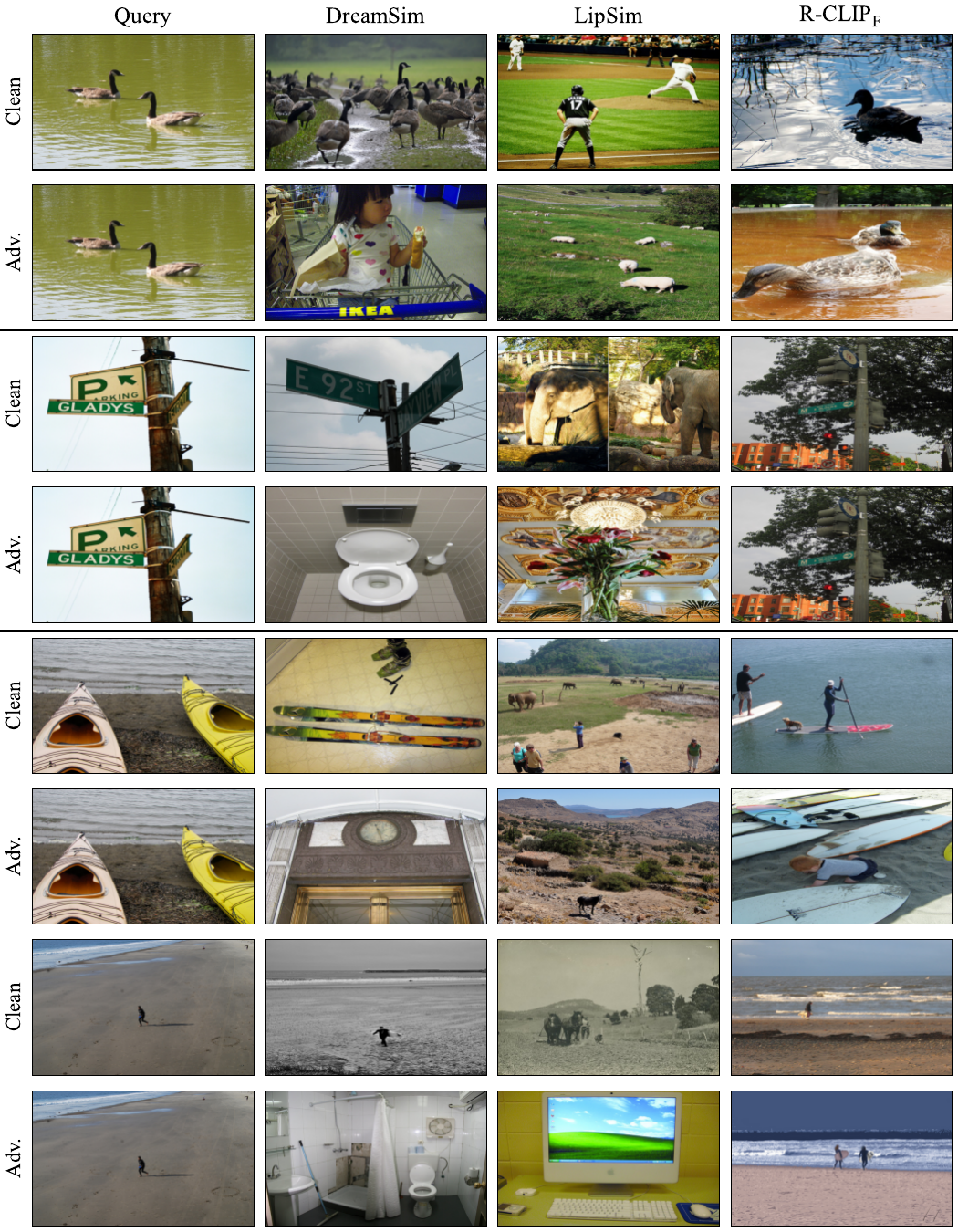}
\caption{\textbf{Clean and adversarial image retrieval on MS-COCO dataset.} Each column shows the nearest neighbor (from 15k random images from the MS-COCO training set) to the `Query' images in the first column. Adversarial images (`Adv.' rows) are generated for $\ell_\infty$ threat model at $\epsilon=4/255$ by maximizing the embedding loss of the respective vision encoders.}
\label{fig:retrieval1}

\end{figure*}

\begin{figure*}[p]
\centering
\includegraphics[width=1.7\columnwidth]{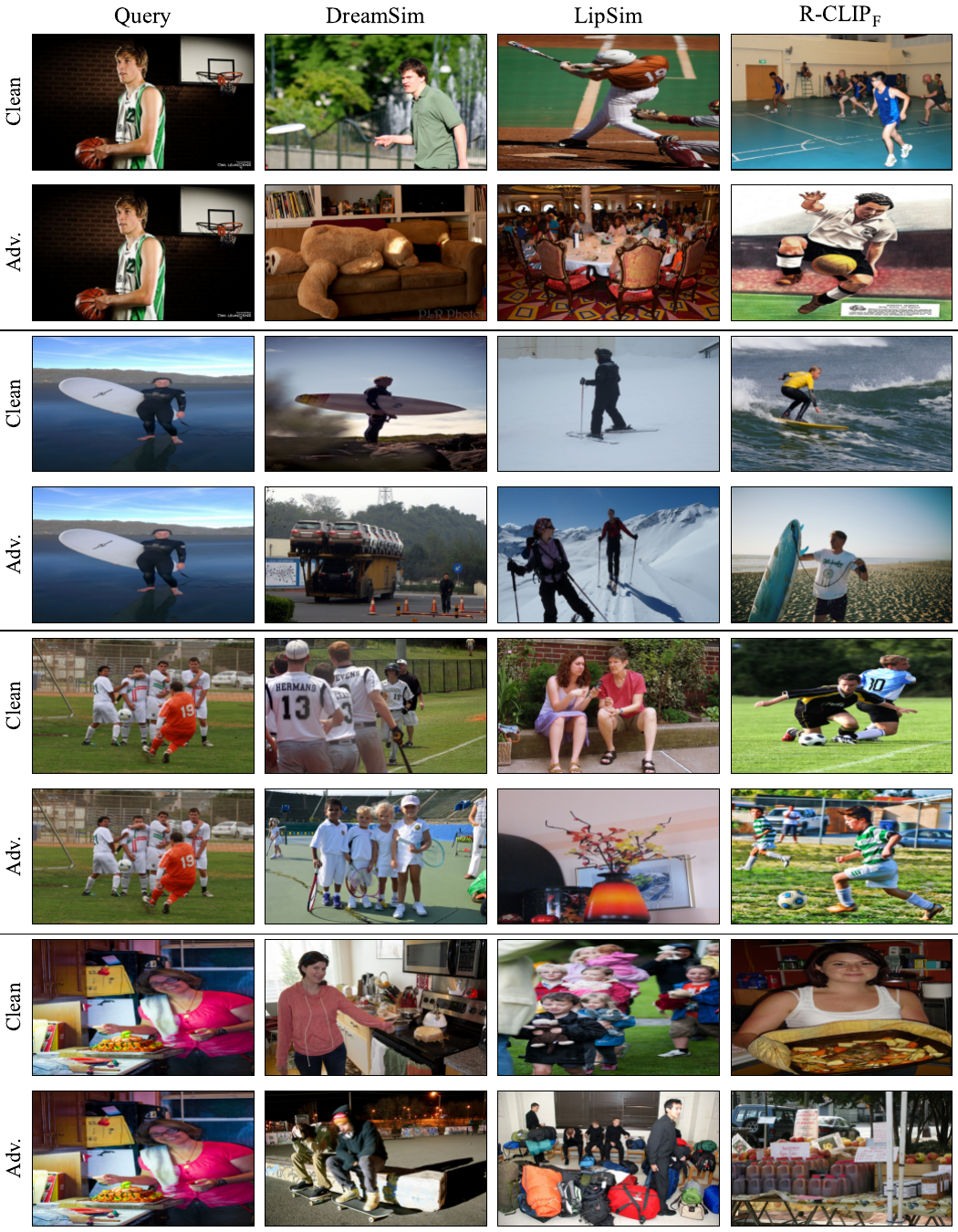}
\caption{\textbf{Clean and adversarial image retrieval on the MS-COCO dataset.} The overall setup is the same as in Fig.~\ref{fig:retrieval1}.}
\label{fig:retrieval2}

\end{figure*}

\textbf{Qualitative results for image retrieval.}
We here expand the qualitative analysis on image retrieval on the MS-COCO~\cite{lin2014coco-dataset} dataset introduced in \cref{sec:nn_retrieval}.
In \cref{fig:retrieval1} and~\cref{fig:retrieval2}, for each query image we first show, similarly to \cref{fig:coco-short}, its nearest neighbour, among a random subset of 15k images from the training data, as identified by the similarity score induced by different models.
Then, we apply on the query image an adversarial perturbation, generated at $\ell_\infty$-radius of $\epsilon=4/255$ using 50 iterations of APGD, which aims at maximizing the squared $\ell_2$-distance between the clean and the adversarial embedding (see \cref{eq:image-retrieval}).
In the `Adv.' row, we show the nearest neighbor assigned to the perturbed query. From these random set of images we can confer that \rclipf performs on this task similarly to DreamSim for clean inputs.
Moreover, we find \rclipf is less susceptible to adversarial perturbations than DreamSim and \lipsim, as in most of the cases it still retrieves a semantically similar image to the adversarially perturbed query image.

\begin{table}[h]
\centering\small
\caption{\textbf{Robust unsafe image detection for \imnet.} We consider both clean and adversarial scenario. For clean performance, we query \imnet samples against retrieval pools of the safe ($\mathcal{S}$) and unsafe ($\mathcal{U}$) sets. For adversarial evaluation, the number of query ($\mathcal{U}$) samples is 1k with $\ell_\infty$-bounded perturbations with targets from $\mathcal{S}$.}
\tabcolsep=2pt
\extrarowheight=1.5pt
\newl=18mm
\newlc=12mm
\begin{tabular}{L{13mm} C{22mm} |
>{\columncolor{lightgreen}}C{\newlc} |
>{\columncolor{lightgreen}}C{\newlc} |
C{\newl}}
 & & \multicolumn{2}{c}{clean} & \multicolumn{1}{c}{$\ell_\infty (\nicefrac{8}{255})$} \\
 \cline{3-5} 
\addlinespace[0.5mm]
 &  & \multicolumn{1}{c}{Qu: $ \mathcal{S}$} & \multicolumn{1}{c|}{Qu: $\mathcal{U}$} & \multicolumn{1}{c}{Qu: $\mathcal{U}$ Tar: $\mathcal{S}$}  \\
 \addlinespace[0.5mm]
Method & Encoder & $\mathcal{S}$ &   $\mathcal{U}$ &    $\mathcal{U}$ \\
\toprule
\addlinespace[1.5mm]

\multicolumn{4}{l}{Perceptual model: \textbf{\lipsim}}\\
\toprule
Pretrained & SLL & 65.3  & 69.3  & 29.0 \\
Margin$_{0.5}$ & SLL & 74.5  & 70.4 &   44.2 \\
\midrule
\addlinespace[1.5mm]

\multicolumn{4}{l}{Perceptual model: \textbf{Robust LPIPS}}\\
\toprule
R-LPIPS & AlexNet & 61.5  & 64.3  & 42.4 \\
\midrule
\addlinespace[1.5mm]

\multicolumn{4}{l}{Perceptual model: \textbf{DreamSim}}\\
\toprule
DINO &  \vit-B/16  & 68.9  & 92.5 &  14.4 \\

Ensemble &  ViT-B/16 ($\times$3) & 79.4 & 98.2  &  19.5 \\
\midrule
\addlinespace[1.5mm]

\multicolumn{4}{l}{Perceptual model: \textbf{DINO}}\\
\toprule
DINO &  \vit-B/16  & 79.9  & 99.0 &  20.5    \\
\rdinof &  \vit-B/16  & 70.9  & 98.0 &  89.6 \\
\midrule
\addlinespace[1.5mm]

\multicolumn{4}{l}{Perceptual model: \textbf{\clip}}\\
\toprule
  \clip & \convnext-B & 83.6  & \textbf{99.1} &  9.9 \\
  \rclipf & \convnext-B & \textbf{85.8}  &  98.0 & \textbf{92.4}  \\
 \rclip & \convnext-B & 83.2 & 97.6 &  87.4 \\
 \midrule
\addlinespace[1.5mm]

\multicolumn{4}{l}{Perceptual model: \textbf{\clip + \lora}}\\
\toprule
  \clip & \convnext-B & 75.7 & 98.2 & 20.4 \\
  \rclipf & \convnext-B & 68.0  & 97.0 & 68.4 \\
  
 \rclip & \convnext-B & 75.6  & 98.7 & 70.0 \\

  \bottomrule
 \label{tab:robust-retrieval-imnet}
\end{tabular}
\end{table}

\begin{table}[h]
\centering\small
\caption{\textbf{\imnet and zero-shot downstream classification evaluation.} We show the clean and robust accuracy for the original \clip models, \rclipf and \rclip on \imnet classification. Moreover, we show the same statistics averaged over 13 zero-shot classification datasets.}
\tabcolsep=1.1pt
\extrarowheight=1.5pt
\newl=9mm
\newlc=9mm
\begin{tabular}{L{12mm} C{17mm} *{2}{|
>{\columncolor{lightgreen}}C{\newlc} C{\newl} C{\newl}}}
 & & \multicolumn{3}{c}{\imnet} & \multicolumn{3}{c}{Avg. other datasets}\\
 \cline{3-5}\cline{5-8}
\addlinespace[0.5mm]
Method & Encoder & clean & \makecell{$\ell_\infty$\\ $ (\nicefrac{2}{255})$} &\makecell{$\ell_\infty$\\ $(\nicefrac{4}{255})$} & clean & \makecell{$\ell_\infty$\\ $(\nicefrac{2}{255})$} & \makecell{$\ell_\infty$\\ $(\nicefrac{4}{255})$} \\
\toprule
\addlinespace[0.5mm]
\clip & \vit-B/32 & 66.1 & 0.0 & 0.0 & 70.4 & 0.0 & 0.0\\
\rclipf & \vit-B/32 & 55.4 & 32.6 & 16.3 & 53.8 & 35.5 & 21.2 \\
 \rclip & \vit-B/32 & 58.3&			41.5&	25.8 & 46.8 & 34.5 & 23.3\\
  \midrule
  \clip & \vit-B/16 & 70.1  & 0.0 & 0.0 & 71.7 & 0.0 & 0.0 \\
  \rclipf & \vit-B/16 & 61.2 & 38.9 & 20.0 & 56.6 & 39.2 & 23.5 \\
 \rclip & \vit-B/16 & 64.0 &		47.9&	31.9& 51.5 & 38.4 & 26.4\\
  \midrule
  \clip & \convnext-B & 71.8 & 0.0 & 0.0 & 71.6 & 0.0 & 0.0\\
  \rclipf & \convnext-B & 62.6 & 42.3 & 23.4 & 60.2 & 44.1 & 28.41 \\ 
 \rclip & \convnext-B & 67.1 & 51.7	& 35.3 & 56.2 & 44.1 & 31.8\\
 
  \bottomrule

 \label{tab:zero-shot-imnet}
\end{tabular}
\end{table}

\begin{table}[h]
\centering 
\small
\tabcolsep=2pt
\caption{\textbf{Comparison of perceptual metrics on THINGS dataset.} We report clean accuracy on the odd-one-out task of THINGS \cite{hebartthings}. In this case fine-tuning on NIGHTS is typically detrimental for clean performance ($^*$~LipSim-Pretrained is distilled from DreamSim which in turn is fine-tuned on NIGHTS).} \label{tab:things}
\begin{tabular}{L{25mm} C{25mm}  
C{18mm} *{1}{>{\columncolor{lightgreen}}C{11mm}}} 
Method & Backbone & 
\makecell{Fine-tuning\\ dataset} & clean 
\\
\toprule
\multicolumn{4}{l}{Perceptual model: \textbf{\clip} \cite{cherti2023openclip}}\\
\toprule
 CLIP & ConvNeXt-B &  None & 50.7 \\
   \midrule
   \addlinespace[1.5mm]
   
\multicolumn{4}{l}{Perceptual model: \textbf{Robust \clip} (ours)}\\
\toprule
 \rclipf & ConvNeXt-B & None & \textbf{51.2} \\
 \rclip & ConvNeXt-B & None &48.1 \\
   \midrule
   \addlinespace[1.5mm]
   
\multicolumn{4}{l}{Perceptual model: \textbf{LipSim} \cite{ghazanfari2023lipsim}}\\
\toprule
Pretrained & SLL & NIGHTS$^*$ &43.6 \\
 $\textrm{Margin}_{0.2}$ & SLL&  NIGHTS &41.3 \\
  $\textrm{Margin}_{0.5}$ & SLL& NIGHTS & 38.5 \\
  \midrule
  \addlinespace[1.5mm]

\multicolumn{4}{l}{Perceptual model: \textbf{Robust LPIPS} \cite{ghazanfari2023rlpips}}\\
 \toprule
 R-LPIPS & AlexNet & BAPPS & 38.3 \\
 \midrule
 \addlinespace[1.5mm]
 
\multicolumn{4}{l}{Perceptual model: \textbf{Fine-tuned DreamSim} \cite{fu2023learning}}\\
\toprule
 OpenCLIP & ViT-B/32 & NIGHTS&47.9 \\
 CLIP & ViT-B/32 & NIGHTS&49.6 \\
 DINO & ViT-B/16 & NIGHTS& 44.3 \\
 Ensemble & ViT-B/16 ($\times$3) &  NIGHTS & 47.5 \\
  \midrule
  \addlinespace[1.5mm]
  
\multicolumn{4}{l}{Perceptual model: \textbf{Fine-tuned Robust \clip} (ours)}\\
\toprule

 \rclip + MLP & ConvNeXt-B & NIGHTS& 
 47.6 \\
 \rclip + \lora & ConvNeXt-B & NIGHTS& 49.9\\
 \bottomrule
\end{tabular}
\end{table}

\subsection{Automated dataset filtering via iamge-to-image retrieval} \label{sec:dataset_filtering}

Finally, in this additional task we study filtering datasets from moderated harmful content by computing similarity to harmful data pools, which might be a particularly sensitive tasks.
For instance, Birhane et al.~\cite{birhane2021multimodal} mention that LAION-400M has troublesome and explicit images and text pairs of pornography, malign stereotypes, and other extremely problematic content.
As test-case of this task, we aim to filter images of dangerous objects from the \imnet training set: the unsafe set $\mathcal{U}$ is represented by all images belonging to the three gun-related classes (\textit{assault\_rifle, rifle, revolver}), while the safe set $\mathcal{S}$ consists of random images from ten other classes including animals and common objects (\textit{desktop\_computer, car\_wheel, frilled\_lizard, lorikeet, snow plow, otter, fire\_engine, koala, tusker, African\_hunting\_dog}).
As retrieval pools, we get from the FlickrAPI\footnote{\url{https://www.flickr.com/services/api}} 500 images related to guns and firearms for the $\mathcal{U}$ class (from tags \textit{handgun, pistol, assault rifle, rifle, shotgun, sniper, bazooka, glock, klashnikov}), and 500 images of common objects and animals for the $\mathcal{S}$ class (from tags \textit{Dog, Cat, Bird, Fish, Horse, Cow, Pig, Rabbit, Chicken, Chair, Table, Pen, Notebook, Bottle, Backpack, Lamp, Wallet, Keys, Clock}).

Then, we conduct the detection experiment for safe and unsafe classes similar to the one in~\cref{sec:nsfw_detection},
but, unlike for the NSFW detection task, we do not have a buffer class $\mathcal{B}$ here. In~\cref{tab:robust-retrieval-imnet}, we show the clean detection accuracy for both cases, i.e. \textit{(i)} when the query belongs to set $\mathcal{S}$ (contains 10k images), and \textit{(ii)} when the query belongs to set $\mathcal{U}$ (contains 4k images).
For the adversarial evaluation, we keep the same setup as in~\cref{sec:nsfw_detection} with query from set $\mathcal{U}$ and target belonging to set $\mathcal{S}$ (we remark that again the query and retrieval pools for each set are disjoint).
Specifically, we again optimize the problem in~\cref{eq:image-retrieval-nsfw} with $\mathcal{S}$ set images as target and 500 iterations of APGD at $\ell_\infty$-radius of $\epsilon_\infty=8/255$.
From~\cref{tab:robust-retrieval-imnet}, we see that for clean evaluation our zero-shot \rclipf yields best detection (85.8\%) for set $\mathcal{S}$, even surpassing the DreamSim and clean \clip (\convnext backbone) models.
\rclipf also performs equally well as DreamSim in detecting unsafe ($\mathcal{U}$) images while being only 1\% behind (98.0\% to 99.1\%) the clean \clip.
For the adversarial case, both \rclipf and \rclip significantly outperform other models, with \rclipf attaining the highest (92.4\%) detection rate.
Finally, similarly to the other retrieval tasks, \lora fine-tuning on NIGHTS typically degrades the performance across models.

\begin{table*}[p]
\centering 
\caption{\textbf{Comparison of perceptual metrics on NIGHTS dataset.}
For each model we show clean and robust accuracy computed with \apgdce with 100 iterations. 
}
\footnotesize
\label{tab:additional_models_nights}
\tabcolsep=3pt
\extrarowheight=1.1pt
\newl=8mm
\newlc=8mm
\begin{tabular}{L{20mm} C{22mm} *{3}{|
>{\columncolor{lightgreen}}C{\newlc} C{\newl} C{\newlc}}}
 & & \multicolumn{3}{c}{\imnet} & \multicolumn{3}{c}{non-\imnet} & \multicolumn{3}{c}{Average}\\
 \cline{3-5}\cline{5-8} \cline{8-11}
\addlinespace[0.5mm]
Model & Backbone & \footnotesize{clean} & \footnotesize{$\ell_\infty
$} &\footnotesize{$\ell_2
$} & \footnotesize{clean} & \footnotesize{$\ell_\infty
$} &\footnotesize{$\ell_2$} & \footnotesize{clean} & \footnotesize{$\ell_\infty
$} &\footnotesize{$\ell_2$} \\
\addlinespace[0.5mm]
\toprule
\multicolumn{11}{l}{Perceptual model: \textbf{\clip}~\cite{radford2021clip, cherti2023openclip} / \textbf{Robust \clip} (ours)}\\
\toprule
 \clip & ViT-B/32 & 85.7 & 0.0 & 0.1 & 84.3 & 0.0 & 0.0 & 85.1 & 0.0 & 0.1 \\
 \rclipf & ViT-B/32 & 92.3 & 74.7 & 72.2 & 89.4 & 67.8 & 68.5 & 91.1 & 71.8 & 70.6 \\
 \rclip & ViT-B/32 & 91.9 & 81.1 & 80.8 & 89.8 & 76.5 & 78.3 & 91.0 & 79.1 & 79.7 \\
 \cmidrule{2-11}
 \clip & ViT-B/16 & 86.3 & 0.0 & 0.0 & 83.5 & 0.0 & 0.0 & 85.1 & 0.0 & 0.0 \\
 \rclipf & ViT-B/16 & 91.1 & 74.7 & 67.1 & 90.0 & 67.2 & 63.3 & 90.6 & 71.5 & 65.5 \\
 \rclip & ViT-B/16 & 93.1 & 82.5 & 78.6 & 90.3 & 75.2 & 75.1 & 91.9 & 79.4 & 77.1 \\
 \cmidrule{2-11}
 \clip & ConvNeXt-B & 88.0 & 0.0 & 0.0 & 86.1 & 0.0 & 0.0 & 87.2 & 0.0 & 0.0 \\
 \rclipf & ConvNeXt-B & 91.5 & 78.3 & 68.0 & 89.3 & 69.1 & 63.6 & 90.6 & 74.3 & 66.1 \\
 \rclip & ConvNeXt-B & 92.9 & 83.6 & 79.2 & 91.4 & 79.7 & 77.5 & 92.3 & 81.9 & 78.5 \\
 \cmidrule{2-11}
\clip & ViT-L/14 & 83.2 & 0.0 & 0.0 & 79.8 & 0.0 & 0.0 & 81.7 & 0.0 & 0.0 \\
 \rclipf & ViT-L/14 & 88.0 & 69.0 & 52.4 & 86.1 & 60.7 & 51.9 & 87.2 & 65.4 & 52.2 \\
 \rclip & ViT-L/14 & 90.4 & 79.3 & 74.6 & 87.5 & 69.2 & 68.8 & 89.1 & 74.9 & 72.1 \\
       \midrule
\multicolumn{11}{l}{Perceptual model: \textbf{DINO}~\cite{caron2021emerging}, \textbf{DINOv2}~\cite{oquab2023dinov2} / \textbf{Robust DINO} (ours)}\\
\toprule
DINO & ViT-B/16 & 90.5 & 1.3 & 1.3 & 89.9 & 0.8 & 1.3 & 90.2 & 1.1 & 1.3 \\
\rdinof & ViT-B/16 & 91.1 & 75.3 & 75.7 & 90.7 & 68.1 & 70.0 & 90.9 & 72.2 & 73.2 \\
\cmidrule{2-11}
DINOv2 & ViT-B/14 & 86.8 & 0.0 & 0.0 & 84.2 & 0.0 & 0.0 & 85.7 & 0.0 & 0.0 \\
\rdinoviif & ViT-B/14 & 89.2 & 74.0 & 67.2 & 87.5 & 65.4 & 63.6 & 88.4 & 70.3 & 65.7 \\
\midrule
\multicolumn{11}{l}{Perceptual model: \textbf{\lipsim}~\cite{ghazanfari2023lipsim}}\\
\toprule
 Pretrained & SLL & 87.0 & 8.0 & 26.5 & 86.1 & 9.5 & 26.6 & 86.6 & 8.6 & 26.5 \\
 $\textrm{Margin}_{0.2}$ & SLL & 90.2 & 22.9 & 46.9 & 86.2 & 23.5 & 46.2 & 88.5 & 23.1 & 46.6 \\
  $\textrm{Margin}_{0.5}$ & SLL & 86.1 & 33.4 & 55.1 & 83.9 & 31.9 & 50.3 & 85.1 & 32.8 & 53.1 \\
      \midrule
\multicolumn{11}{l}{Perceptual model: \textbf{R-LPIPS}~\cite{ghazanfari2023rlpips}}\\
\toprule
 R-LPIPS & AlexNet & 72.4 & 15.7 & 27.8 & 70.5 & 17.0 & 25.8 & 71.6 & 16.2 & 26.9 \\
     \midrule
\multicolumn{11}{l}{Perceptual model: \textbf{DreamSim}~\cite{fu2023learning}}\\
\toprule
 OpenCLIP & ViT-B/32 & 96.4 & 1.6 & 2.9 & 94.1 & 2.0 & 3.8 & 95.4 & 1.8 & 3.3 \\
 CLIP & ViT-B/32 & 94.1 & 0.1 & 0.3 & 93.6 & 0.1 & 0.4 & 93.9 & 0.1 & 0.3 \\
 DINO & ViT-B/16 & 94.6 & 3.1 & 5.8 & 94.4 & 4.2 & 6.8 & 94.5 & 3.6 & 6.2 \\
 Ensemble & ViT-B/16 ($\times$3) & 96.6 & 0.4 & 0.7 & 95.5 & 0.6 & 1.3 & 96.2 & 0.5 & 0.9 \\
 \midrule

 \multicolumn{11}{l}{Perceptual model: \textbf{MLP Fine-tuned \clip} (ours)} \\
\toprule
 \clip & ViT-B/32 & 91.1 & 0.0 & 0.0 & 87.5 & 0.0 & 0.0 & 89.5 & 0.0 & 0.0 \\
 \rclipf & ViT-B/32 & 94.1 & 77.5 & 74.4 & 92.7 & 69.6 & 70.1 & 93.5 & 74.1 & 72.6 \\
 \rclip & ViT-B/32  & 93.6 & 84.5 & 83.2 & 90.5 & 79.9 & 80.8 & 92.3 & 82.6 & 82.2 \\
  \cmidrule{2-11}
 \clip & ViT-B/16  & 90.2 & 0.0 & 0.0 & 87.1 & 0.0 & 0.0 & 88.9 & 0.0 & 0.0 \\
 \rclipf & ViT-B/16  & 93.2 & 79.6 & 70.4 & 92.1 & 70.9 & 65.8 & 92.7 & 75.9 & 68.4 \\
 \rclip & ViT-B/16  & 95.5 & 84.9 & 82.0 & 91.3 & 78.4 & 75.9 & 93.7 & 82.1 & 79.4 \\
 \cmidrule{2-11}
 \clip & ConvNeXt-B  & 91.2 & 0.0 & 0.0 & 89.0 & 0.0 & 0.0 & 90.2 & 0.0 & 0.0 \\
 \rclipf & ConvNeXt-B  & 93.0 & 80.7 & 70.8 & 92.0 & 74.8 & 66.7 & 92.5 & 78.2 & 69.0 \\
 \rclip & ConvNeXt-B  & 95.1 & 87.0 & 81.0 & 93.6 & 80.8 & 78.3 & 94.5 & 84.4 & 79.8 \\
\midrule

 \multicolumn{11}{l}{Perceptual model: \textbf{\lora Fine-tuned CLIP / DINO} (ours)} \\
\toprule
 \rdinof & ViT-B/16 & 95.2 & 80.5 & 80.4 & 94.8 & 75.2 & 77.5 & 95.0 & 78.2 & 79.2 \\
  \cmidrule{2-11}
\clip & ViT-B/32 & 95.6 & 0.3 & 1.0 & 93.7 & 0.4 & 0.9 & 94.8 & 0.5 & 0.9 \\
 \rclipf & ViT-B/32  & 96.1 & 83.2 & 82.1 & 94.4 & 77.5 & 79.8 & 95.3 & 80.8 & 81.1 \\
 \rclip& ViT-B/32  & 94.9 & 82.8 & 83.2 & 93.5 & 78.2 & 80.8 & 94.3 & 80.8 & 82.2 \\
 \cmidrule{2-11}
 \clip & ViT-B/16  & 95.2 & 0.0 & 0.0 & 93.6 & 0.0 & 0.0 & 94.5 & 0.0 & 0.0 \\
 \rclipf & ViT-B/16  & 95.8 & 83.0 & 78.8 & 95.5 & 78.0 & 78.3 & 95.7 & 80.9 & 78.6 \\
 \rclip & ViT-B/16  & 95.0 & 84.1 & 82.6 & 94.0 & 78.0 & 79.4 & 94.6 & 81.5 & 81.2 \\
 \cmidrule{2-11}
 \clip & ConvNeXt-B  & 95.5 & 0.0 & 0.0 & 95.3 & 0.0 & 0.0 & 95.4 & 0.0 & 0.0 \\
 \rclipf & ConvNeXt-B  & 96.0 & 87.9 & 82.2 & 94.5 & 82.5 & 80.7 & 95.3 & 85.6 & 81.6 \\
 \rclip & ConvNeXt-B  & 95.6 & 89.3 & 85.2 & 94.3 & 84.3 & 83.7 & 95.0 & 87.2 & 84.5 \\
 \bottomrule
\end{tabular}
\end{table*}

\begin{table*}[p]
\centering 
\small
\tabcolsep=3pt
\caption{\textbf{Detailed comparison of perceptual metrics on different splits of the BAPPS dataset.} 
We report the clean accuracy of each model on the 6 splits of the BAPPS dataset, together with their mean. Robust \clip encoders provide consistent improvements across splits.
} \label{tab:details_bapps}
\begin{tabular}{L{20mm} C{22mm} | *{6}{*{1}{C{10mm}}}>{\columncolor{lightgreen}}C{10mm}
} 
model & Backbone & 
cnn & color & deblur & frameint. & superr. & trad. & mean\\

\toprule
\multicolumn{9}{l}{Perceptual model: \textbf{\clip}~\cite{radford2021clip, cherti2023openclip} / \textbf{Robust \clip} (ours)}\\
\toprule
\clip & ViT-B/32 & 83.1 & 61.1 & 58.6 & 63.0 & 70.3 & 78.2 & 69.1 \\
 \rclipf & ViT-B/32 & 86.7 & 70.8 & 65.0 & 67.3 & 76.1 & 78.7 & 74.1 \\
 \rclip & ViT-B/32 & 86.5 & 71.5 & 65.0 & 67.4 & 76.6 & 77.7 & 74.1 \\
  \cmidrule{2-9}
 \clip & ViT-B/16 & 81.9 & 60.3 & 55.5 & 65.2 & 68.9 & 77.9 & 68.3 \\
 \rclipf & ViT-B/16 & 87.3 & 70.1 & 64.9 & 68.2 & 75.9 & 78.3 & 74.1 \\
 \rclip & ViT-B/16 & 86.4 & 71.2 & 64.9 & 67.2 & 76.4 & 77.7 & 74.0 \\
  \cmidrule{2-9}
  \clip & ViT-L/14 & 77.4 & 57.4 & 54.6 & 64.8 & 65.0 & 74.5 & 65.6 \\
  \rclipf & ViT-L/14 \cite{schlarmann2024robust} & 86.7 & 68.0 & 64.5 & 66.8 & 75.0 & 77.9 & 73.2 \\ 
  \rclip & ViT-L/14 \cite{schlarmann2024robust} & 86.6 & 70.8 & 65.1 & 67.2 & 75.4 & 79.0 & 74.0 \\
  \cmidrule{2-9}
 \clip & ConvNeXt-B & 82.3 & 60.0 & 55.3 & 65.9 & 67.3 & 78.5 & 68.2 \\
 \rclipf & ConvNeXt-B & 86.5 & 70.5 & 64.8 & 67.7 & 75.0 & 79.3 & 74.0 \\
\rclip & ConvNeXt-B & 86.6 & 71.4 & 64.9 & 67.5 & 75.8 & 78.6 & 74.1 \\
   \midrule
\multicolumn{9}{l}{Perceptual model: \textbf{DINO}~\cite{caron2021emerging}, \textbf{DINOv2}~\cite{oquab2023dinov2} / \textbf{Robust DINO} (ours)}\\
\toprule
DINO & ViT-B/16 & 85.1 & 62.0 & 62.2 & 67.1 & 72.9 & 79.7 & 71.5 \\
\rdinof & ViT-B/16 & 87.6 & 67.6 & 65.1 & 67.7 & 76.9 & 81.5 & 74.4 \\
\cmidrule{2-9}
DINOv2 & ViT-B/14 & 81.0 & 57.8 & 59.0 & 66.4 & 63.2 & 75.6 & 67.2 \\
\rdinoviif & ViT-B/14 & 86.9 & 67.6 & 64.6 & 68.0 & 75.5 & 78.1 & 73.5 \\
\midrule
\multicolumn{9}{l}{Perceptual model: \textbf{LipSim}~\cite{ghazanfari2023lipsim}}\\
\toprule
Pretrained & SLL & 86.4 & 69.9 & 65.6 & 66.7 & 76.8 & 79.5 & 74.2 \\
 $\textrm{Margin}_{0.2}$ & SLL & 85.2 & 71.9 & 64.7 & 66.8 & 77.2 & 77.9 & 74.0 \\
 $\textrm{Margin}_{0.5}$ & SLL & 83.6 & 71.1 & 64.1 & 66.0 & 76.8 & 77.0 & 73.1 \\
    \midrule
\multicolumn{9}{l}{Perceptual model: \textbf{R-LPIPS}~\cite{ghazanfari2023rlpips}}\\
\toprule
 R-LPIPS & AlexNet & 87.5 & 67.4 & 63.7 & 66.5 & 76.1 & 75.7 & 72.8 \\
    \midrule
\multicolumn{9}{l}{Perceptual model: \textbf{DreamSim}~\cite{fu2023learning}}\\
\toprule
OpenCLIP & ViT-B/32 & 86.4 & 67.0 & 63.0 & 65.6 & 74.7 & 81.7 & 73.1 \\
 CLIP & ViT-B/32 & 83.9 & 63.3 & 58.2 & 63.3 & 70.0 & 79.1 & 69.6 \\
 DINO & ViT-B/16 & 85.7 & 67.5 & 62.7 & 67.3 & 73.3 & 80.1 & 72.8 \\
 Ensemble & ViT-B/16 ($\times$3) 
 & 86.7 & 67.6 & 62.4 & 66.3 & 74.3 & 81.3 & 73.1 \\
    \midrule
 \multicolumn{9}{l}{Perceptual model: \textbf{MLP Fine-tuned \clip} (ours)} \\
\toprule
 \clip  & ViT-B/32 & 84.4 & 63.1 & 58.6 & 63.9 & 70.1 & 78.8 & 69.8 \\
 \rclipf  & ViT-B/32  & 86.8 & 71.4 & 65.1 & 67.0 & 76.4 & 78.2 & 74.2 \\
 \rclip & ViT-B/32 & 86.3 & 72.1 & 64.7 & 67.2 & 76.6 & 77.2 & 74.0 \\
   \cmidrule{2-9}

 \clip & ViT-B/16 & 83.2 & 62.5 & 55.8 & 65.5 & 69.4 & 78.4 & 69.1 \\
 \rclipf  & ViT-B/16  & 87.3 & 71.4 & 65.2 & 68.2 & 76.1 & 78.3 & 74.4 \\
 \rclip  & ViT-B/16 & 86.5 & 71.2 & 64.9 & 67.1 & 76.5 & 77.7 & 74.0 \\
   \cmidrule{2-9}

 \clip & ConvNeXt-B & 82.4 & 62.2 & 55.7 & 65.6 & 67.2 & 78.9 & 68.7 \\
 \rclipf  & ConvNeXt-B  & 86.8 & 70.4 & 64.9 & 67.2 & 75.6 & 78.6 & 73.9 \\
 \rclip  & ConvNeXt-B  & 86.8 & 72.0 & 65.1 & 67.4 & 75.6 & 78.3 & 74.2 \\
\midrule
 \multicolumn{9}{l}{Perceptual model: \textbf{\lora Fine-tuned CLIP / DINO} (ours)} \\
\toprule
 \rdinof & ViT-B/16 & 86.3 & 72.8 & 65.0 & 67.8 & 76.9 & 79.9 & 74.8 \\
 \cmidrule{2-9}
\clip  & ViT-B/32 & 86.0 & 67.9 & 60.9 & 64.5 & 72.7 & 81.3 & 72.2 \\
 \rclipf 
  & ViT-B/32 & 86.8 & 73.2 & 65.2 & 68.2 & 77.2 & 79.8 & 75.1 \\
 \rclip  & ViT-B/32 & 86.2 & 72.6 & 65.3 & 66.9 & 76.9 & 77.4 & 74.2 \\
   \cmidrule{2-9}

 \clip & ViT-B/16& 85.5 & 66.9 & 57.4 & 64.0 & 72.8 & 81.0 & 71.3 \\
 \rclipf  & ViT-B/16  & 86.6 & 72.1 & 65.4 & 66.4 & 76.9 & 79.7 & 74.5 \\
 \rclip & ViT-B/16 & 86.3 & 72.3 & 65.0 & 67.7 & 76.7 & 78.5 & 74.4 \\
  \cmidrule{2-9}
 \clip  & ConvNeXt-B & 85.6 & 65.8 & 58.1 & 65.3 & 72.2 & 80.4 & 71.2 \\
 \rclipf & ConvNeXt-B  & 87.5 & 72.6 & 65.0 & 67.2 & 76.3 & 80.6 & 74.9 \\
 \rclip & ConvNeXt-B  & 87.3 & 72.4 & 65.3 & 67.3 & 76.0 & 79.9 & 74.7 \\
 \bottomrule
\end{tabular}

\end{table*}

\begin{table*}[p]
\centering \small
\tabcolsep=2.5pt
\newl=7mm
\newlc=7mm
\caption{\textbf{Comparison of perceptual models on the BAPPS dataset with \apgdce 100x1.} We report both $\ell_\infty$ and $\ell_2$ robust accuracy evaluated at radii $\nicefrac{4}{255}$ and $3$ respectively for 1k samples on every split of the BAPPS dataset, and their mean.}
\label{tab:details_robustness_bapps}
\vspace{2mm}
\begin{tabular}{L{16mm} C{20mm} *{7}{|
>{\columncolor{lightgreen}}C{\newlc}C{\newl}
}}
 & & \multicolumn{2}{c|}{cnn} & \multicolumn{2}{c|}{color} & \multicolumn{2}{c|}{deblur} & \multicolumn{2}{c|}{frameinterp.} & \multicolumn{2}{c|}{superres} & \multicolumn{2}{c|}{trad.} & \multicolumn{2}{c}{mean} \\
 \addlinespace[0.5mm]
 \cline{3-16}  
  \addlinespace[0.5mm]

model & Backbone &   \scriptsize{$\ell_\infty$} &\scriptsize{$\ell_2$} &  \scriptsize{$\ell_\infty$} &\scriptsize{$\ell_2$} &  \scriptsize{$\ell_\infty$} &\scriptsize{$\ell_2$} & \scriptsize{$\ell_\infty$} &\scriptsize{$\ell_2$} &  \scriptsize{$\ell_\infty$} &\scriptsize{$\ell_2$}  & \scriptsize{$\ell_\infty$} &\scriptsize{$\ell_2$} & \scriptsize{$\ell_\infty$} &\scriptsize{$\ell_2$}
\\
\toprule
\multicolumn{16}{l}{Perceptual model: \textbf{\clip}~\cite{radford2021clip, cherti2023openclip} / \textbf{Robust \clip} (ours)}\\
\toprule
\clip & ViT-B/32 & 0.0 & 0.0 & 0.0 & 0.0 & 0.1 & 0.1 & 0.0 & 0.0 & 0.2 & 0.2 & 0.0 & 0.0 & 0.0 & 0.0 \\
 \rclipf & ViT-B/32 & 36.1 & 31.7 & 17.3 & 14.5 & 7.9 & 4.3 & 7.0 & 4.8 & 17.5 & 14.5 & 35.9 & 29.4 & 20.3 & 16.5 \\
 \rclip & ViT-B/32 & 46.3 & 44.0 & 27.9 & 28.0 & 17.4 & 14.7 & 11.6 & 10.6 & 30.3 & 25.9 & 41.7 & 38.7 & 29.2 & 27.0 \\
\cmidrule{2-16}
 \clip & ViT-B/16 & 0.0 & 0.1 & 0.0 & 0.0 & 0.1 & 0.1 & 0.0 & 0.0 & 0.3 & 0.3 & 0.0 & 0.0 & 0.1 & 0.1 \\
 \rclipf & ViT-B/16 & 34.1 & 16.7 & 15.2 & 6.8 & 6.8& 1.9 & 5.0 & 2.0 & 16.4 & 6.0 & 38.4 & 18.6 & 19.3 & 8.7 \\
 \rclip & ViT-B/16 & 45.4 & 33.7 & 24.2 & 19.5 & 16.0 & 8.2 & 10.2 & 7.0 & 27.0 & 16.8 & 44.1 & 32.3 & 27.8 & 19.6 \\
 \cmidrule{2-16}
\clip & ViT-L/14 & 0.0 & 0.0 & 0.0 & 0.0 & 0.1 & 0.1 & 0.0 & 0.0 & 0.2 & 0.2 & 0.0 & 0.0 & 0.1 & 0.1 \\
\rclipf & ViT-L/14 \cite{schlarmann2024robust} & 25.5 & 9.7 & 9.2 & 2.7 & 4.4 & 0.3 & 2.7 & 0.9 & 10.9 & 3.9 & 32.6 & 11.6 & 14.2 & 4.9 \\ 
\rclip & ViT-L/14 \cite{schlarmann2024robust} & 38.6 & 23.5 & 16.9 & 10.4 & 11.0 & 2.7 & 7.3 & 4.2 & 17.9 & 10.2 & 37.4 & 21.6 & 21.5 & 12.1 \\
\cmidrule{2-16}
 \clip & ConvNeXt-B & 0.0 & 0.0 & 0.0 & 0.0 & 0.1 & 0.1 & 0.0 & 0.0 & 0.2 & 0.2 & 0.0 & 0.0 & 0.0 & 0.0 \\
  \rclipf & ConvNeXt-B & 34.1 & 11.4 & 14.9 & 5.2 & 7.0 & 0.7 & 5.5 & 1.7 & 15.9 & 3.5 & 36.4 & 13.6 & 19.0 & 6.0 \\
 \rclip & ConvNeXt-B & 45.2 & 28.1 & 21.3 & 13.0 & 13.6 & 4.30 & 11.3 & 5.5 & 26.7 & 15.0 & 42.8 & 29.1 & 26.8 & 15.8 \\
    \midrule
\multicolumn{16}{l}{Perceptual model: \textbf{DINO}~\cite{caron2021emerging}, \textbf{DINOv2}~\cite{oquab2023dinov2} / \textbf{Robust DINO} (ours)}\\
\toprule
DINO & ViT-B/16 & 0.0 & 0.0 & 0.0 & 0.0 & 0.1 & 0.1 & 0.0 & 0.0 & 0.2 & 0.2 & 0.0 & 0.0 & 0.1 & 0.1 \\
\rdinof & ViT-B/16 & 44.1 & 36.9 & 10.6 & 7.9 & 11.3 & 4.5 & 9.1 & 7.2 & 22.4 & 15.1 & 44.7 & 37.5 & 23.7 & 18.2 \\
\cmidrule{2-16}
DINOv2 & ViT-B/14 & 0.0 & 0.0 & 0.0 & 0.0 & 0.1 & 0.1 & 0.0 & 0.0 & 0.2 & 0.2 & 0.0 & 0.0 & 0.1 & 0.1 \\
\rdinoviif & ViT-B/14 & 35.0 & 16.2 & 7.6 & 3.0 & 9.0 & 1.3 & 6.1 & 2.4 & 18.8 & 4.7 & 33.7 & 15.3 & 18.4 & 7.2 \\
\midrule
\multicolumn{16}{l}{Perceptual model: \textbf{\lipsim}~\cite{ghazanfari2023lipsim}}\\
\toprule
Pretrained & SLL& 2.8 & 15.5 & 0.3 & 3.2 & 0.1 & 1.0 & 0.0 & 0.2 & 0.9 & 4.9 & 2.4 & 19.3 & 1.1 & 7.4 \\
 Margin$_{0.2}$ &SLL & 9.7 & 25.6 & 3.8 & 11.3 & 0.1 & 2.2 & 0.0 & 1.1 & 1.7 & 7.3 & 9.8 & 28.8 & 4.2 & 12.7 \\
 Margin$_{0.5}$ & SLL & 14.0 & 28.8 & 8.3 & 20.5 & 0.2 & 2.2 & 0.0 & 1.0 & 1.7 & 6.0 & 10.7 & 31.9 & 5.8 & 15.1 \\
     \midrule
\multicolumn{16}{l}{Perceptual model: \textbf{R-LPIPS}~\cite{ghazanfari2023rlpips}}\\
\toprule
 R-LPIPS & AlexNet & 20.8 & 31.3 & 8.8 & 13.9 & 0.2 & 0.8 & 1.2 & 2.3 & 3.0 & 5.6 & 8.3 & 20.0 & 7.0 & 12.3 \\
    \midrule
\multicolumn{16}{l}{Perceptual model: \textbf{DreamSim}~\cite{fu2023learning}}\\
\toprule
OpenCLIP & ViT-B/32 & 0.0 & 0.0 & 0.0 & 0.0 & 0.1 & 0.1 & 0.0 & 0.0 & 0.2 & 0.2 & 0.0 & 0.0 & 0.0 & 0.0 \\
 CLIP & ViT-B/32 & 0.0 & 0.0 &0.0 & 0.0 & 0.1 & 0.1 & 0.0 & 0.0 & 0.2 & 0.2 & 0.0 & 0.0 & 0.0 & 0.0 \\
 DINO & ViT-B/16 & 0.0 & 0.0 & 0.0 & 0.0 & 0.1 & 0.1 & 0.0 & 0.0 & 0.2 & 0.2 & 0.1 & 0.0 & 0.0 & 0.1 \\
 Ensemble & ViT-B/16 ($\times$3) & 0.0 & 0.0 & 0.0 & 0.0 & 0.1 & 0.1 &0.0 & 0.0 & 0.2 & 0.2 &0.0 & 0.0 & 0.0 & 0.0 \\

   \midrule
\multicolumn{16}{l}{Perceptual model: \textbf{MLP Fine-tuned \clip} (ours)}\\
\toprule
\clip & ViT-B/32  & 0.0 & 0.0 & 0.0 & 0.0 & 0.1 & 0.1 & 0.0 & 0.0 & 0.2 & 0.2 & 0.0 & 0.0 & 0.0 & 0.0 \\
 \rclipf & ViT-B/32  & 38.7 & 32.6 & 21.2 & 17.5 & 9.1 & 4.7 & 7.1 & 5.5 & 18.5 & 13.8 & 38.5 & 28.9 & 22.2 & 17.2 \\
 \rclip & ViT-B/32  & 47.2 & 43.5 & 32.3 & 31.3 & 17.4 & 13.7 & 11.8 & 10.4 & 30.3 & 24.5 & 41.9 & 37.8 & 30.1 & 26.9 \\
\cmidrule{2-16}
 
 \clip & ViT-B/16 & 0.0 & 0.1 & 0.0 & 0.0 & 0.1 & 0.2 & 0.0 & 0.0 & 0.3 & 0.6 & 0.0 & 0.0 & 0.1 & 0.2 \\
 \rclipf & ViT-B/16  & 37.7 & 18.2 & 20.5 & 10.6 & 8.1 & 1.6 & 6.2 & 2.3 & 18.0 & 6.0 & 38.7 & 16.8 & 21.5 & 9.3 \\
 \rclip & ViT-B/16  & 46.3 & 33.4 & 29.0 & 21.4 & 16.7 & 7.2 & 10.6 & 6.8 & 28.3 & 16.3 & 43.7 & 31.8 & 29.1 & 19.5 \\
\cmidrule{2-16}
 \clip & ConvNeXt-B  & 0.0 & 0.0 &0.0 & 0.0 & 0.1 & 0.1 & 0.0 & 0.0 & 0.2 & 0.2 & 0.0 & 0.0 & 0.0 & 0.0 \\
\rclipf & ConvNeXt-B  & 38.6 & 14.1 & 20.7 & 7.7 & 8.6 & 0.6 & 6.7 & 2.5 & 17.4 & 3.7 & 38.2 & 14.2 & 21.7 & 7.1\\
\rclip & ConvNeXt-B  & 47.4 & 28.7 & 28.0 & 18.4 & 14.2 & 4.1 & 11.5 & 6.0 & 26.6 & 12.6 & 43.2 & 28.0 & 28.5 & 16.3 \\
  \midrule
\multicolumn{16}{l}{Perceptual model: \textbf{LoRA Fine-tuned CLIP / DINO} (ours)}\\
\toprule
\rdinof & ViT-B/16 & 43.5 & 44.3 & 27.2 & 30.6 & 10.0 & 8.8 & 10.1 & 9.6 & 19.4 & 17.6 & 44.4 & 42.3 & 25.8 & 25.5 \\
\cmidrule{2-16}
\clip & ViT-B/32  & 0.0 & 0.0 & 0.0 & 0.0 & 0.1 & 0.1 & 0.0 & 0.0 & 0.2 & 0.2 & 0.0 & 0.0 & 0.0 & 0.0 \\
 \rclipf & ViT-B/32  & 38.1 & 35.4 & 24.7 & 23.7 & 7.1 & 5.1 & 7.6 & 7.0 & 18.6 & 17.4 & 36.9 & 30.6 & 22.2 & 19.9 \\
 \rclip & ViT-B/32  & 39.3 & 42.1 & 27.0 & 28.0 & 6.8 & 7.7 & 5.6 & 6.7 & 15.6 & 18.5 & 35.3 & 35.6 & 21.6 & 23.1 \\
\cmidrule{2-16}
 
 \clip & ViT-B/16  & 0.0 & 0.0 & 0.0 & 0.0 & 0.1 & 0.1 & 0.0 & 0.0 & 0.2 & 0.2 & 0.0 & 0.0 & 0.0 & 0.0 \\
 \rclipf & ViT-B/16  & 43.3 & 31.9 & 28.9 & 22.8 & 8.5 & 3.6 & 8.3 & 4.6 & 21.5 & 9.5 & 40.7 & 27.9 & 25.2 & 16.7 \\
 \rclip & ViT-B/16  & 45.7 & 41.7 & 34.8 & 31.2 & 11.4 & 7.7 & 10.2 & 8.5 & 24.2 & 19.5 & 43.9 & 36.0 & 28.4 & 24.1 \\
 \cmidrule{2-16}
 \clip & ConvNeXt-B  & 0.0 & 0.0 & 0.0 & 0.0 & 0.1 & 0.1 & 0.0 & 0.0 &0.2 & 0.2 & 0.0 & 0.0 & 0.0 & 0.0 \\
 \rclipf & ConvNeXt-B  & 48.2 & 35.8 & 32.8 & 24.3 & 12.5 & 3.7 & 10.7 & 5.4 & 27.4 & 16.2 & 49.5 & 34.9 & 30.2 & 20.1\\
 \rclip & ConvNeXt-B  & 48.1 & 35.1 & 28.7 & 23.1 & 13.4 & 6.1 & 10.3 & 5.6 & 26.7 & 14.5 & 47.7 & 35.8 & 29.2 & 20.0 \\
\bottomrule
\end{tabular}
\end{table*}

\end{appendices}

\end{document}